\documentclass[12pt]{article}
\usepackage{etoolbox} % For patching environments

\AtBeginEnvironment{thebibliography}{%
    \small % or \footnotesize, \scriptsize, etc.
    \setlength{\baselineskip}{0.8\baselineskip} % Reduce line-spacing (adjust 0.8 as needed)
    \setlength{\itemsep}{2pt} % Reduce spacing between entries
}

\usepackage{amsmath}
\usepackage{graphicx}
\usepackage{enumerate}
\usepackage{natbib}
\usepackage{ulem} % sout

\usepackage{amsfonts}
\usepackage{amsmath}
\usepackage{amssymb}
\usepackage{amsthm}
\usepackage{thmtools}

\theoremstyle{plain}% default
\newtheorem{theorem}{Theorem}[section] %
\newtheorem{lemma}[theorem]{Lemma} %
\newtheorem{proposition}[theorem]{Proposition} %
\newtheorem{definition}{Definition}[section] %
\newtheorem{assumption}{Assumption}[section] %
\newtheorem{corollary}{Corollary} %
\newtheorem*{remark}{Remark} %

\usepackage{placeins}
\usepackage{hyperref} 
\usepackage{float}
\usepackage[font=small,labelfont=bf,tableposition=top]{caption}
\usepackage{pgfplots}
\pgfplotsset{compat=1.7}
\usepackage[export]{adjustbox}% 
\usepackage{array}
\usepackage{fancyhdr}
\usepackage{algorithm}
  \makeatletter
  \renewcommand{\ALG@name}{Framework}
  \makeatother
\usepackage{algpseudocode}
\usepackage{xr}
\usepackage{xcite}
\DeclareMathOperator*{\argmin}{arg\,min}

\newcommand{\R}{\mathbb{R}} % Real line
\newcommand{\E}{\mathbb{E}} % Expectation
\newcommand{\dnorm}[1]{ \Vert #1 \Vert } % Norm
\newcommand{\Wd}[2]{W_1\left( \Prob_{#1} ,  \Prob_{#2} \right)} % 1-WD
\newcommand{\eqd}{\stackrel{\text{\tiny d}}{=}} % equal in distribution
\newcommand{\Prob}{\mathbb{P}}
\newcommand{\simiid}{\overset{\mathrm{iid}}{\sim}}

\newcommand{\Lip}{ \text{Lip}}

\newcommand{\tD}{ \widetilde{D}}
\newcommand{\tE}{ \widetilde{E}}

\newcommand{\hProb}{ \widehat{\Prob}}

\newcommand{\cD}{ \mathcal{D}}
\newcommand{\cE}{ \mathcal{E}}

\newcommand{\cG}{ \mathcal{G}}
\newcommand{\cH}{ \mathcal{H}}

\newcommand{\cL}{ \mathcal{L}}

\newcommand{\cN}{ \mathcal{N}}
\newcommand{\cO}{ \mathcal{O}}
\newcommand{\cP}{ \mathcal{P}}

\newcommand{\cR}{ \mathcal{R}}
\newcommand{\cS}{ \mathcal{S}}

\newcommand{\cU}{ \mathcal{U}}

\newcommand{\cX}{ \mathcal{X}}
\newcommand{\cY}{ \mathcal{Y}}

\newcommand{\cZ}{ \mathcal{Z}}

\newcommand{\hD}{ \widehat{D}}
\newcommand{\hE}{ \widehat{E}}

\newcommand{\hG}{ \widehat{G}}
\newcommand{\hH}{ \widehat{H}}

\usepackage{caption}
\newcommand{\blind}{1}

\begin{document}

\usetikzlibrary{matrix}
\usetikzlibrary{shapes.geometric} % For the triangle shape

\def\spacingset#1{\renewcommand{\baselinestretch}%
{#1}\small\normalsize} \spacingset{1}

%%%%%%%%%%%%%%%%%%%%%%%%%%%%%%%%%%%%%%%%%%%%%%%%%%%%%%%%%%%%%%%%%%%%%%%%%%%%%%

\if1\blind
{
  \title{\bf Semi-Supervised Generative Learning via Latent Space Distribution Matching}
  \author{Kwong Yu Chong \ 
    and \
   Long Feng \\
    School of Computing \& Data Science, University of Hong Kong}
    \date{}
  \maketitle
} \fi

\if0\blind
{
  \bigskip
  \bigskip
  \bigskip
  \begin{center}
    % {\LARGE\bf Semi-supervised Generative Learning via Autoencoding Wasserstein Generator}
        {\LARGE\bf Semi-Supervised Generative Learning via Latent Space Distribution Matching}
\end{center}
  \medskip
} \fi

\bigskip

\begin{abstract}
We introduce Latent Space Distribution Matching (LSDM), a novel framework for semi‑supervised generative modeling of conditional distributions. LSDM operates in two stages: (i) learning a low‑dimensional latent space from both paired and unpaired data, and (ii) performing joint distribution matching in this space via the 1‑Wasserstein distance, using only paired data. 
This two‑step approach minimizes an upper bound on the 1‑Wasserstein distance between joint distributions, reducing reliance on scarce paired samples while enabling fast one‑step generation. Theoretically, we establish non‑asymptotic error bounds and demonstrate a key benefit of unpaired data: enhanced geometric fidelity in generated outputs. Furthermore, by extending the scope of
its two core steps,
LSDM provides a coherent statistical perspective that connects to a broad
class of latent-space approaches. Notably, Latent Diffusion Models (LDMs) can be viewed as a variant of LSDM, in which joint distribution matching is achieved indirectly via score matching. Consequently, our results also provide theoretical insights into the consistency of LDMs.
Empirical evaluations on real‑world image tasks, including class‑conditional generation and image super‑resolution, demonstrate the effectiveness of LSDM in leveraging unpaired data to enhance generation quality.
\end{abstract}

\noindent%

\vfill

\newpage
\spacingset{1.9} % DON'T change the spacing!

\section{Introduction}
\label{Section:Introduction}
Generative learning has achieved impressive success across a wide range of applications, such as language modeling and image generation. This paper focuses on conditional generative learning in a semi-supervised setting. Semi-supervised learning (SSL) enhances learning by leveraging limited labeled data alongside abundant unlabeled data,  offering particular value in domains where labeled data are scarce or expensive to acquire.

Given inputs $X$, generating outputs $Y$ according to the conditional distribution $\Prob_{Y|X}$ is highly valuable for a wide range of applications. In a fully supervised setting, conditional generation requires paired data $(X,Y)$ in order to learn the conditional mapping between inputs and outputs. However, acquiring authentic paired data can be challenging, and artificially constructing such pairs by assuming a particular generative process may introduce biases. For instance, generative learning has shown impressive success in image super-resolution tasks \citep{ImgSupRes1,ImgSupRes2}. Training these models requires paired datasets containing both low- and high-resolution versions of the same images. However, such pairs are rarely available, as most natural images exist at a single resolution. Moreover, artificially generating low-resolution images via down-sampling can lead to domain shift, potentially diminishing the model’s effectiveness in real-world applications \citep{DomainShift}.

Given these challenges, unpaired data, which are often easier to obtain, hold particular significance. Although such data do not directly reveal the dependence between variables, they can still capture the inherent structure of the response variable $Y$. For instance, when $Y$ has a low-dimensional geometric structure, such as lying close to a manifold, generated samples need to preserve this geometric structure to maintain realism. Therefore, having access to abundant unpaired response data can improve generation quality by enabling better estimation of the underlying data structure. Building on this idea, this paper studies semi-supervised generative learning that draws on additional response data. We now proceed to review existing studies on semi-supervised learning and generative learning.

\subsection{Related Work}
%In this subsection, we review works on semi-supervised learning and generative learning. 

A central theme in SSL is that unlabeled data can enhance estimation even when the underlying model is misspecified. In regression settings, unlabeled covariates have been shown to improve inference for the regression parameter when the mean function is nonlinear \citep{Buja2019}, facilitate more accurate estimation of the response mean \citep{zhang2019semi}, and produce estimators with reduced asymptotic variance \citep{SemiSupLinReg2022}. Recent theoretical developments have further demonstrated minimax optimality and efficiency gains for semi-supervised estimators across a wider range of misspecification scenarios \citep{OptimalSafe2024, GeneralMEstSS2024}. Additional benefits of incorporating unlabeled data have been established in \citet{Cai1, Cai2, Cai3}. In the machine learning literature, SSL has been extensively explored for classification tasks, often relying on structural assumptions such as cluster organization or manifold geometry in the data \citep{ClusterAssump2006, ManifoldAssump2004, lowdenAssump2005}.

%Given the proven value of unlabeled data in regression and classification, it is natural to ask whether semi-supervised methods can also advance generative learning, where the aim is to synthesize rather than predict data. 
Generative learning is arguably one of the most actively researched fields in modern AI. Leading frameworks include Variational Autoencoders (VAEs; \citealp{VAE}), Generative Adversarial Networks (GANs; \citealp{GAN}), and Diffusion Models (DMs; \citealp{DDPM2020}). VAEs model data distributions via variational inference but often produce blurry outputs; GANs generate sharper samples through adversarial training, yet suffer from instability and mode collapse; diffusion models achieve high fidelity through iterative refinement, though at high computational cost.
Recently, generative learning has also drawn growing interest in statistics. \citet{GANError} analyzed the convergence rates of GANs, demonstrating their adaptability to inherently low-dimensional data; \citet{iWGAN} proposed a GAN–VAE hybrid with a probabilistic interpretation; \citet{AdaptiveGAN} extended this model to adaptively select latent dimensions; and \citet{TemperFlow} used Wasserstein gradient flow theory to identify limitations in transport-based models, introducing TemperFlow to address multi-modality.

Compared to unconditional generative learning, sampling from conditional distributions is often more valuable for downstream applications, and research in this area has expanded rapidly. An important line of work in conditional generative learning is based on distribution matching \citep{DCG, WassDCG, WGR}. Leveraging the noise outsourcing lemma, these methods ensure the existence of a conditional generator and estimate it by minimizing a statistical divergence between the joint distribution of generated pairs and the target joint distribution. 
The distribution matching approach has several notable advantages. First, the dual representation of divergences enables single-step training through a GAN-style adversarial game. Second, the generation is fast, requiring only one forward pass through the generator. Third, the objective function directly reflects generation quality, aiding model evaluation. However, this approach does not admit a straightforward mechanism for incorporating unlabeled data into the objective, limiting its applicability in semi-supervised settings.

Another prominent approach is latent‑space generative modeling, which naturally supports semi‑supervised learning by building a low‑dimensional latent representation. Representative examples include Latent Diffusion Models (LDM; \citealp{LDM}), Latent Space Flow Matching (LFM; \citealp{FlowMatchingLatent}), and related variants. %These methods generally adopt a two‑stage scheme: first, an autoencoder is trained on both labeled and unlabeled data to learn a compact latent space; second, with the autoencoder frozen, a conditional diffusion model is trained on labeled data to map inputs to the corresponding latent codes of output.
These methods typically begin by training an autoencoder to capture a compact latent space, then train a conditional generative model to map inputs to the corresponding output latent codes.
Despite their state‑of‑the‑art performance, several open questions remain:
(1) Diffusion and flow‑based models rely on multiple iterative steps for generation, leading to slow generation. (2) The exact mechanism through which unpaired data enhances generation quality is unclear. (3) While existing latent‑space methods exhibit notable structural similarities, there is currently no unified theoretical framework that explains their shared behavior.

\subsection{Contributions}
This paper addresses semi-supervised generative learning by introducing Latent Space Distribution Matching (LSDM), a framework that integrates distribution matching with latent space representation learning. LSDM operates in two stages: first, an autoencoder is trained on paired and unpaired data to learn a compact latent space; second, a latent code generator is trained on paired data to match joint distributions in this space, while the autoencoder remains fixed.
LSDM offers two training variants: composite LSDM (cLSDM), which preserves a composite generator architecture during training, and direct LSDM (dLSDM), which optimizes only the latent generator. These approaches present complementary trade-offs: cLSDM delivers more stable training and higher sample quality, while dLSDM offers faster training and reduced computational cost.

The LSDM framework offers at least three key contributions to conditional generation.
(1) LSDM unifies joint distribution matching and latent space learning within a single objective function. The design naturally supports semi‑supervised learning and enables efficient single‑pass generation. (2) We establish finite‑sample convergence rates that reveal how generation quality in LSDM is jointly influenced by latent smoothness, latent dimension and intrinsic dimension of response data. %, paired and unpaired sample sizes.} 
% \sout{We establish finite‑sample convergence rates that reveal how reconstruction error, latent smoothness, latent dimensionality jointly influence generation quality in LSDM.}
Moreover, we prove that incorporating unpaired response data improves the approximation of the underlying data structure under mild assumptions, thereby enhancing the geometric fidelity of generated samples. (3)  LSDM formalizes a general two‑step paradigm for latent conditional generative learning, encompassing many latent conditional models as special cases. Notably, LDM can be viewed as a variant of dLSDM, where joint distribution matching is optimized indirectly via score matching and the latent generator is defined implicitly through the learned score network. Consequently, our results also provide theoretical insights into the consistency of LDMs.

%Our contributions are summarized as follows:
%\begin{itemize}
 %   \item We propose LSDM, a principled framework that integrates \ckyB{joint distribution matching} with latent space approach, overcoming key limitations of both paradigms. LSDM naturally supports semi-supervised learning, enables fast single-pass generation, and is accompanied by transparent theoretical guarantees.
%    \item  \ckyB{We provide a theoretical analysis of how key variables in two‑step latent‑space models affect generation quality. Our results show that the optimal generation quality is achieved when (i) the autoencoder achieves low reconstruction error, (ii) the latent space is smooth and regular, and (iii) its dimension moderately exceeds the intrinsic dimension of the data. Furthermore, we show that incorporating unpaired response data improves the approximation of the underlying data structure under mild conditions, thereby enhancing the geometric fidelity of the generated samples.}
%    \item We establish connections between LSDM, latent diffusion models, and GAN-based models, showing that latent diffusion models and f‑GANs can be viewed as compatible variants of LSDM.
%\end{itemize}

The remainder of this paper is organized as follows. Section~\ref{Section:Framework} introduces the LSDM framework. Section~\ref{Section:Implementation} details its implementation and connections to pre‑train–fine‑tune approaches. Section~\ref{Section:Connections} discusses links to existing generative models. Section~\ref{Section:TheoreticalAnalysis} presents the theoretical analysis. Section~\ref{Section:Experiments} presents the empirical results on class‑conditional generation and super‑resolution. Section~\ref{Section:Discussion} concludes and outlines future directions.
% The rest of the paper is organized as follows. Section~\ref{Section:Framework} introduces the LSDM framework. Section~\ref{Section:Implementation} details the implementation of LSDM and discusses its connections to pre-train–fine-tune approaches. \ckyB{Section~\ref{Section:Connections} discusses the connections to existing generative models.} Section~\ref{Section:TheoreticalAnalysis} presents the theoretical analysis. Section~\ref{Section:Experiments} provides empirical results on class-conditional generation and super-resolution tasks. Finally, Section~\ref{Section:Discussion} concludes the paper and outlines future directions for LSDM.
\subsection{Notations}
We write $X\sim \Prob_X$ to indicate that a random variable follows distribution $\Prob_X$, with support $\cX$, and denote its empirical distribution by $\hProb_{X}$ . For random variables $X$ and $Y$, $X \eqd  Y$ denotes equality in distribution. For $v \in \R^d$, $\dnorm{v}_2 = \sqrt{\sum_{i=1}^d v_i^2}$ is the $l_2$ norm. We use $x \wedge y:= \min\{x,y\}$ and $x \vee y := \max\{x,y\}$. We say that $a_n \lesssim b_n$, or $a_n = \cO(b_n)$, if there exists $C>0$ such that $a_n \le C \hspace{0.1em}b_n$. We say $a_n \asymp b_n$ if $a_n \lesssim b_n$ and $b_n \lesssim a_n$. We denote the composition of functions $f$ and $g$ by $f\circ g$, where $\left(f\circ g\right)\left(x\right) = f\left(g\left(x\right)\right)$. We use $ a.s.$ to denote almost surely. For a Lipschitz function $f:\cX \subseteq \R^p \to \R^q$, we denote its Lipschitz constant by $\text{Lip}(f) := \sup_{x,y \in \cX, x\neq y}\frac{\dnorm{f(x) - f(y)}_2}{\dnorm{x-y}_2}$.

\section{Semi-supervised Generative Learning}\label{Section:Framework}
Let $(X,Y) \in \cX \times \cY$ be a pair of random vector with joint distribution $\Prob_{X,Y}$, where $\cX \subseteq  \R^p$ and $\cY \subseteq \R^q$ are the supports of $X$ and $Y$, respectively. We denote their marginal distributions by $\Prob_X$ and $\Prob_Y$. Suppose that we are provided with a paired dataset of size $n$ and an unpaired dataset of size $N$, denoted respectively as
\begin{equation}
    \cP = \{X_i,Y_i\}_{i=1}^n \simiid \Prob_{X,Y}, \quad \cU = \{Y_i\}_{i=n+1}^{n+N}\simiid \Prob_Y.
\end{equation}
Our objective is to learn a measurable function $G: \cX\times \R^d \to \cY$ using $\cP\cup\,\cU$, such that for $\Prob_X$-almost every $x$,
\begin{equation}
        G(x, \eta) \sim \Prob_{Y|X=x}, \hspace{0.5em}
\end{equation}
where $\eta \in\mathbb{R}^d$ is an independent noise vector that accounts for the randomness in the conditional distribution. A natural choice for $\eta$ is a Gaussian vector $\eta\sim\mathcal{N}(0,I_d)$.  We focus on settings where $Y$ possesses low‑dimensional intrinsic structure, for instance, when its support concentrates near a manifold. A detailed discussion of this low‑dimensional structure is formalized in Section~\ref{Section:TheoreticalAnalysis}.

Unlike many semi-supervised settings where unlabeled data contains additional predictor $X$, we focus on scenarios with abundant unpaired response $Y$. This choice stems from two main considerations.
First, unpaired responses are often straightforward to obtain. For example, in image super‑resolution, high‑resolution images are much more common than precisely aligned low‑/high‑resolution pairs, since modern cameras capture high‑resolution images by default. Large collections of such images expose the geometric structure of realistic data, and exploiting this structure can enhance the perceptual quality of super‑resolved outputs.
Second, unpaired responses can boost generation quality by enabling better estimation of the low‑dimensional structure underlying $Y$, which is the primary setting considered in this paper.

\subsection{The composite LSDM (cLSDM)}
A key formulation in our approach is to model $G$ as a composite of two functions:
\begin{equation}
    G(x,\eta) = D\circ H(x,\eta).
\end{equation}
Here, $H: \cX \times \R^d \to \cZ \subseteq \R^m$ is the latent code generator, $\cZ$ is the latent space, and $D: \cZ \to \R^q$ is the decoder. This formulation resembles an autoencoder, with a bottleneck latent dimension $m<q$. The rationale for this composite structure is justified by Theorem \ref{Theorem:cLSDM}, whose analysis relies on the 1-Wasserstein distance defined below. %following notion of distributional distance.
\begin{definition}[1-Wasserstein Distance] \label{Definition:WassersteinDistance} The 1-Wasserstein distance between two probability distributions $\Prob_X, \Prob_Y$ on $\R^q$ with finite first moment is
\begin{equation}
    \Wd{X}{Y} = \inf_{\gamma \in \Gamma(\Prob_X, \Prob_Y)} \E_{(X,Y)\sim \gamma}\dnorm{X-Y}_2,
\end{equation}
where $\Gamma(\Prob_X, \Prob_Y)$ is the set of all joint distributions with marginals $\left(\Prob_X, \Prob_Y\right)$.
\end{definition}
% Our theoretical results are stated in terms of the 1-Wasserstein distance , which serves as our primary measure of divergence between probability distributions. 
Definition \ref{Definition:WassersteinDistance} presents the 1-Wasserstein distance, a metric used to measure the space of Borel probability distributions with finite first moment.
Other forms of Wasserstein distance, such as the 2-Wasserstein distance, have also been explored in the literature across a range of contexts, including data clustering \citep{W2_clustering} and domain adaptation \citep{ W2_domainadaptation}. We restrict our attention to the 1-Wasserstein distance, as it plays a crucial role in Theorem \ref{Theorem:cLSDM} presented below:

\begin{theorem}[Risk Decomposition for the Composite Generator]\label{Theorem:cLSDM} Let $E: \cY \to \cZ$ be an encoder. Suppose the generator $G$ has the form $G = D\circ H$, where $D: \cZ \to \R^q$ and $H: \cX \times \R^d \to \cZ$. Then, the 1-Wasserstein distance between the joint distributions of  $\left(X, G(X,\eta)\right)$ and $(X,Y)$ satisfies the bound:
\begin{equation}
\Wd{X, G(X, \eta)}{X,Y} \leq \E\dnorm{Y - D\circ E(Y)}_2 + \Wd{X, D\circ H(X, \eta)}{X, D\circ E(Y)},
\end{equation}
Consequently, if $(D,E,H)$ is a triplet that satisfies
\begin{equation}
    \E\dnorm{Y - D\circ E(Y)}_2=0,\quad 
    \Wd{X, D\circ H(X, \eta)}{X, D\circ E(Y)}=0,
\end{equation}
then the generator achieves conditional distribution matching:
\begin{align*}
    G(x,\eta) \sim \Prob_{Y|X=x}.
\end{align*}
\end{theorem}

Motivated by the theorem, we propose a two‑step procedure for learning the composite generator $G$. In the first step, we learn an decoder-encoder pair $(D,E)$ by reconstructing $Y$ from the combined response data. When the latent dimension \textit{m} is substantially smaller than the input dimension, this step effectively performs representation learning, capturing the intrinsic structure of the data and inducing a compact, low-dimensional latent space. In the second step, using the learned pair $(\widehat{D},\widehat{E})$, we estimate $\hH$ by minimizing the 1-Wasserstein distance between the joint distributions of $(X,\widehat{D}\circ H(X,\eta)) $ and $(X, \widehat{D}\circ \widehat{E}(Y))$, based on the paired data. Due to the composite structure $\widehat{G}=\hD\circ \hH$, the generated samples are constrained to the geometric structure learned during the first-stage representation step.

Given estimated pair $(\hD, \hE)$, let $\widehat{\Prob}_{X, \hD\circ H (X, \eta)} $ and $\widehat{\Prob}_{X, \hD\circ \hE(Y)}$ denote
the empirical distributions of samples $\bigl\{X_i, \hD\circ H\left(X_i, \eta_i\right)\bigr\}_{i=1}^n$ and $\bigl\{X_i, \hD\circ \hE \left(Y_i\right)\bigr\}_{i=1}^n$, respectively. Our framework is summarized below.
\begin{algorithm}[H]
\caption{composite LSDM (cLSDM) \label{Algo:cLSDM}}
\begin{algorithmic}[1]
\Require $\{X_i,Y_i\}_{i=1}^n$, $\{Y_i\}_{i=n+1}^{n+N}$
\State Representation learning with combined data $\cP\cup \cU$:
\begin{equation}
    (\hD, \hE) = \argmin_{D,E} \ \frac{1}{n+N}\sum_{i=1}^{n+N}\dnorm{Y_i - D\circ E(Y_i)}_2 \label{eq:obj1}
    \end{equation}
\State Distribution matching with paired data $\cP$:
 \begin{equation}
            \hH = \ \argmin_{H} \; W_1\left(\widehat{\Prob}_{X, \hD\circ H (X, \eta)}, \widehat{\Prob}_{X, \hD\circ \hE(Y)}\right) \label{Equation:cLSDMDistributionMatchingObjective}
            \end{equation}
\end{algorithmic}
\end{algorithm}

The paired and unpaired data play distinct roles in this two-step approach: paired observations encode the conditional mapping between $X$ and $Y$, while unpaired observations capture the intrinsic structure of the response $Y$. For example, consider a setting where $Y$ represents facial images and $X$ corresponds to attributes. In Step 1, we learn a latent representation whose dimensions may correspond to semantically meaningful features such as gender or hairstyle. In Step 2, the model maps attributes to these latent codes, which are then decoded into realistic images. A key advantage of this framework is that even when the conditional mapping from the predictor space $\cX$ to the latent space $\cZ$ is imperfect, the decoder, which is trained on abundant response data, can still generate realistic outputs by respecting the underlying geometric structure of $\cY$. 

We note that a seemingly natural alternative in the conditional generation step would be to directly minimize $W_1\left(\widehat{\Prob}_{X, \hD\circ H (X, \eta)}, \widehat{\Prob}_{X, Y}\right) $ while keeping $\hD$ fixed. However, this approach is fundamentally limited: the range of the decoder $\hD$ may not cover the full support of $Y$, which can cause two difficulties. 
First, in adversarial training (Section \ref{Subsection:DualityImplementation}), the critic can easily discriminate between two distributions when their supports are different, resulting in unstable training. Second, the objective cannot be minimized to zero, no matter how flexible $H$ is. In contrast, when both distributions have the same support as in (\ref{Equation:cLSDMDistributionMatchingObjective}), the first difficulty disappears. Moreover, as the following proposition shows, a sufficiently flexible $H$ can drive the objective (\ref{Equation:cLSDMDistributionMatchingObjective}) to zero.
\begin{proposition}
[Existence of $H$ for Arbitrary $\hD,\hE$]\label{Proposition:ExistenceOfH} For arbitrary $(\hD,\hE)$, there exists a measurable function $H: \cX \times \R^d \to \cZ$ such that $(X,  H(X,\eta)) = (X,  \hE(Y)) \hspace{0.5em} a.s.$ and $(X, \hD\circ H(X,\eta)) = (X, \hD\circ \hE(Y)) \hspace{0.5em} a.s.$ If they have finite first moment, then both $ \Wd{X, \hD\circ H(X, \eta)}{X,\hD\circ \hE(Y)} $ and $ \Wd{X,  H(X, \eta)}{X,\hE(Y)} $ are zero.
\end{proposition}

\FloatBarrier
\begin{figure}[ht]
\centering
\includegraphics[max width=\linewidth]{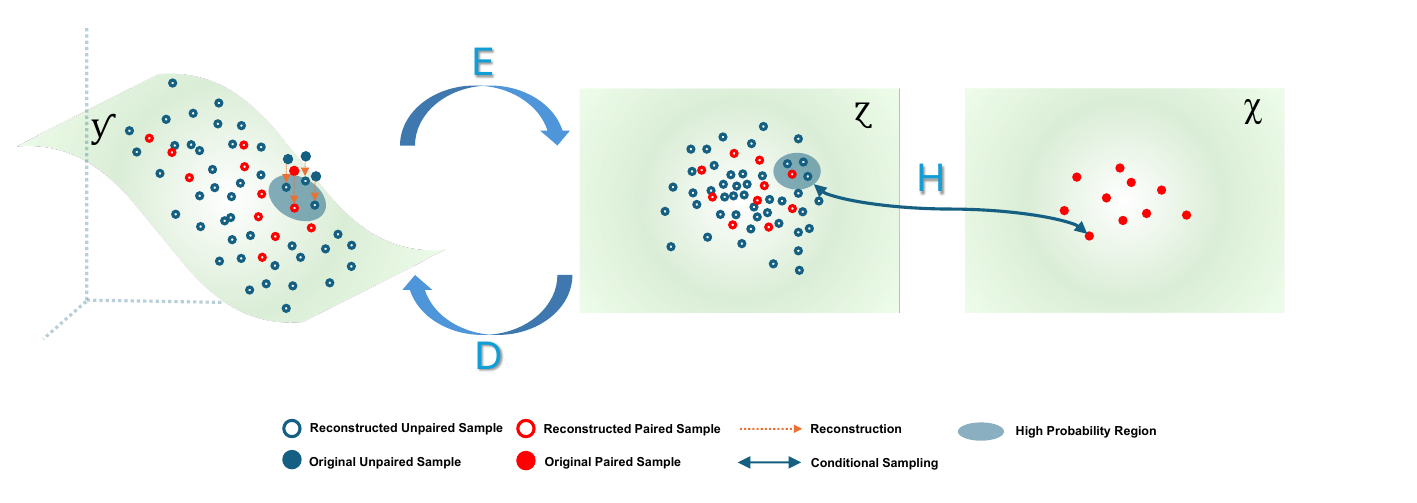}
\caption{An Illustrative Figure of LSDM.}\label{Figure:LSDM}
\end{figure}
\FloatBarrier

We name this framework Latent Space Distribution Matching (LSDM). An illustration is provided in Figure \ref{Figure:LSDM}. Framework \ref{Algo:cLSDM} is referred to as composite LSDM (cLSDM), because the composite generator structure is preserved during distribution matching. In the next subsection, we introduce a variant of LSDM called direct LSDM (dLSDM), which employs a simpler joint distribution matching objective in Step 2.

\subsection{The direct LSDM (dLSDM)}
The direct LSDM is motivated by the following theorem:
\begin{theorem}\label{Theorem:dLSDM}
    If the decoder $D$ is $K$-Lipschitz continuous, then
    \begin{equation}
        \Wd{X, D\circ H(X,\eta)}{X, D\circ E(Y)} \leq \left(1 \vee K\right) \hspace{0.25em}\Wd{X, H(X,\eta)}{X,E(Y)}.
    \end{equation}
Consequently, if $(D,E,H)$ is a triplet that satisfies
\begin{equation}
    \E\dnorm{Y - D\circ E(Y)}_2=0,\quad 
    \Wd{X, H(X, \eta)}{X, E(Y)}=0,
\end{equation}
then the generator achieves conditional distribution matching:
\begin{align*}
 G(x,\eta) \sim \Prob_{Y|X=x}.
\end{align*}
\end{theorem}
When the decoder $D$ is Lipschitz continuous (e.g., when implemented as a ReLU neural network), Theorem \ref{Theorem:dLSDM} decomposes the distribution matching objective into two components: a reconstruction gap and a latent distribution matching error. The latent generator $H$ can then be learned by matching the joint distributions of $(X, H(X,\eta)) $ and $(X, \hE(Y))$. The existence of $H$ is similarly guaranteed by Proposition \ref{Proposition:ExistenceOfH}. This decomposition yields the following framework.
\begin{algorithm}[H]
\caption{direct LSDM (dLSDM) \label{Algo:dLSDM}}
\begin{algorithmic}[1]
\Require $\{X_i,Y_i\}_{i=1}^n$, $\{Y_i\}_{i=n+1}^{n+N}$
\State Representation learning with combined data $\cP\cup \cU$ as in (\ref{eq:obj1}).
\State Latent distribution matching with paired data $\cP$: 
 \begin{equation}
            \hH = \argmin_{H} W_1\left(\widehat{\Prob}_{X,  H (X, \eta)}, \widehat{\Prob}_{X,  \hE(Y)}\right)\label{Equation:dLSDMDistributionMatchingObjective}
            \end{equation}
\end{algorithmic}
\end{algorithm}

Under dLSDM, distribution matching is explicitly performed in the latent space $\cZ$. The latent generator $\hH$ is trained to generate samples from the conditional distribution  $\Prob_{\hE(Y)|X=x}$, with the encoder $\hE$ learned from the first step. Under the ideal case that the learned pair $(\hD,\hE)$ can perfectly reconstruct $Y$ (i.e., $\hD\circ \hE(Y)=Y$),  learning the conditional distribution $\Prob_{Y|X=x}$ is equivalent to learning $\Prob_{\hE(Y)|X=x}$.

\subsection{Choosing Between dLSDM and cLSDM}

The cLSDM and dLSDM offer two complementary approaches to semi-supervised generative learning, each with distinct advantages.

Under mild regularity assumptions, both cLSDM and dLSDM attain the same statistical convergence rate as demonstrated in Section~\ref{Section:TheoreticalAnalysis}. In practice, however, the interaction between the optimization algorithm, adversarial training dynamics, and network architecture produces distinct trade‑offs. When a GAN‑like adversarial training is adopted, cLSDM typically yields more stable training and higher sample quality, whereas dLSDM enables faster training with lower computational cost. Therefore, cLSDM is preferable when training stability and output quality are important; dLSDM is better suited when computational resources are limited or training speed is a priority.

In cLSDM, both the “real” and “generated” samples are outputs of the same decoder. Consequently, they share the same data support and exhibit similar structure. This makes it harder for the critic to distinguish between the two distributions, leading to a more balanced adversarial game. In dLSDM, the critic compares the latent generator’s output $H(X,\eta)$ with the learned latent code $\hE(Y)$, which may initially lie on different supports or possess dissimilar geometries. If the latent generator is insufficiently expressive or the critic becomes too powerful early in training, the critic can easily separate the two, resulting in unstable training. Since a balanced critic‑generator game is critical for adversarial optimization, cLSDM generally delivers more stable training and, as a result, higher sample quality.

The dLSDM avoids forward passes through the decoder $\hD$ during the distribution matching step. When the decoder is a large neural network, decoding the latent codes in each iteration of training adds substantial computational overhead. By omitting this step, dLSDM reduces training time and memory requirements. Thus, dLSDM generally offers faster training with lower computational cost. In our experiments, dLSDM trained on average twice as fast as cLSDM.

The two-step approach of LSDM is closely related to the pre-training and fine-tuning paradigm. We elaborate on this perspective and detail the implementation of LSDM in the next section.

\section{Implementation of LSDM} \label{Section:Implementation}
\subsection{Pre-training and Fine-tuning}
Pre-training and fine-tuning have become widely used strategies in various machine learning tasks, particularly in the development of language models and image generation, due to their ability to leverage large-scale unlabeled data and improve generalization across downstream tasks. The two-step approach of LSDM is closely related to the generative pre-training and fine-tuning paradigm.

During the pre-training stage, models are trained on unlabeled data to learn meaningful feature representations. In the subsequent fine-tuning stage, some components of the model are frozen while the remaining parts are trained on labeled data to adapt the model parameters for a specific task. Alternatively, the model can be extended with an adapter component, such as additional layers or parameters, which is trained during fine-tuning with the core model remaining frozen. For a more detailed discussion, see \cite{GenerativePretraining2010}, which offers a regularization-based perspective on the benefits of integrating unsupervised pre-training into supervised learning tasks.

LSDM adapts this pre-training and fine-tuning framework specifically to conditional generation. In Step 1 (pre-training), we train an undercomplete autoencoder 
 on the combined dataset $\cP \cup \, \cU$ 
 to learn efficient representations of the response data. The is done by learning to reconstruct response with a bottleneck dimension. Step 2 serves as the fine-tuning (or, alignment) stage, where the generator’s adapter component, $H$, is trained on paired data for conditional generation, while the autoencoder remains frozen.

LSDM can thus be viewed as a pre‑train–fine‑tune extension of the conditional Wasserstein GANs. It fuses latent space learning with 1-Wasserstein distribution matching via Theorem \ref{Theorem:cLSDM} and \ref{Theorem:dLSDM}. This integration yields several key advantages: it naturally supports semi‑supervised learning, enables fast single‑pass generation, and provides a set of transparent theoretical guarantees.

\subsection{Implementation and Regularization}
\label{Subsection:DualityImplementation}
The pre-training and fine-tuning framework underlying LSDM offers substantial flexibility, enabling the integration of diverse autoencoders in the Step 1 and a variety of generative models in Step 2.

In pre-training stage of LSDM, different variants of the autoencoder can be applied. For instance, we may consider a regularized version of the autoencoder, where the objective function takes the form
\begin{equation}
  \E \dnorm{Y - D\circ E(Y)}_2  + R(E, D),
\end{equation}
where $R(E,D)\ge 0$ is a non‑negative regularization term on the deterministic autoencoder. This objective provides an upper bound on the reconstruction error in (\ref{eq:obj1}). 
The regularization term $R(E,D)$ can be designed to control the variance and smoothness of the latent space, which directly impacts the subsequent distribution matching step. %\sout{Theoretically, our results rely on the continuity of the learned decoder. } 
Empirical experiments suggest that proper regularization improves generation quality.

One widely adopted approach is distributional regularization, exemplified by Wasserstein autoencoders (WAE, \citealp{WAE}). In this formulation,
\begin{equation}
    R(E, D)= \lambda \hspace{0.1em} W_1\left(\Prob_{E(Y)}, \Prob_{Z}\right),
\end{equation}
where $\Prob_Z$ is a prescribed prior (e.g., standard Gaussian). By encouraging the latent codes to follow a continuous distribution, one promotes a smooth and regular latent space, which can benefit the subsequent distribution‑matching step. 

In LSDM, optimization of the distribution matching step is also highly flexible. A standard and effective approach for minimizing the objectives (\ref{Equation:cLSDMDistributionMatchingObjective}) and (\ref{Equation:dLSDMDistributionMatchingObjective}) is to apply the Kantorovich–Rubinstein duality of the 1‑Wasserstein distance \citep{villani2008}:
\begin{equation} 
W_1(\Prob_X,\Prob_Y) = \sup_{f\in \mathcal{F}^1}\E_{x\sim\Prob_X} f(x) - \E_{y\sim \Prob_Y}f(y),
\end{equation}
where $\mathcal{F}^1  = \Lip(\Omega, 1)$ denotes the class of 1-Lipschitz functions. The function $f$, called the critic, can be parameterized by a deep neural networks. This duality transforms the minimization into a min‑max problem similar to that in GANs. Such duality‑based optimization is well‑studied, with non‑asymptotic error bounds that account for approximation errors in both the critic and the generator  \citep{WGR}. We adopt this GAN‑based formulation as the default optimization strategy for LSDM. Implementation details are provided in Section D of the supplementary material. 

Beyond this formulation, the LSDM framework is compatible with a broader class of conditional generative models. As we establish in the next section, this includes not only various f‑GAN objectives but also latent diffusion models.

\section{Connections with LDMs and f-GANs}\label{Section:Connections}

\subsection{Connection with Latent Diffusion Models}
Latent diffusion models (LDMs) are closely related to dLSDM. Rather than performing explicit distribution matching, LDMs generate latent codes using a conditional diffusion model. We outline the connection for latent generation here; a more detailed comparison, including autoencoder training, is provided in the supplementary material.
% Latent diffusion models (LDMs) are closely related to dLSDM. Instead of performing explicit distribution matching, LDMs generate latent codes using a conditional diffusion model. We outline the connection here; a more detailed comparison is provided in the supplementary material.

We start with an overview of the conditional diffusion framework. In LDMs that use a deterministic autoencoder with vector-quantization regularization \citep{VQGAN}, the generation target is the continuous, pre‑quantised latent code $Z= E(Y)$. Following the setup in \citet{OverviewDM}, we consider a forward diffusion process which adds noise to $Z$ according to
\begin{equation}
dZ_t = -\frac{1}{2}Z_tdt + dW_t,\qquad Z_0 \sim \Prob_{Z \mid X=x},
\end{equation}
where $W_t$ is a Wiener process. The process is run until a sufficiently large time $T$, at which $Z_T $ is approximately standard Gaussian. Let $\Prob_ {Z_t|X=x}$ 
denote the marginal distribution of $Z_t $ conditional on $X=x$, and let its score function be written as
$\nabla\log p_{t}(z|x)$. To sample from $\Prob_{Z|X=x}$, noise is removed via the backward process
\begin{equation}
    d\overleftarrow{Z_t} = \left\{\frac{1}{2}\overleftarrow{Z_t} + \nabla \log p_{T-t}\left(\overleftarrow{Z_t}|x\right)\right\}dt + d\overline{W}_t, \quad \overleftarrow{Z_0} \sim \cN(0, I_m),
\end{equation}
where $\overline{W}_t,$ is a time-reversed Wiener process. The true conditional score is unknown and is approximated by a conditional score network $s(z,x,t)\approx \nabla\log p_{t}(z|x)$. A standard training objective is the score‑matching loss, defined as
\begin{equation}
    \cL_{SM}(s) =  \E_{t \sim U([0,T])}\E_{x\sim \Prob_X}\E_{z_t\sim \Prob_{Z_t|X=x}} \dnorm{s(z_t,x, t) - \nabla\log p_t(z_t|x)}_2^2 ,
\end{equation}
An equivalent denoising formulation is frequently adopted instead (see Section 5.1 of \citet{OverviewDM}). However, for clarity, we maintain the original form here.

For any conditional score network $s(z,x,t)$, the latent code generator $H_s$ is defined implicitly by the backward process and the score network. A notable special case occurs when the reverse SDE is discretized using the Euler–Maruyama scheme \citep{Euler-Maruyama}, in which $H_s$ has an explicit form. Suppose the backward process is simulated over $N$ steps on the interval $[0,T]$, with step size $\Delta=T/N$ and discrete times $t_n = n\Delta$ for $n=0,1,\dots, N-1$. Define the update as $f_t^{s}(z,x,\eta) := z + \left(\frac{1}{2}z +  s(z,x,T-t) \right) \Delta + \eta \sqrt{\Delta }$. The generator $H_s$ is then given by
\begin{equation}
    H_s(x,\eta) = f^s_{t_{N-1}}(\dots (f^s_{t_1}( f^s_{t_0}(\eta_0, x,\eta_1), x, \eta_2)\dots ),
\end{equation}
where $\eta = (\eta_0, \dots , \eta_{N})$ and $\eta_i\sim \cN(0, I_{m})$. 

We let $\Prob_{X,H_s}$ denotes the distribution obtained by running the continuous-time backward process with a given score network $s(z,x,t)$ and approximating $\Prob_{Z_T|X=x}$ by a standard Gaussian. The following proposition extends the results of  \citet{ConditionalDiffusionConvergence} to show that the score matching objective upper‑bounds the dLSDM Step 2 objective.
\begin{proposition}[Bound on the 1‑Wasserstein Distance by Score Matching Objective] \label{Proposition:ConnectionsWithLDM}Suppose there exist constants $C_1,C_2, C_3,C_4>0$ such that 
(1) $\mathbb{D}_{KL}(\Prob_{Z|X=x}, \cN(0, I_m)) < C_1$ for all $x \in \cX$,
(2) $\E_{Z_t|X=x}\dnorm{s(z_t,x,t) - \nabla \log p_t(z_t|x) }_2^2<C_2$ for all $t\in [0,T]$ and $x \in \cX$,
(3) $\E _{v \sim Z|X=x}\dnorm{v}_2^2<C_3$ and $\E _{v \sim H_s|X=x}\dnorm{v}_2^2<C_4$ for all $x \in \cX$. 
Then
$$ W_1(\Prob_{X,H_s}, \Prob_{X,Z})\lesssim  \left(T \cL_{SM}(s)\right)^{1/4} + e^{-T/2}.$$
\end{proposition}
The proposition implies that if a score model is consistent
and $T \to \infty$ at a proper rate, then the model achieves joint distribution matching consistency under regularity conditions on the conditional density and score network. This implies that a diffusion model can be used to generate latent codes within the dLSDM framework, and LDMs can be viewed as an instance of dLSDM in which the Step 2 objective is minimized implicitly via score matching.

\subsection{Connection with f-GANs}
The 1‑Wasserstein distance in the distribution matching step of LSDM may be substituted with various $ f$-divergences. This substitution is motivated by a general inequality that bounds the 1‑Wasserstein distance in terms of commonly used 
$f$-divergences, as formalized in the following proposition.
\begin{proposition}[Bound on the 1‑Wasserstein Distance by f‑divergences]\label{Proposition:ConnectionsWithfGAN}
     Let $\mathbb{D}_f\left(\mu, \nu\right)$ denote one of the following divergences between two probability measures $\mu$ and $\nu$ with bounded support $\Omega$: (i) Kullback–Leibler, (ii) $\chi^2$, (iii) Jensen-Shannon, or (iv) total variation (formal definitions are provided in the supplementary material). Then
    \begin{equation}
    W_1\left(\mu, \nu\right)\leq 2\, \text{diam}(\Omega) \max\biggl\{\mathbb{D}_f\left(\mu, \nu\right), \sqrt{\frac{1}{2}\mathbb{D}_f\left(\mu, \nu\right)}\biggr\}.
\end{equation}
where $\text{diam}(\Omega)=\sup_{x,y \in \Omega}\dnorm{x-y}_2$ denotes the diameter of $\Omega$.
\end{proposition}

The $f$-divergences listed above are those used in classical 
$f$-GANs. When the data distribution has bounded support, the 1‑Wasserstein objectives in (\ref{Equation:cLSDMDistributionMatchingObjective}) and (\ref{Equation:dLSDMDistributionMatchingObjective}) can be replaced by any of these divergences, enabling LSDM to be interpreted as a latent space extension of the classical f-GAN \citep{DCG, fGAN}. Optimizing these divergences through their variational dual formulations leads to a GAN‑style adversarial game.

While the substitution with f-divergences is theoretically admissible, the 1‑Wasserstein objective is generally preferred in practice. The duality‑based minimization of the Wasserstein distance enforces a Lipschitz constraint on the critic, which typically leads to more stable training dynamics than those of f‑divergence objectives.

\section{Theoretical Analysis}\label{Section:TheoreticalAnalysis}
In this section, we present theoretical analyses of the LSDM framework, including error bounds on reconstruction, distribution matching, and the benefits of unpaired data. We first introduce ReLU neural networks, as our decoder $D$, encoder $E$ and latent generator $H$ will be modeled by this class of function:
\begin{definition}[ReLU Neural Networks] Let $L \in \mathbb{N}$ and $N_i \in \mathbb{N}, \forall i=0,...,L$. A $L$-layer ReLU neural network $f: \R^{N_0} \to \R^{N_L}$ is defined as
\begin{equation}
f(x) = T_L \circ \sigma \circ T_{L-1} \circ \dots \circ \sigma \circ T_1 (x),
\end{equation}
where $\sigma(x) = \max \{x,0\}$ is applied component-wise, $T_i(z) = W_i z + b_i$ with $W_i \in \R^{N_{i} \times N_{i-1}}$ and $b_i \in \R^{N_{i}}$. We define the depth of the network as $\cL(f) : = L$, the number of parameters as $\cS(f)=\sum_{i=1}^L (N_iN_{i-1} +N_i)$, and the class of ReLU neural networks as:
\begin{align}
    \cR \cN(N_0, N_{L}, L, S, B) = \{ & f: \cL(f) \le L , \, \cS(f) \le S, \sup_{x \in \R^{N_0}}\|f(x)\|_\infty \le B  \}.
\end{align}
\end{definition}
We denote ReLU network classes for the decoder $D$, encoder $E$, and latent generator $H$ as $\cD=\cR\cN(m, q,L_d, S_d, B_d)$, $\cE=\cR\cN(q,m,L_e,S_e,B_e)$, and $\cH=\cR\cN(p+d,m,L_h,S_h,B_h)$, respectively. 
Here, $m$ denotes the latent dimension of the autoencoder pair, typically smaller than the ambient dimension $q$. As will be shown later, $m$ has a significant impact on model performance. Another function class relevant to our analysis is the Hölder class, whose definition is given below:
\begin{definition}[Hölder Class] For any  $\beta=s+r, s\in \mathbb{N}_0, r \in (0,1]$, $\cX \subseteq \R^d, \cY \subseteq \R$, the Hölder class $\cH^{\beta}(\cX, \cY, C)$  is defined as $\cH^{\beta}(\cX, \cY, C) := \bigl\{f: \cX \to \cY: \dnorm{f}_{\cH^\beta} \leq C \bigr\}$, where
$$\dnorm{f}_{\cH^\beta} := \sum_{|\alpha|\leq s} \dnorm{\partial^{\alpha}f}_\infty + \sum_{|\alpha| = s}\sup_{x\neq y} \frac{|\partial^\alpha f(x) - \partial^\alpha f(y)|}{\dnorm{x-y}_2^r}$$ $\alpha = (\alpha_1, ..., \alpha_d)$ is a multi-index, $|\alpha| = \sum_{i=1}^d \alpha_i$, $\partial^{\alpha}f = \frac{\partial^{|\alpha|}f}{\partial x_1^{\alpha_1}\dots \partial x_d^{\alpha_d}}$. For vector-valued functions $f: \cX \to \R^q$, let $\dnorm{f}_{\cH^{\beta}} = \sum_{j=1}^q \dnorm{f_j}_{\cH^{\beta}}$, where $f_j$ denotes the jth component of $f$.
\end{definition}
The Hölder class generalizes the notion of Lipschitz continuous function. When $\beta=1$, it reduces to the class of Lipschitz functions with Lipschitz constant $C$. 
This function class is widely studied in the theoretical analysis of neural networks, as universal approximation theorems have shown that ReLU-activated networks can approximate any Hölder-class function arbitrarily well, with the approximation error depending on the network depth and width. Throughout our analysis, we impose the following condition:
\begin{assumption}[Boundedness Condition]\label{Assumption:Boundedness} Assume that $\cX, \cY$ and $\cZ $ are bounded. Without loss of generality, assume $\cX \times \cY \subseteq [0,1]^{p+q}$ and $\cZ \subseteq [0,1]^m$.
\end{assumption}
Assumption \ref{Assumption:Boundedness} is naturally satisfied by many types of data, such as images and videos. Without loss of generality, we set the upper bound constants $B_d$, $B_e$ and $B_h$ to 1. 

The low-dimensional structure of $Y$ can be characterized through various intrincis dimension of its support $\cY$. Here, we consider the Minkowski dimension and denote it by \( d_{\mathcal{Y}} \).
When the response data lies on a low-dimensional set, we have $d_\cY < q$. A formal definition of the Minkowski dimension is provided in the supplementary material.

\subsection{Reconstruction error analysis}
We begin by analyzing the reconstruction error in Step 1. The empirical reconstruction loss in Step 1 is defined as
$\widehat{\cL}_{\text{recon}}(D,E) := \frac{1}{n+N} \sum_{i=1}^{n+N} \dnorm{Y_i - D\circ E(Y_i)}_2$, and its population counterpart is given by:
\begin{equation}
    \cL_{\text{recon}}(D,E) := \E_Y \dnorm{Y - D\circ E(Y)}_2.
\end{equation}
We impose the following realizability assumption.
\begin{assumption}[Realizability]\label{Assumption:Existence} There exists $D^* \in \cH^{\beta_d}([0,1]^{d_\cY}, [0,1]^q, C)$ and $E^* \in \cH^{\beta_e}([0,1]^q, [0,1]^{d_\cY}, C)$ such that $\cL_{\text{recon}}(D^*,E^*) = 0$, for some $\beta_d,\beta_e,C>0$, and $d_\cY<q$.
\end{assumption}
Assumption~\ref{Assumption:Existence} posits that $Y$ is intrinsically low-dimensional and admits a perfect reconstruction via continuous maps. While this perfect reconstruction is idealized for an autoencoder with a latent dimension smaller than the ambient dimension $q$, the optimal reconstruction error in practice is typically small, suggesting that our theoretical results remain relevant in practice. This assumption allows us to analyze the influence of the latent dimension $m$ and autoencoder smoothness $\beta_d,\beta_e$ on the statistical convergence rate.
An important consequence of the assumption is that, for any $m \geq d_\cY$, one can construct $\tD \in \cH^{\beta_d}([0,1]^{m}, [0,1]^q, C)$ and $\tE \in \cH^{\beta_e}([0,1]^q, [0,1]^{m}, C)$ satisfying $\cL_{\text{recon}}(\tD,\tE) = 0$, by ignoring the redundant dimensions $m-d_\cY$. In other words, under the assumption, the Step 1 loss can be driven to zero whenever the latent dimension $m$ is at least the intrinsic dimension $d_\cY$. For our analysis we consider the empirical risk minimizers:
\begin{align} \label{Equation:ReconstructionERM}
   (\hD, \hE) \in \argmin_{D \in \cD, E \in \cE} \  \widehat{\cL}_{\text{recon}}(D,E).
\end{align}
With these definitions we obtain the following non‑asymptotic error bound:
\begin{theorem}[Reconstruction Error Bound]\label{Theorem:ReconstructionError}  Suppose Assumptions \ref{Assumption:Boundedness} and \ref{Assumption:Existence} hold. There exists a constant $\alpha>0$ such that if the neural network classes $\cD, \cE$ are chosen with latent dimension $m\geq d_\cY$ and parameters $L_e, L_d= \alpha \log (n+N)$, $S_d = \alpha (n+N)^{\frac{m}{2\beta_d+m}}\log (n+N)$, and  $S_e = \alpha (n+N)^\frac{s}{2\beta_e(\beta_d\wedge1)+s}\log (n+N)$, then
\begin{align*}
        \E_{\cP \, \cup\,\cU } \cL_{\text{recon}}(\widehat{D}, \widehat{E})\lesssim  (n+N)^{-1/(2+\kappa)}\log^2(n+N), 
    \end{align*}
where $\kappa = \max\left\{\frac{m}{\beta_d}, \frac{s}{\beta_e(\beta_d\wedge1)} \right\}$ and $s > d_\cY$ is arbitrary.
\end{theorem}
Theorem \ref{Theorem:ReconstructionError} establishes a non‑asymptotic bound on the reconstruction loss, expressed in terms of the latent dimension $m$ and the smoothness of the optimal autoencoder. Greater smoothness of the target autoencoder, reflected in larger values of $\beta_d,\beta_e$, leads to faster statistical convergence.
 
\begin{corollary}
\label{Corollary:FullReconstruction}  Suppose $\beta_d,\beta_e=1$. Under the conditions of Theorem \ref{Theorem:ReconstructionError}, for any $s>d_\cY$,
\begin{align*}
        \E_{\cP \, \cup\,\cU } \cL_{\text{recon}}(\widehat{D}, \widehat{E})\lesssim  (n+N)^{-1/\left(2+m\vee s\right)} \log^2(n+N). 
    \end{align*}
\end{corollary}
Corollary \ref{Corollary:FullReconstruction} suggests that, when the target autoencoder is Lipschitz, one should choose $m = d_\cY$. Choosing $m < d_\cY$ loses the consistency guarantee, while choosing $m>d_\cY$ leads to a slower statistical convergence rate. 

\begin{remark}%[Estimating the intrinsic dimension $d_\cY$] 
Estimation of the intrinsic dimension $d_\cY$ has been studied in the literature. One practical approach is to train GANs with varying latent dimensions and select the smallest dimension that yields good performance. Notably, the latent Wasserstein GAN \citep{AdaptiveGAN} incorporates a rank penalty that consistently learns the appropriate latent dimension. The resulting dimension can be taken as an estimate of $d_\cY$ and used to set $m$ when training the autoencoder.
\end{remark}

\subsection{Distribution matching error analysis}\label{Subsection:DistributionMatching}
Now we present the distribution matching error analysis in Step 2. 
We focus on the following empirical risk optimizers:
\begin{align}
   &\hH_{\text{cLSDM}} \in \argmin_{H \in \cH}  W_1\left(\hProb_{X, \hD \circ H(X,\eta)}, \hProb_{X, \hD \circ \hE(Y)} \right),\\
   &\hH_{\text{dLSDM}} \in \argmin_{H \in \cH}  W_1\left(\hProb_{X, H(X,\eta)}, \hProb_{X,  \hE(Y)} \right).
\end{align}
We assume access to exact empirical risk minimizers for the 1‑Wasserstein distance. 
In practice, when a dual formulation is employed, the critic is parameterized by a neural network, which introduces an approximation gap between the true distance and its estimate. 
Moreover, optimization via stochastic gradient descent may converge to a local rather than a global minimizer. 
Although these gaps are not explicitly addressed in our theorems, they can be incorporated into the analysis by quantifying the critic’s approximation error, as in~\cite{WGR}, and by adding an explicit optimization error term.

% The distribution matching in Step 2 relies on the continuity of the learned decoder $\hD$. For a well‑behaved mapping from the predictor space $\cX$ to the latent space $\cZ$ to yield an equally good mapping from $\cX$ to the response space $\cY$,  the decoder must be sufficiently regular. Although $\hD$ and $\hE$ are implemented as Lipschitz-continuous ReLU networks, their Lipschitz constants can in principle grow with network complexity and the combined sample size $n+N$. To enable a tractable analysis, we adopt the following continuity assumption:

Before presenting the non‑asymptotic error bound for distribution matching under both dLSDM and cLSDM, we impose an additional smoothness condition to ensure that any sufficiently good encoder–decoder pair is regular.

\begin{assumption} [Smoothness Condition]\label{Assumption:Smoothness} There exists constants $\delta,K>0$ such that for any $(D,E) \in \cD \times \cE$ with $\cL_{\text{recon}}(D,E)<\delta$, we have $\operatorname{Lip}(D) < K$ and $\operatorname{Lip(E)} < K$. Moreover, there exists $H^* \in \cH^{\beta_h}([0,1]^{p+d}, [0,1]^m, C)$ such that $(X,  H^*(X,\eta)) = (X, E(Y)) \hspace{0.5em} a.s.$
\end{assumption}

Assumption \ref{Assumption:Smoothness} ensures that any sufficiently good ReLU autoencoder obtained from Step 1 is regular, with a Lipschitz constant not exceeding $K$, and that the corresponding target latent generator $H^*$ (whose existence is established in Proposition \ref{Proposition:ExistenceOfH})  is Hölder smooth with exponent $\beta_h$. The assumption does not automatically follow from standard universal approximation results for ReLU networks, since the Lipschitz constant of the decoder can grow with the size of the network.
In practice, one may use the regularity term $R(D, E)$ introduced in Section~\ref{Section:Implementation} to enforce the smoothness of the autoencoder, for example, by constraining the spectral norms of the weight matrices.

\begin{theorem}[Distribution Matching Consistency]\label{Theorem:DistributionMatchingConsistency}  Suppose Assumptions \ref{Assumption:Boundedness}, \ref{Assumption:Existence} and \ref{Assumption:Smoothness} hold. There exists a constant $\alpha>0$ such that if  (1)  $n \asymp N$, (2) the network classes $\cH$ are set with  $L_h= \alpha \log n$,  $S_h = \alpha n^{\frac{m}{2\beta_h+m}}\log n$ (3) the network classes $\cD,\cE$ are set as in Theorem~\ref{Theorem:ReconstructionError}, then
$$\E_{\cP \, \cup\,\cU }  W_1\left(\Prob_{X, \hD \circ \hH(X,\eta)}, \Prob_{X, \hD \circ \hE(Y)} \right)\lesssim  n^{-1/(2+\gamma)}  \log^2n.$$
where $\gamma = \max\left\{ \frac{m}{\beta_d},\frac{s}{\beta_e(\beta_d\wedge 1)} , \frac{p+d}{\beta_h}, p+m-2\right\}$, $s > d_\cY$ is arbitrary, and $\hH$ can be either $\hH_{cLSDM}$ or $\hH_{dLSDM}$. 
\end{theorem}
Theorem \ref{Theorem:DistributionMatchingConsistency} suggests that cLSDM and dLSDM converge at the same rate, which increases with component smoothness (larger $\beta_e,\beta_d,\beta_h$) and decreases with latent dimension $m$.
The autoencoder smoothness parameters $\beta_d,\beta_e$ enter the rate through the generalization error of the encoder $\hE$. Since the size of $\hE$ scales with the total sample size $n+N$ but only $n$ paired samples are available for Step 2, controlling this error necessitates $n \asymp N$.

% \ckyB{Under the stated assumptions, cLSDM and dLSDM converge at the same rate, which increases with component smoothness (larger $\beta_e,\beta_d,\beta_h$) and decreases with latent dimension $m$.
% \ckyB{The autoencoder smoothness parameters $\beta_d,\beta_e$ enter the rate through the generalization error of the encoder $\hE$ and Assumption \ref{Assumption:Smoothness}.} Since the size of $\hE$ scales with the total sample size $n+N$ but only $n$ paired samples are available for Step 2, controlling this error necessitates $n \asymp N$.}

\begin{remark}[Choice of noise dimension $d$] Based on Theorem~\ref{Theorem:DistributionMatchingConsistency}, a smaller noise dimension $d$ leads to a better convergence rate, suggesting that one may choose a small value of $d$, such as $d=1$. In practice, however, values of $d>1$ often yield better performance. \cite{WGR} proposed a BIC‑type criterion for selecting $d$. A similar approach can be adopted here by minimizing the objective in (\ref{Equation:cLSDMDistributionMatchingObjective}), augmented with a penalty term proportional to the noise dimension $d$ and the paired sample size $n$.
\end{remark}

\begin{remark}[Comparison with cWGAN]
In the conditional WGAN framework, \cite{WGR} derived a rate (ignoring log terms) involving $n^{-1/(2+\gamma)}$ with \(\gamma =  \max\{\frac{3(p+d)}{\beta_g},2(p+q+1)\}\), where \(\beta_g\) is the Hölder smoothness of the optimal conditional generator \(G^*\). In LSDM, $\gamma = \max\left\{ \frac{m}{\beta_d},\frac{s}{\beta_e(\beta_d\wedge 1)} , \frac{p+d}{\beta_h}, p+m-2\right\}$ where $s>d_\cY$ is arbitrary. When all functions are Lipschitz (i.e., \(\beta_g=\beta_d=\beta_e=\beta_h = 1\)), LSDM attains a faster distribution matching rate as $d_\cY<q$ and $m < q$.
\end{remark}

\subsection{Benefits of unpaired data}
Building on the analyses of reconstruction and distribution-matching errors, we can establish the consistency of the composite generator $\hG = \hD \circ \hH$. This consistency provides a clear understanding of the benefits of incorporating unpaired data in LSDM.
\begin{corollary}[Consistency of LSDM] \label{Corollary:TwoStepRate} Suppose Assumptions \ref{Assumption:Boundedness}, \ref{Assumption:Existence} and \ref{Assumption:Smoothness} hold. If $n \asymp N$, $m\geq d_\cY$, and the neural network classes $\cD,\cE,\cH$ are set as in Theorem \ref{Theorem:DistributionMatchingConsistency}, then
$$\E_{\cP \, \cup\,\cU }  W_1\left(\Prob_{X, \hD \circ \hH(X,\eta)}, \Prob_{X, Y} \right) \leq \;\underbrace{\cO\left((n+N)^{-1/(2+\kappa)}\log^2(n+N)\right)}_{\text{Reconsturction gap}} \quad  + \;\underbrace{\cO\left(n^{-1/(2+\gamma)}  \log^2n\right)}_{\text{Latent distribution matching}} ,$$
where $\kappa = \max\left\{\frac{m}{\beta_d}, \frac{s}{\beta_e(\beta_d\wedge1)} \right\}$, $\gamma =\max\left\{ \frac{m}{\beta_d},\frac{s}{\beta_e(\beta_d\wedge 1)} , \frac{p+d}{\beta_h}, p+m-2\right\}$, and $\hH$ can be either $\hH_{cLSDM}$ or $\hH_{dLSDM}$.
\end{corollary}
Corollary~\ref{Corollary:TwoStepRate} shows that the learning of the generative mapping $\cZ \to \cY$ is decoupled from that of the conditional mapping $\cX \to \cZ$, each with its own convergence rate and sample size. A large total sample size $n+N$ improves generation quality via a reduced autoencoder reconstruction error, tightening the upper-bound. The distribution‑matching rate depends on the paired sample size $n$ and the low‑dimensional quantities $d_\cY$ and $m$, rather than the full ambient dimension $q$. When the target autoencoder and latent generator are smooth (i.e., large $\beta_d,\beta_e,\beta_h$), the Step 2 convergence rate improves. Theoretically, this improvement manifests as a smaller required network size to attain the optimal rate.

Beyond the improved reconstruction error rate, incorporating unpaired data in LSDM offers an additional advantage: a more accurate approximation of the data support by the learned decoder~$\widehat{D}$. When $\cY$ exhibits a low-dimensional structure, Step~1 of LSDM captures the underlying data geometry by learning a better decoder from the combined sample of size $n+N$. Recall that the overall generator is composite, $\widehat{G} = \widehat{D} \circ \widehat{H}$, so the generated samples inherit the geometric structure encoded in~$\widehat{D}$. By fixing the decoder in Step~2, the samples are constrained to its range, which serves as an implicit form of regularization.
For many data modalities, adherence to the intrinsic geometry is critical for perceptual quality. In facial images, for example, a realistic appearance requires correctly positioned eyes, ears, and mouth. By exploiting the structure learned from abundant unpaired data, the generated samples conform to this geometry even when the conditional mapping in Step~2 is imperfect. Consequently, the outputs remain visually plausible even if they do not precisely match a given predictor~$X$. We formalize this observation under suitable regularity conditions below.

Let $\widehat{Z} = \hE(Y) \in \cZ$ denote the learned latent code, and assume it admits a density $p(\hat{z})$ with respect to Lebesgue measure. Define the distance from a point to the response support as $\text{dist}(x, \cY) = \inf_{y\in \cY} \dnorm{x-y}_2$. 
\begin{theorem}[Uniform Approximation of the Data Support]  \label{Theorem:RangeProximity} Suppose the conditions of Corollary \ref{Corollary:FullReconstruction} hold. Let $g(z) = \text{dist}(\hD(z), \cY)$ and assume there exists a constant $\rho \in (0,1)$ such that $\int_{B}p(\hat{z})d\hat{z}\geq \rho, B = \bigl\{z\in \cZ:g(z)\geq \sup_{z \in \cZ}g(z)/2\bigr\}$ almost surely. Then,
$$ \E_{\cP \, \cup\,\cU }  \; \sup_{z \in \cZ} \,\text{dist}(\hD(z), \cY) \lesssim \rho^{-1} (n+N)^{-1/(2+\xi)} \log^2(n+N),$$
where $\xi = m\vee s$ and $s>d_\cY$ is arbitrary. 
\end{theorem}
Theorem~\ref{Theorem:RangeProximity} implies that, as the combined sample size $n+N$ increases, the range of the learned decoder converges to the support of the response distribution. Consequently, the generated samples lie close to~$\mathcal{Y}$ with high probability. This, in turn, indicates that incorporating unpaired data in Step~1 enhances the geometric fidelity of the outputs, yielding perceptually more realistic samples. 
We visualize this effect in Figure~\ref{MNIST:latentspace}
 through an example, which shows that, for fixed~$n$, the generated samples become progressively sharper as~$N$ increases.

\section{Numerical Experiments} 
In this section, we present two real-image data experiments to assess the performance of LSDM in semi-supervised generative learning tasks. We begin with an class-conditional image generation task on the MNIST dataset, followed by super-resolution on the CelebA dataset. 

%We conduct comprehensive experiments to thoroughly evaluate LSDM with varying number of labeled and unlabeled samples, as well as the latent dimension.
\label{Section:Experiments}

\subsection{MNIST Conditional Generation}

We consider a conditional generation task on the MNIST dataset, which consists of $28\times 28$ grayscale images of handwritten digits $0$ to $9$. Owing to its intrinsically low‑dimensional structure, MNIST is widely used for benchmarking generative and dimension‑reduction models. The class label $x \in \{0, \dots, 9\}$ serves as the predictor $X$ and the image as the response $Y$. The goal is to generate images $Y$ corresponding to a given label $X=x$.

We also examine discrete latent embeddings as a form of regularization, using a vector quantized variational autoencoder (VQVAE; \citealp{VQVAE}). Our experiment examines: (1) the impact of unpaired data on generation quality via an ablation study; (2) the effects of autoencoder regularization (WAE, VQVAE) and Step 2 divergence measures (KL, JS, $W_1$); and (3) performance comparisons against common baselines. We adopt a semi‑supervised setting where paired data $(X,Y)$ are scarce but unpaired responses $Y$ are abundant. Specifically, we use 
$n=125$ to $1{,}500$ paired samples while varying the number of unpaired samples from $N=0$ to $29{,}750$. This setup is particularly suitable for evaluating the semi‑supervised capabilities of LSDM, as the mapping from $X$ to the latent space $Z$ is relatively straightforward, whereas generating an image from a latent code is more complex. This contrast is expected to emphasize the effect of incorporating additional unpaired 
 data on generation quality. 

Following \citet{Pope2021}, an appropriate intrinsic dimension of MNIST  lies between 7 and 13. We choose the upper bound, $m=13$, for our latent space.
For comparison, we include the following baselines: conditional GAN (cGAN; \citealp{cGAN}), conditional Wasserstein GAN (cWGAN; \citealp{WassDCG}), conditional variational autoencoder (cVAE; \citealp{cVAE}), and latent diffusion models (LDM; \citealp{LDM}). For LDMs we consider two variants: (i) the autoencoder is pre‑trained using only the paired data (denoted pLDM), and (ii) the autoencoder is pre‑trained using both paired and unpaired data (denoted LDM). This setup enables a direct comparison between the influence of unpaired data in LSDM and its effect within a conventional pre‑train–fine‑tune framework. All models employ the same convolutional architecture and are trained with standard hyperparameters. Further implementation details are provided in the supplementary material.
The quality of generated images is quantitatively assessed using the Fréchet Inception Distance (FID; \citealp{FID}). This metric evaluates how closely the distribution of generated images matches that of real images by comparing the mean and covariance of features obtained from a pre‑trained Inception‑v3 network. A lower FID indicates better quality. Definition of FID is provided in the supplementary material. All quantitative results are computed on a test set of sample size $10{,}000$.

\begin{figure}[ht]
\centering
\includegraphics[scale=0.3]{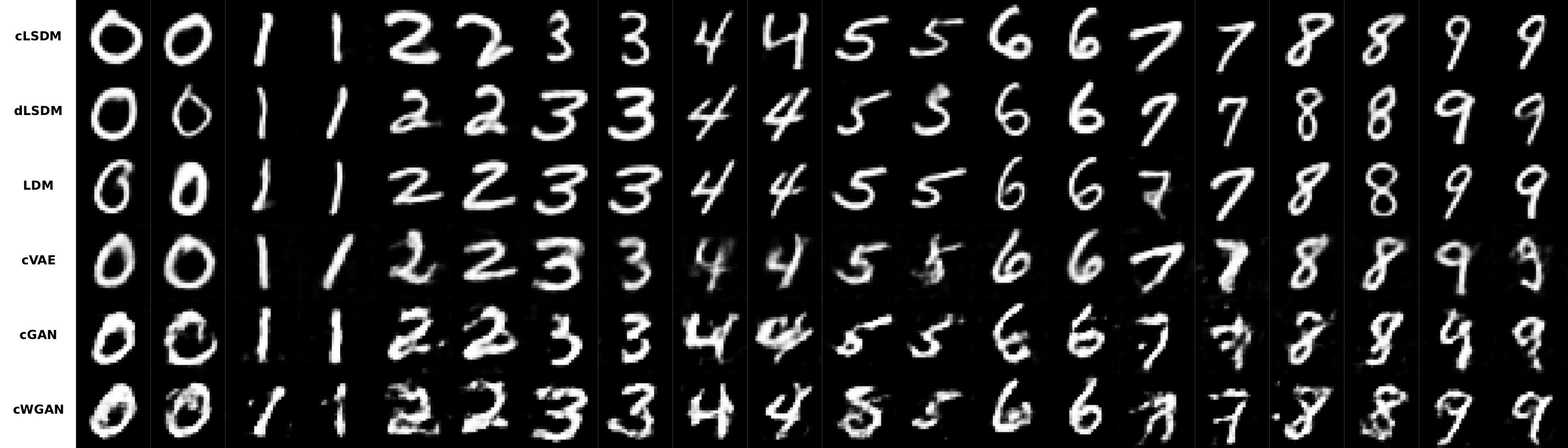}
\small
\caption{Qualitative result on the MNIST dataset ($n=250$, $N=29{,}750$, $m=13$).}\label{MNIST:qualitative}
\end{figure}

Table \ref{MNIST:quantitative} reports FID scores for models with varying values of $n$. Models that exploit unpaired $Y$ (cLSDM, dLSDM, LDM) achieve noticeably lower FID scores than their fully‑supervised counterparts. LDM performs slightly worse than LSDM when $n$ is small, but the gap narrows as $n$ grows, consistent with the empirical observation that GAN‑based methods often excel with limited data while diffusion models scale more effectively. For both LSDM and LDM, generation quality improves as $n$ increases while $n+N$ is fixed at $30{,}000$. For $n=250$, the FID of LSDM is about half that of the fully‑supervised baselines. Moreover, LSDM performs well when the Step 2 objective is replaced by JS divergence but poorly with KL divergence, whose training dynamics are highly unstable. This supports the practical preference for the original 1‑Wasserstein formulation. For cLSDM, autoencoder regularization via WAE improves generation quality. Qualitative results are displayed in Figure \ref{MNIST:qualitative}.

% \FloatBarrier
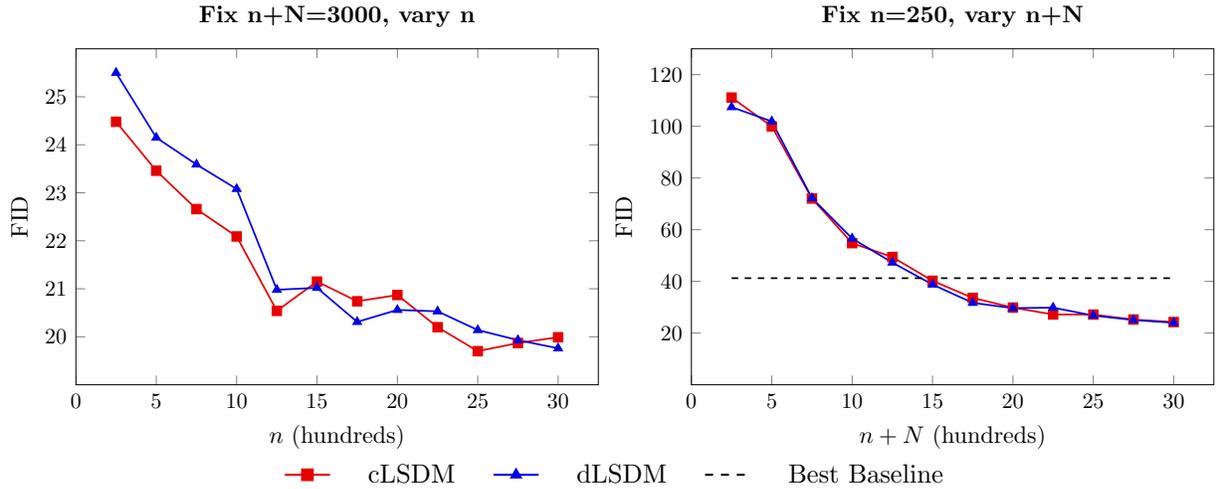
\begin{figure}[ht]
\centering
\begin{tikzpicture}[scale=0.825] 
\begin{axis}[
    name=ax1,
    title={Fix n+N=3000, vary n},
    width=10cm,
    height=7cm,
    xlabel={$n$ (hundreds)},
    ylabel={FID},
    xmin=0, xmax=32.5,
    ymin=19, ymax=26,
    xtick={0, 5, 10, 15, 20, 25, 30},
    ytick={ 20, 21,22, 23,24, 25},
    legend pos=north east,
    tick style={thin},
    ticklabel style={font=\footnotesize},
    label style={font=\small},
    title style={font=\small\bfseries},
    legend style={font=\small, draw=none, fill=white, fill opacity=0.9, text opacity=1},
    every axis plot/.append style={thick},
    scaled y ticks=false,
    y tick label style={
        /pgf/number format/fixed,
        /pgf/number format/.cd,
        fixed zerofill,
        precision=0
    },
]
\addplot[
    color=red!90!black,
    mark=square*,
    ]
    coordinates {
   (2.5, 24.48) (5, 23.46) (7.5, 22.66) (10, 22.09) (12.5, 20.54) (15, 21.15) (17.5, 20.74) (20, 20.87) (22.5, 20.20) (25, 19.70) (27.5, 19.87) (30, 19.99)};
\label{ablation:cLSDM}
\addplot[
    color=blue!90!black,
    mark=triangle*,
    ]
    coordinates {(2.5, 25.50) (5, 24.15) (7.5,23.59) (10, 23.08) (12.5, 20.98) (15, 21.02) (17.5, 20.31) (20, 20.56) (22.5, 20.53) (25,20.14) (27.5, 19.93) (30, 19.76)
};
\label{ablation:dLSDM}
\end{axis}

\begin{axis}[
        at={(ax1.south east)},
        xshift=1.5cm,
            width=10cm,
            height=7cm,
    title={Fix n=250, vary n+N},
    xlabel={$n+N$ (hundreds)},
    ylabel={FID},
    xmin=0, xmax=32.5,
    ymin=0, ymax=130,
    xtick={0, 5, 10, 15, 20,  25, 30 },
    ytick={20, 40, 60, 80, 100, 120},
    legend pos=north east,
    tick style={thin},
    ticklabel style={font=\footnotesize},
    label style={font=\small},
    title style={font=\small\bfseries},
    legend style={font=\small, draw=none, fill=white, fill opacity=0.9, text opacity=1},
    every axis plot/.append style={thick},
    scaled y ticks=false,
    y tick label style={
        /pgf/number format/fixed,
        /pgf/number format/.cd,
        fixed zerofill,
        precision=0
    },
]
\addplot[
    color=red!90!black,
    mark=square*,
    ]
    coordinates {
    (2.5,111.12)(5,99.97)(7.5,72.07)(10, 54.80)(12.5,49.42)(15,40.24)(17.5,33.61)(20, 29.86) (22.5, 27.20)(25,27.16)(27.5,25.24)(30,24.26)};\label{plot:cLSDM}
\addplot[
    color=blue!90!black,
    mark=triangle*,
    ]
    coordinates {
    (2.5,107.43380)(5,101.90734)(7.5,72.22736)(10, 56.60303)(12.5,47.28194)(15,38.780715)(17.5,31.69911)(20, 29.66109) (22.5, 29.83943)(25,26.78973)(27.5,25.04384)(30,24.0118)};\label{plot:dLSDM}

    \addplot[dashed,
    color=black,
    mark=none,
    ]
    coordinates {
    (2.5,41.24643325805664)(30,41.24643325805664)
    };\label{plot:BestBaseline}

\end{axis}
\end{tikzpicture}
\vspace*{-18pt} 
\begin{center}
\footnotesize
    \begin{tikzpicture}
        \matrix[
            matrix of nodes, 
            anchor=center,   
            nodes={inner sep=2pt, font=\footnotesize}, 
            column sep=10pt, 
            row sep=0pt,     
        ]  {
            \ref{plot:cLSDM} & cLSDM & 
            \ref{plot:dLSDM} &dLSDM & 
            \ref{plot:BestBaseline} & Best Baseline\\
            % \ref{plot:n300} &dLSDM &
            % \ref{plot:n400} & LDM \\
        };
    \end{tikzpicture}
\end{center}
\vspace*{-10pt} 
\small
\caption{Ablation study of LSDM on the MNIST dataset.
Left: Total sample size $n + N$ is fixed at $3{,}000$ while $n$ varies.
Right: Number of paired samples $n$ is fixed at $250$ while the total sample size $n + N$ varies. Model architecture and training parameters are fixed within each study but differ between the two studies.
}
\label{MNIST:ablation}
\end{figure}

To examine the effect of unpaired $Y$ on generation quality, we conduct two ablation studies with the same architecture and training settings within each study. In the first experiment, the total sample size is fixed at $n + N = 3{,}000$, while the number of paired samples $n$ is varied.
In the second experiment, $n$ is held constant at $250$, and the number of unpaired samples $N$ is varied. Results are presented in Figure \ref{MNIST:ablation}.
In the study with a fixed total $n + N = 3{,}000$, generation quality generally improves as the proportion of paired samples $n$ increases, consistent with expectations.
In the study with $n = 250$, generation quality, measured by FID, improves as $N$ increases.
For total sample sizes $n + N \ge 1{,}500$, LSDM surpasses the strongest baseline, cWGAN. 
Below this threshold, LSDM performs worse.  This crossover is consistent with Corollary~\ref{Corollary:TwoStepRate}, which shows that the upper‑bound objective can be loose when $N$ is small. Notably, the quality obtained with $n = 3{,}000$ paired samples and $N=0$ is comparable to that achieved with $n = 250$ paired and $N = 2{,}750$ unpaired samples.
This suggests that for class‑conditional generation, where the mapping from $X$ to $Z$ is relatively simple, the total number of response samples has a greater influence than the number of strictly paired observations.

\begin{figure}[ht]
\begin{minipage}{0.5\linewidth}
\centering
\includegraphics[scale=0.3]{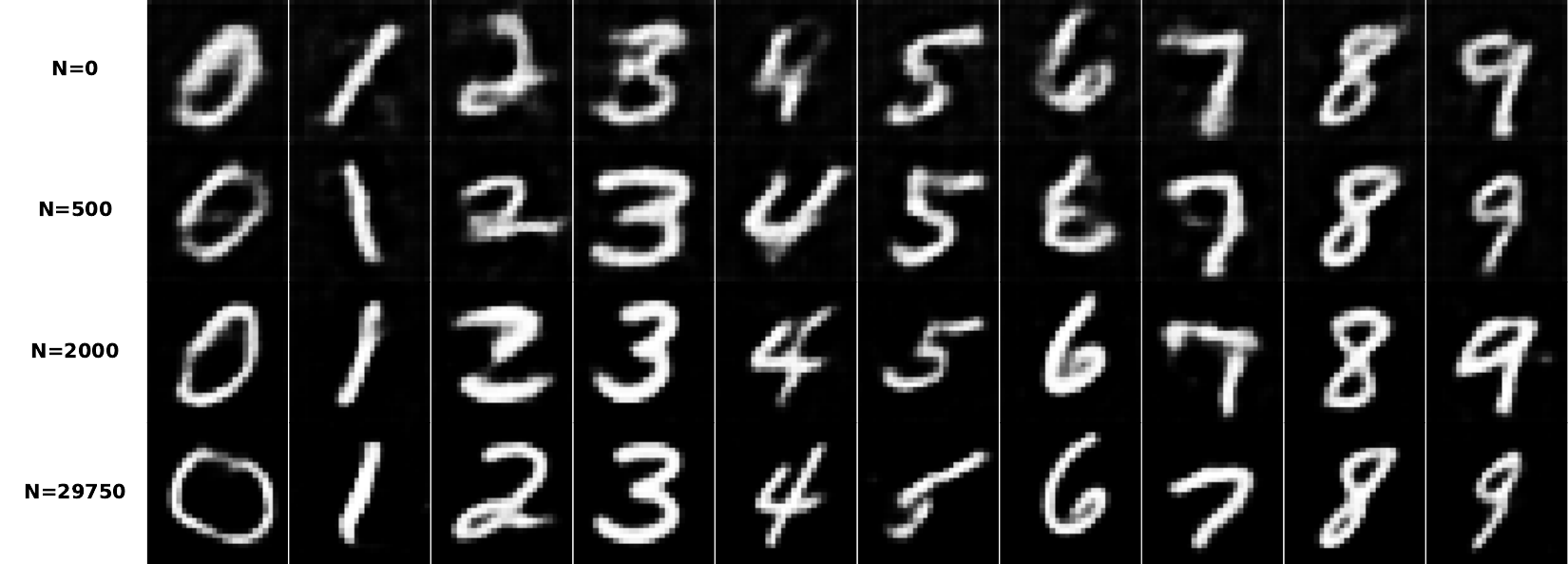}
\end{minipage}
\begin{minipage}{0.44\linewidth}
\centering
\includegraphics[scale=0.25]{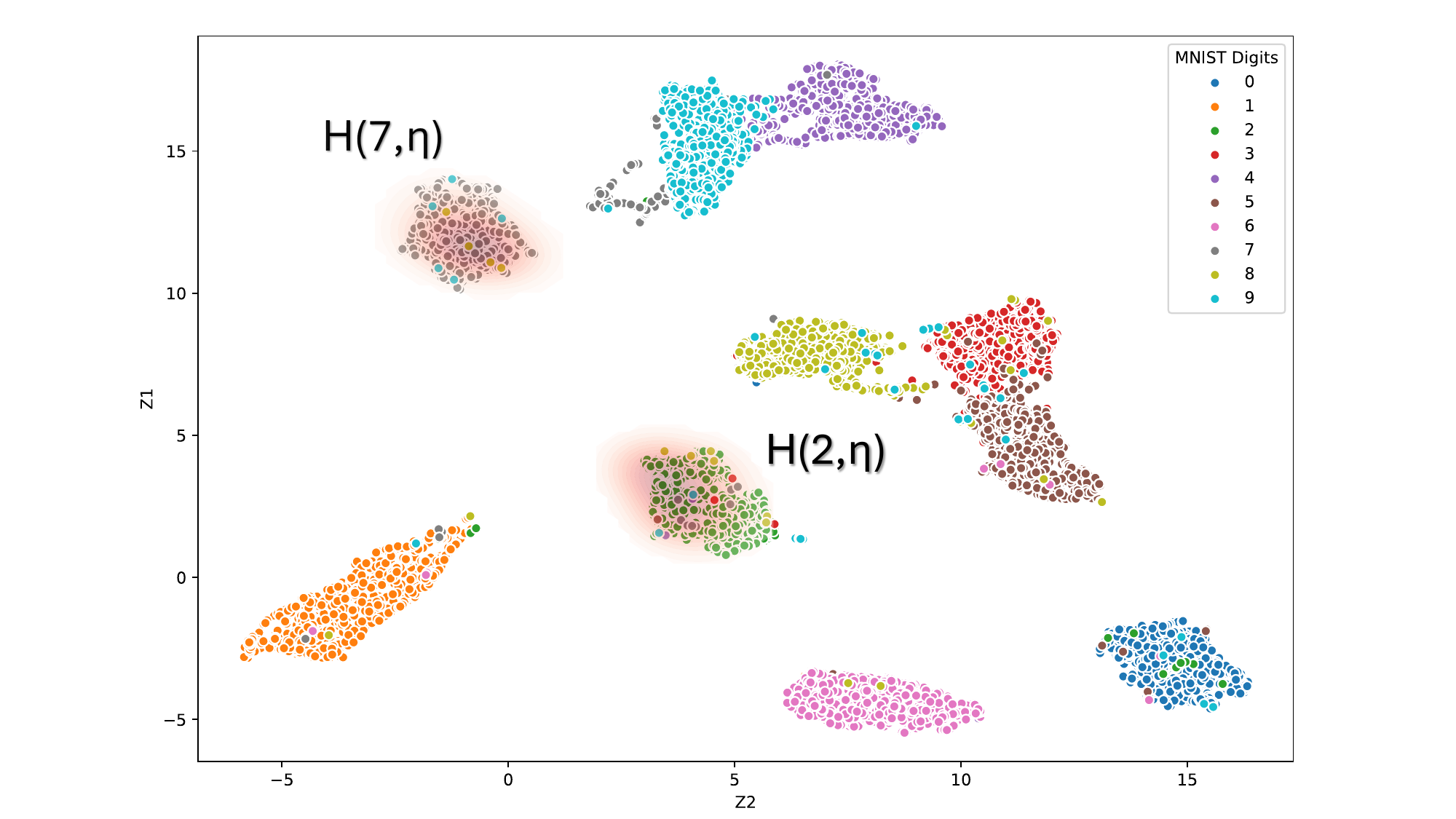}
\end{minipage}
\small
\caption{Left: Qualitative changes of cLSDM on the MNIST dataset for fixed $n=250$ and varying $N$. Right: UMAP-reduced latent space of an autoencoder on the MNIST dataset. Red contours represent the distribution of generated latent codes $H(X, \eta)$ in cLSDM.}
\label{MNIST:latentspace}
\end{figure}

Finally, we illustrate with real data that the latent generator in LSDM induces a structured distribution in the latent space, as illustrated in Figure \ref{Figure:LSDM}. We first apply Uniform Manifold Approximation and Projection (UMAP; \citealp{Umap}) to project the latent codes from a standard autoencoder onto two dimensions, coloring points by their digit class. We then generate $5{,}000$ samples from the latent generator $H$ of cLSDM. As shown in Figure \ref{MNIST:latentspace}, conditioned on a label $X=x$ (e.g., $x=2$ or $x=7$), the generator produces latent codes that cluster around the latent code of the corresponding digit class. A similar structure is observed for dLSDM.

\begin{table}[ht]
\centering
\small
\setlength{\tabcolsep}{5pt}
\renewcommand{\arraystretch}{0.6}
\begin{minipage}{0.55\linewidth}
\centering
\begin{tabular}{llcll}
\hline
  &\multicolumn{4}{c}{FID$\downarrow$}\\
\hline
Models &n=125&n=250&n=500  &n=1500\\
\hline
cLSDM &20.49&19.50&17.72 &15.05\\
dLSDM &23.86&21.41&21.17 &18.43\\
LDM&31.63&25.29&22.82 &18.72\\
pLDM&144.72&105.06&96.69 &30.17\\
cVAE &66.41&47.13&33.80 &24.15\\
cGAN &68.84& 61.42& 30.19&18.20\\
 cWGAN & 52.38& 41.25& 25.64 &18.44\\
  && & &\\
\hline
\end{tabular}
\end{minipage}
\hfill
\begin{minipage}{0.36\linewidth}
\centering
\begin{tabular}{lc}
\hline
 &FID$\downarrow$\\
\hline
Models& n=250\\
\hline
cLSDM-VQVAE-W1& 21.59\\
cLSDM-WAE-W1& \textbf{18.96}\\
dLSDM-VQVAE-W1& 21.37\\
dLSDM-WAE-W1& 22.17\\
cLSDM-AE-KL&105.05\\
cLSDM-AE-JS&22.37\\
dLSDM-AE-KL&120.05\\
 dLSDM-AE-JS&30.25\\
 \hline
\end{tabular}
\end{minipage}
\vspace*{5pt} 
\caption{Quantitative Result on the MNIST dataset ($N=29{,}750$, $m=13$).}
\label{MNIST:quantitative}
\end{table}

\FloatBarrier

\subsection{CelebA Image Super-resolution}
The image super‑resolution task seeks to reconstruct a high‑resolution image from a low‑resolution input.
We conduct experiments on the CelebA dataset \citep{CelebA}, a widely used benchmark for generative models due to its inherent compressibility.

CelebA comprises $202{,}599$ RGB images of celebrity faces at a resolution of $178 \times 218$ pixels.
To reduce computational overhead, we crop the central $140 \times 140$ region and downsample it to $(3,64,64)$, which serves as the high‑resolution response $Y$.
The low‑resolution input $X$ is obtained by further downsampling $Y$ to $(3,16,16)$.
The task is to sample from the conditional distribution $\Prob_{Y \mid X}$, corresponding to a challenging $4\times$ up‑sampling problem in a data‑scarce regime.
Our study examines three main aspects:
(1) the impact of unpaired data on generation quality via an ablation study;
(2) the effect of latent dimension size on generation quality; and
(3) performance comparison with mainstream baselines.

For image tasks that demand preservation of fine details under a bottleneck constraint, it is common to apply vector‑quantization (VQ) regularization to the spatial latent code, as in \citet{LDM}.
Following this approach, we use spatial latent codes of shape $(m_c,16,16)$ with $m_c \in \{2,4,6\}$, corresponding to $24\times$, $12\times$, and $8\times$ reductions from the original $(3,64,64)$ images. Prior work on latent diffusion models identified $(4,16,16)$ as achieving the best trade-off between FID and dimension reduction. Besides FID, we also report Learned Perceptual Image Patch Similarity (LPIPS; \citealp{LPIPS}) for quantitative evaluation. LPIPS measures the perceptual similarity between two images using the activation differences of a pre-trained network. We employ the VGG backbone \citep{VGG} to compute LPIPS. Lower values indicate greater perceptual similarity between  the super‑resolved and true images. We also include the Structural Similarity Index Measure (SSIM), a pixel‑based metric that compares luminance and contrast. Definitions of the metrics are provided in the supplementary material. All metrics are computed on a test set of $19{,}962$ images.

Table \ref{CelebA:quantitative}
presents quantitative comparisons between LSDM and baseline models, while Figure~\ref{CelebA:Qualitative} illustrates generated samples.
%compares LSDM with common baselines and Figure \ref{CelebA:Qualitative} provides a visual comparison of generated samples.
Models that leverage unpaired $Y$ data (cLSDM, dLSDM, LDM) achieve notably lower FID and LPIPS scores and better SSIM scores than their fully supervised counterparts. LSDM slightly outperforms LDM, although the gap narrows as the paired sample size $n$ increases, consistent with the empirical observation that GAN‑based methods often excel with limited data while diffusion models scale more effectively. %Figure \ref{CelebA:Qualitative} provides a visual comparison of generated samples. The output of cLSDM, dLSDM, and LDM is perceptually superior to that of baselines that do not exploit unpaired data.

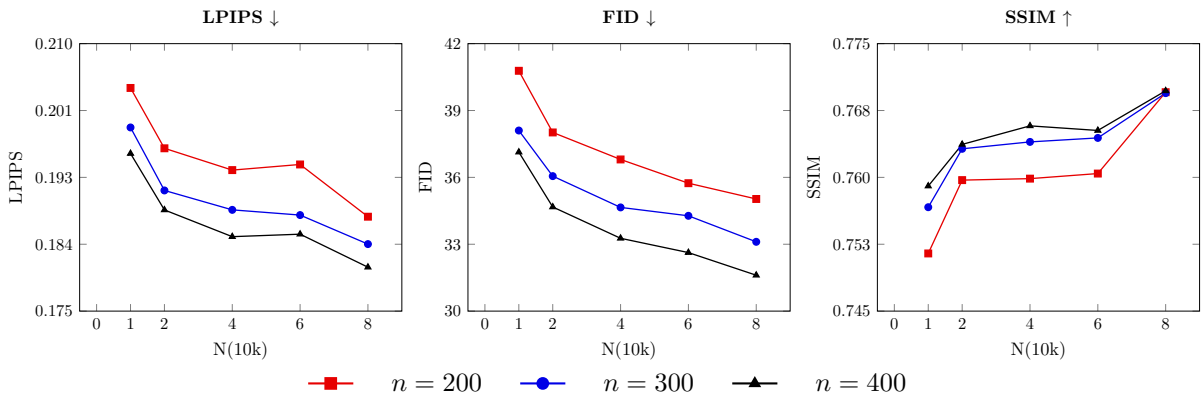
\begin{figure}[ht]
\centering
% First plot: FID
\begin{tikzpicture}[scale=0.625] 
\begin{axis}[
    title={LPIPS $\downarrow$},
    xlabel={N(10k)},
    ylabel={LPIPS},
    xmin=-0.5, xmax=9,
    ymin=0.175, ymax=0.21,
    xtick={0,1,2,4,6,8},
    ytick={0.175, 0.18375, 0.1925, 0.20125,0.21},
    tick style={thin},
    ticklabel style={font=\footnotesize},
    label style={font=\small},
    title style={font=\small\bfseries},
    every axis plot/.append style={thick},
    scaled y ticks=false,
    y tick label style={
        /pgf/number format/fixed,
        /pgf/number format/.cd,
        fixed zerofill,
        precision=3
    },
]
\addplot[
color=red!90!black,
mark=square*]
    coordinates {
    % (1,0.205609180372853)(2,0.199553850525761)(4,0.197840901776027)(6,0.194184274141254)(8,0.1887204270362)
    (1,0.204203407339958)(2,0.196305819305888)(4,0.193451528641877)(6,0.194189677340224)(8,0.1873540123459)
    };
    \label{plot:n200}
\addplot[
color=blue!90!black,
mark=*]
    coordinates {
    % (1,0.199426498409264)(2,0.193618478928858)(4,0.190745522056023)(6,0.18768643601936)(8,0.183724711479016)
    (1,0.199032644866739)(2,0.190789490111091)(4,0.1882493406364)(6,0.18756743731975)(8,0.183770098627456)
    };
        \label{plot:n300}
\addplot[
color=black!90!black,
mark=triangle*]
    coordinates {
    % (1,0.197007751694161)(2,0.190384240490128)(4,0.189211036528103)(6,0.185430704836687)(8,0.180680777312113)
    (1,0.195623694793362)(2,0.188240831952429)(4,0.184741210870616)(6,0.18506695571852)(8,0.180747232818374)
    };
        \label{plot:n400}
\end{axis}
\end{tikzpicture}
\begin{tikzpicture}[scale=0.625] 
\begin{axis}[
    title={FID $\downarrow$},
    xlabel={N(10k)},
    ylabel={FID},
    xmin=-0.5, xmax=9,
    ymin=30, ymax=42,
    xtick={0,1,2,4,6,8},
    ytick={30,33,36, 39, 42},
    tick style={thin},
    ticklabel style={font=\footnotesize},
    label style={font=\small},
    title style={font=\small\bfseries},
    every axis plot/.append style={thick},
    scaled y ticks=false,
    y tick label style={
        /pgf/number format/fixed,
        /pgf/number format/.cd,
        fixed zerofill,
        precision=0
    },
]
\addplot[
color=red!90!black,
mark=square*]
    coordinates {
    % (1,40.0326995849609)(2,39.5853195190429)(4,37.5370368957519)(6,37.1157913208007)(8,36.973632812)
    (1,40.7819709777832)(2,38.0157661437988)(4,36.8039779663086)(6,35.7385101318359)(8,35.0311546325684)
    };
    
\addplot[
    color=blue!90!black,
mark=*]
    coordinates {
    % (1,38.4035568237304)(2,37.0430488586425)(4,35.742622375488)(6,34.9532432556152)(8,34.9900398254394)
    (1,38.1023292541504)(2,36.0562438964844)(4,34.6524429321289)(6,34.2758445739746)(8,33.1088447570801)
    };
    
\addplot[
    color=black!90!black,
mark=triangle*]
    coordinates {
    % (1,37.5320)(2, 35.7544)(4, 35.5128)(6, 34.2354)(8, 33.3583)
(1,37.1333122253418)(2,34.6719398498535)(4,33.2667541503906)(6,32.6258773803711)(8,31.6101951599121)
    };
    
\end{axis}
\end{tikzpicture}
% Third plot: SSIM
\begin{tikzpicture}[scale=0.625] 
\begin{axis}[
    title={SSIM $\uparrow$},
    xlabel={N(10k)},
    ylabel={SSIM},
    xmin=-0.5, xmax=9,
    ymin=0.745, ymax=0.775,
    xtick={0,1,2,4,6,8},
    ytick={0.745,0.7525, 0.76, 0.7675, 0.775},
    tick style={thin},
    ticklabel style={font=\footnotesize},
    label style={font=\small},
    title style={font=\small\bfseries},
    every axis plot/.append style={thick},
    scaled y ticks=false,
    y tick label style={
        /pgf/number format/fixed,
        /pgf/number format/.cd,
        fixed zerofill,
        precision=3
    },
]
\addplot[
color=red!90!black,
mark=square*]
    coordinates {
    % (1,0.750907396126003)(2,0.753598803975111)(4,0.755392458500741)(6,0.75926362998785)(8,0.767543325681221)
    (1,0.751467633588963)(2,0.75969217592986)(4,0.759858713779692)(6,0.760428200001206)(8,0.769552743932955)
    };
    
\addplot[
color=blue!90!black,
mark=*]
    coordinates {
    % (1,0.756165195354633)(2,0.759567585417603)(4,0.76075133593213)(6,0.763151330874303)(8,0.769034186270442)   
    (1,0.756653852303202)(2,0.763204907095393)(4,0.763978838693665)(6,0.764426669858522)(8,0.769449799658719)
    };
    
\addplot[
color=black!90!black,
mark=triangle*]
    coordinates {
    % (1,0.758368478654059)(2,0.760625819932505)(4,0.762801910184832)(6,0.764353233400846)(8,0.76975420416864)        
    (1,0.759026920779914)(2,0.763698730950317)(4,0.765778045188018)(6,0.765262575095073)(8,0.769707899415628)
    };
\end{axis}
\end{tikzpicture}
\vspace*{-18pt} 
\begin{center}
\footnotesize
    \begin{tikzpicture}
        \matrix[
            matrix of nodes, 
            anchor=center,   
            nodes={inner sep=2pt, font=\footnotesize}, 
            column sep=10pt, 
            row sep=0pt,     
        ]  {
            \ref{plot:n200} & $n=200$ & 
            \ref{plot:n300} & $n=300$ &
            \ref{plot:n400} & $n=400$ \\
        };
    \end{tikzpicture}
\end{center}
\vspace*{-12pt} 
\caption{\footnotesize Quantitative results  on the CelebA dataset for varying $n$ and $N$ ($m_c=4$, cLSDM).}
\label{CelebA:figure1}
\end{figure}

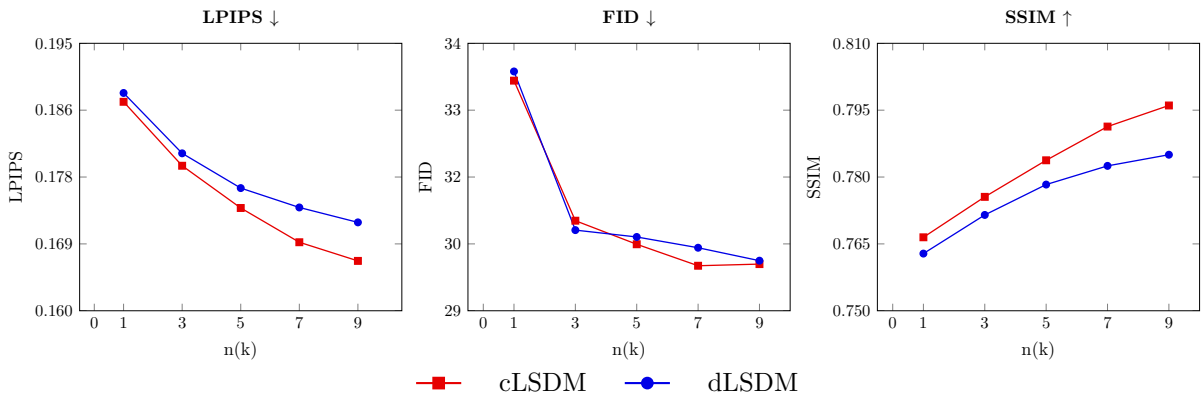
\begin{figure}[ht]
\centering
% First plot: FID
\begin{tikzpicture}[scale=0.625] 
\begin{axis}[
    title={LPIPS $\downarrow$},
    xlabel={n(k)},
    ylabel={LPIPS},
    xmin=-0.5, xmax=10.5,
    ymin=0.16, ymax=0.195,
    xtick={0,1,3,5,7,9},
    ytick={0.16, 0.16875, 0.1775,0.18625, 0.195},
    tick style={thin},
    ticklabel style={font=\footnotesize},
    label style={font=\small},
    title style={font=\small\bfseries},
    every axis plot/.append style={thick},
    scaled y ticks=false,
    y tick label style={
        /pgf/number format/fixed,
        /pgf/number format/.cd,
        fixed zerofill,
        precision=3
    },
]
\addplot[
color=red!90!black,
mark=square*]
    coordinates {
    % (1,0.187305321543886)(3, 0.178792093120468)(5,0.172953896646257)(7,0.169901318402486)(9,0.16737143485296)
    (1,0.187346230789626)(3,0.178976574828875)(5,0.173449391031681)(7,0.168966460880088)(9,0.166526475921087)
    };
\addplot[
color=blue!90!black,
mark=*]
    coordinates {
    % (1, 0.189798374690078)(3,0.182505884458134)(5,0.176914690587363)(7,0.174578120672395)(9, 0.172719109644811)
    (1,0.188499563431384)(3,0.180601351223636)(5,0.176052914080454)(7,0.173517809018333)(9,0.171566725422177)
    };

\end{axis}
\end{tikzpicture}
\begin{tikzpicture}[scale=0.625] 
\begin{axis}[
    title={FID $\downarrow$},
    xlabel={n(k)},
    ylabel={FID},
    xmin=-0.5, xmax=10,
    ymin=29, ymax=34,
    xtick={0,1,3,5,7,9},
    ytick={29,30.25,31.5,32.75,34},
    tick style={thin},
    ticklabel style={font=\footnotesize},
    label style={font=\small},
    title style={font=\small\bfseries},
    every axis plot/.append style={thick},
    scaled y ticks=false,
    y tick label style={
        /pgf/number format/fixed,
        /pgf/number format/.cd,
        fixed zerofill,
        precision=0
    },
]
\addplot[
color=red!90!black,
mark=square*]
    coordinates {
    % (1,32.7894477844238)(3,30.4653739929199)(5,29.4855918884277)(7,29.135398864746)(9,29.1315402984619)
    (1,33.3020133972168)(3,30.6837215423584)(5,30.2432250976563)(7,29.8415927886963)(9,29.8722877502441)
    };
    
\addplot[
    color=blue!90!black,
mark=*]
    coordinates {
    % (1,33.0830993652343)(3,30.604621887207)(5,29.8868694305419)(7,29.5512599945068)(9,29.4628829956054)
        (1,33.4737854003906)(3,30.5077705383301)(5,30.3794212341309)(7,30.1781005859375)(9,29.9362392425537)
    };

\end{axis}
\end{tikzpicture}
% Third plot: SSIM
\begin{tikzpicture}[scale=0.625] 
\begin{axis}[
    title={SSIM $\uparrow$},
    xlabel={n(k)},
    ylabel={SSIM},
    xmin=-0.5, xmax=10,
    ymin=0.75, ymax=0.81,
    xtick={0,1,3,5,7,9},
    ytick={0.75,0.765, 0.78, 0.795, 0.81},
    tick style={thin},
    ticklabel style={font=\footnotesize},
    label style={font=\small},
    title style={font=\small\bfseries},
    every axis plot/.append style={thick},
    scaled y ticks=false,
    y tick label style={
        /pgf/number format/fixed,
        /pgf/number format/.cd,
        fixed zerofill,
        precision=3
    },
]
\addplot[
color=red!90!black,
mark=square*]
    coordinates {
    % (1,0.766876122478779)(3,0.775405797603887)(5,0.783520727116506)(7,0.78844361572296)(9,0.79315491303594)
    (1,0.766481185124234)(3,0.775554727497565)(5,0.783756627980125)(7,0.791319870516432)(9,0.796053026759449)
    };
    
\addplot[
color=blue!90!black,
mark=*]
    coordinates {
    % (1,0.762898756332405)(3,0.770969312901177)(5,0.777957267100031)(7,0.781498526896232)(9,0.784270637129246)
    (1,0.762826702659492)(3,0.771496184279978)(5,0.778311328221008)(7,0.782503365777607)(9,0.785005035148141)
    };
    
\end{axis}
\end{tikzpicture}
\vspace*{-18pt} 
\begin{center}
\footnotesize
    \begin{tikzpicture}
        \matrix[
            matrix of nodes, 
            anchor=center,   
            nodes={inner sep=2pt, font=\footnotesize}, 
            column sep=10pt, 
            row sep=0pt,     
        ]  {
            \ref{plot:n200} & cLSDM & 
            \ref{plot:n300} & dLSDM \\
        };
    \end{tikzpicture}
\end{center}
\vspace*{-12pt} 
\caption{\footnotesize Quantitative results on the CelebA dataset for varying $n$ while fixing $n+N=10{,}000$, $m_c=4$.}
\label{CelebA:figure4}
\end{figure}

Figures~\ref{CelebA:figure1} and~\ref{CelebA:figure4} present ablation studies examining the effect of paired sample size $n$ and unpaired sample size $N$ on generation quality. In Figure~\ref{CelebA:figure1}, $n$ is held constant while $N$ varies, and quality improves across all three metrics as $N$ increases, consistent with Theorem~\ref{Theorem:RangeProximity}. A decoder trained on abundant unpaired data produces more realistic samples that better conform to the data geometry learned in Step~1, thereby mitigating artifacts and perceptually implausible patterns.  
In Figure~\ref{CelebA:figure4}, the total sample size $n+N$ is fixed at $10{,}000$ while $n$ varies. Performance improves with increasing $n$. Notably, when $n = 9{,}000$ and $n+N=10{,}000$, FID and LPIPS scores are only approximately $6\%$ and $9\%$ better, respectively, than in the case with $n = 400$ and $n+N = 80{,}000$. This highlights the pronounced effect of the unpaired sample size $N$ on perceptual quality.

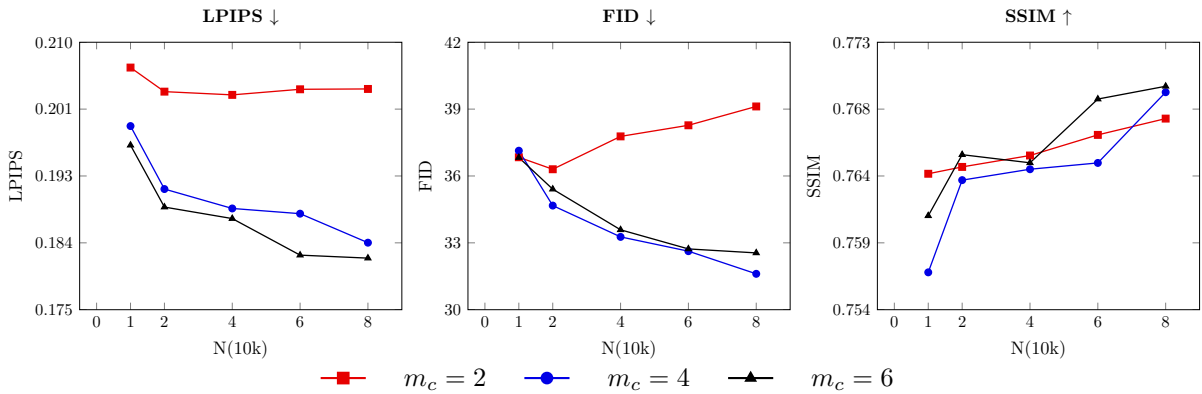
\begin{figure}[ht]
\centering
% First plot: FID
\begin{tikzpicture}[scale=0.625] 
\begin{axis}[
    title={LPIPS $\downarrow$},
    xlabel={N(10k)},
    ylabel={LPIPS},
    xmin=-0.5, xmax=9,
    ymin=0.175, ymax=0.21,
    xtick={0,1,2,4,6,8},
    ytick={0.175,
0.18375,
0.1925,
0.20125,
0.21},
    tick style={thin},
    ticklabel style={font=\footnotesize},
    label style={font=\small},
    title style={font=\small\bfseries},
    every axis plot/.append style={thick},
    scaled y ticks=false,
    y tick label style={
        /pgf/number format/fixed,
        /pgf/number format/.cd,
        fixed zerofill,
        precision=3
    },
]
\addplot[
color=red!90!black,
mark=square*]
    coordinates {
% (1,0.204444226873696)(2,0.20426140752635)(4,0.203709841467266)(6,0.204561444158222)(8,0.202875350546638)
(1,0.206687147805333)(2,0.203536697122451)(4,0.203114134457245)(6,0.203843867966293)(8,0.203889965569133)
    };
\addplot[
color=blue!90!black,
mark=*]
    coordinates {
% (1,0.197007751694161)(2,0.190384240490128)(4,0.189211036528103)(6,0.185430704836687)(8,0.180680777312113)
(1,0.199032644866739)(2,0.190789490111091)(4,0.1882493406364)(6,0.18756743731975)(8,0.183770098627456)

    };
\addplot[
color=black!90!black,
mark=triangle*]
    coordinates {
% (1,0.198027372049695)(2,0.189731446073069)(4,0.186036358733846)(6,0.181725956253985)(8,0.182262927828254)
(1,0.19652792012664)(2,0.188430707209359)(4,0.186926089860958)(6,0.182141466162723)(8,0.181750334667163)
    };
\end{axis}
\end{tikzpicture}
\begin{tikzpicture}[scale=0.625] 
\begin{axis}[
    title={FID $\downarrow$},
    xlabel={N(10k)},
    ylabel={FID},
    xmin=-0.5, xmax=9,
    ymin=30, ymax=42,
    xtick={0,1,2,4,6,8},
    ytick={30,
33,
36,
39,
42},
    tick style={thin},
    ticklabel style={font=\footnotesize},
    label style={font=\small},
    title style={font=\small\bfseries},
    every axis plot/.append style={thick},
    scaled y ticks=false,
    y tick label style={
        /pgf/number format/fixed,
        /pgf/number format/.cd,
        fixed zerofill,
        precision=0
    },
]
\addplot[
color=red!90!black,
mark=square*]
    coordinates {
% (1,37.407566070556)(2,37.9107322692871)(4,39.8219528198242)(6,38.9214172363281)(8,39.8368835449218)    
(1,36.8398780822754)(2,36.3014297485352)(4,37.7761154174805)(6,38.2738494873047)(8,39.1189498901367)
    };
    
\addplot[
    color=blue!90!black,
mark=*]
    coordinates {
% (1,37.5320053100585)(2,35.7544364929199)(4,35.512802124023)(6,34.2354927062988)(8,33.3583068847656)    
(1,37.1333122253418)(2,34.6719398498535)(4,33.2667541503906)(6,32.6258773803711)(8,31.6101951599121)
    };
    
\addplot[
    color=black!90!black,
mark=triangle*]
    coordinates {
% (1,38.8259010314941)(2,35.4560890197753)(4,35.5620956420898)(6,33.3512954711914)(8,34.30660629272)        
(1,36.8278541564941)(2,35.4087448120117)(4,33.5848503112793)(6,32.7270469665527)(8,32.5483169555664)
    };
\end{axis}
\end{tikzpicture}
% Third plot: SSIM
\begin{tikzpicture}[scale=0.625] 
\begin{axis}[
    title={SSIM $\uparrow$},
    xlabel={N(10k)},
    ylabel={SSIM},
    xmin=-0.5, xmax=9,
    ymin=0.754, ymax=0.773,
    xtick={0,1,2,4,6,8},
    ytick={0.754,
0.75875,
0.7635,
0.76825,
0.773},
    tick style={thin},
    ticklabel style={font=\footnotesize},
    label style={font=\small},
    title style={font=\small\bfseries},
    every axis plot/.append style={thick},
    scaled y ticks=false,
    y tick label style={
        /pgf/number format/fixed,
        /pgf/number format/.cd,
        fixed zerofill,
        precision=3
    },
]
\addplot[
color=red!90!black,
mark=square*]
    coordinates {
    
    % (1,0.762512816251512)(2,0.762559523839007)(4,0.766118153674415)(6,0.766348422499269)(8,0.766960099517146)
    (1,0.763657706998033)(2,0.76414410134688)(4,0.764954008471809)(6,0.766418919106092)(8,0.767575750799077)
    
    };
    
\addplot[
color=blue!90!black,
mark=*]
    coordinates {
    % (1,0.758368478654059)(2,0.760625819932505)(4,0.762801910184832)(6,0.764353233400846)(8,0.76975420416864)    
    (1,0.756653852303202)(2,0.763204907095393)(4,0.763978838693665)(6,0.764426669858522)(8,0.769449799658719)
    };
    
\addplot[
color=black!90!black,
mark=triangle*]
    coordinates {
    % (1,0.75963773651933)(2,0.762467815741999)(4,0.766488834010755)(6,0.767790752426558)(8,0.768462364258693)        
    (1,0.760673500017465)(2,0.765012386984367)(4,0.764435814659792)(6,0.768964101861897)(8,0.769872405320964)
    };
\end{axis}
\end{tikzpicture}
\vspace*{-18pt} 
\begin{center}
\footnotesize
    \begin{tikzpicture}
        \matrix[
            matrix of nodes, 
            anchor=center,   
            nodes={inner sep=2pt, font=\footnotesize}, 
            column sep=10pt, 
            row sep=0pt,     
        ]  {
            \ref{plot:n200} & $m_c=2$ & 
            \ref{plot:n300} & $m_c=4$ &
            \ref{plot:n400} & $m_c=6$ \\
        };
    \end{tikzpicture}
\end{center}
\vspace*{-12pt} 
\caption{\footnotesize Quantitative results on the CelebA dataset for varying $m_c$ ($n=400$, cLSDM).}
\label{CelebA:figure2}
\end{figure}

\begin{figure}[ht]
\centering
\includegraphics[scale=0.55]{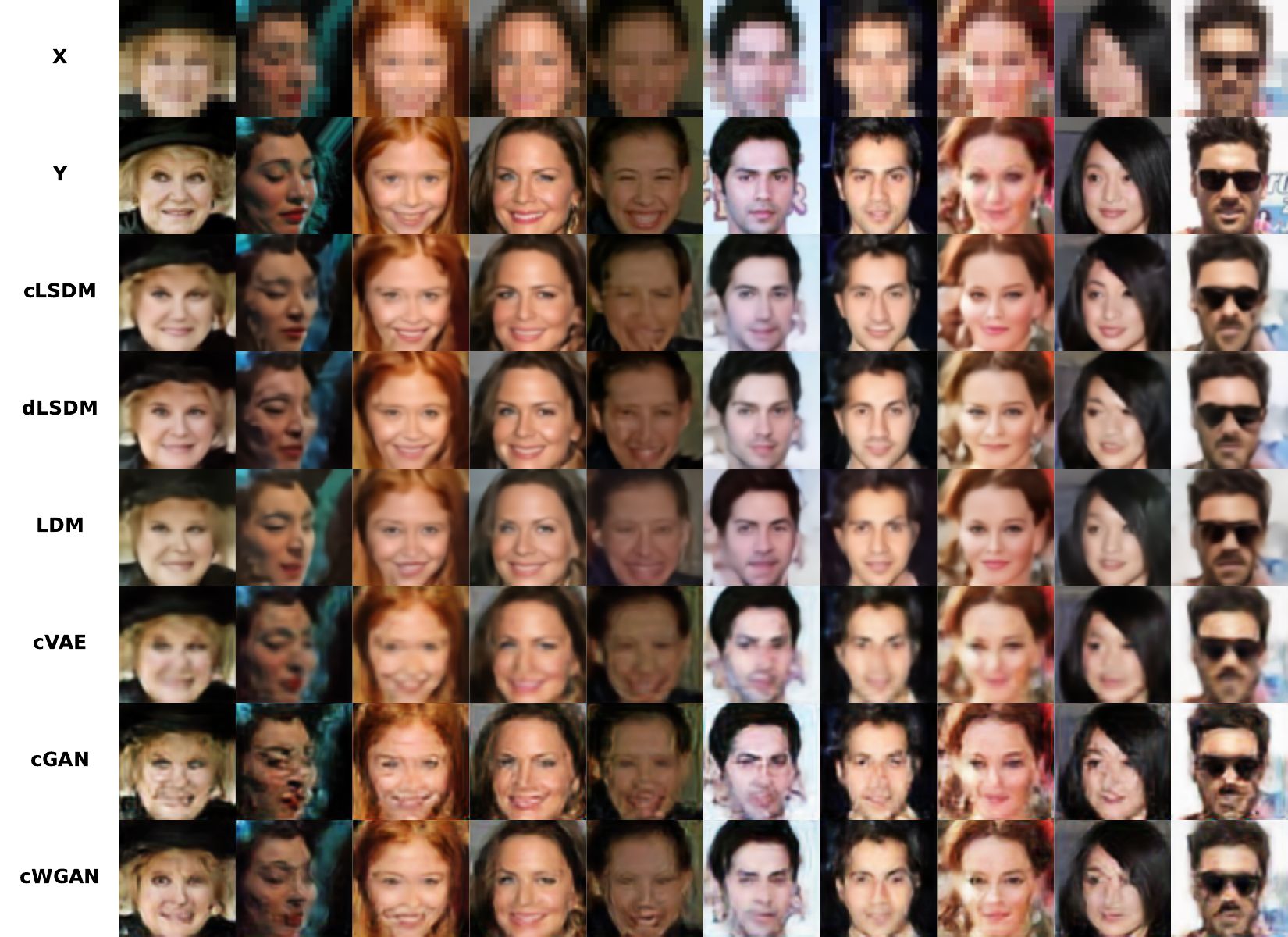}
\small
\vspace{0.05in}
\caption{Qualitative result on the CelebA dataset ($n=400$, $N=80{,}000$, $m_c=4$). }\label{CelebA:Qualitative}
\end{figure}

Figure~\ref{CelebA:figure2} shows results with $n = 400$ while varying the latent channel dimension $m_c$ and unpaired size $N$. Generation quality drops sharply when $m_c$ is too small (e.g., $m_c = 2$), confirming that $m$ must exceed $d_{\mathcal{Y}}$ for consistency in Step~1. 
Interestingly, for $m_c = 2$, the FID score worsens as $N$ increases. A possible explanation is that, with insufficient latent dimension and a large unpaired sample, realistic image points in latent space become densely clustered. As a result, the latent generator is more likely to converge to interpolated points that correspond to blurry images, degrading perceptual quality. This effect disappears once $m_c$ is sufficiently large.
Conversely, an excessively large channel count ($m_c = 6$) provides limited benefit and can even degrade performance (in FID).

\FloatBarrier
\begin{table}[ht]
\centering
\small
\setlength{\tabcolsep}{5pt}
\renewcommand{\arraystretch}{0.6}
\begin{minipage}{0.85\linewidth}
\centering
\begin{tabular}{llclllllll}
\hline
  &\multicolumn{3}{c}{FID$\downarrow$}& \multicolumn{3}{c}{LPIPS$\downarrow$} & \multicolumn{3}{c}{SSIM$\uparrow$}\\
\hline
Models &n=200&n=300&n=400&n=200& n=300& n=400& n=200& n=300&n=400\\
\hline
cLSDM &35.0&33.1&31.6& 0.187& 0.184&  0.181& 0.770& 0.769&0.770\\
dLSDM &34.4&32.4&31.3& 0.191& 0.186&  0.184& 0.769& 0.771&0.770\\
LDM&35.4&34.9&33.3& 0.198& 0.190&  0.179& 0.737& 0.746&0.758\\
cVAE &54.7&44.7&41.4& 0.245& 0.224&  0.215& 0.739& 0.751&0.757\\
cGAN &83.9& 71.9& 56.2& 0.298& 0.268&  0.240& 0.615& 0.663&0.689\\
  cWGAN &44.3& 41.8& 37.2&  0.243& 0.229&  0.217& 0.689& 0.702&0.712\\
\hline
\end{tabular}
\end{minipage}
\caption{Quantitative Result on CelebA test set of $19{,}962$ ($n+N=80{,}000$, $m_c=4$).}
\label{CelebA:quantitative}
\end{table}

\FloatBarrier

\section{Discussions} \label{Section:Discussion}
This paper introduced LSDM, a novel semi‑supervised framework for conditional generative modeling that combines joint distribution matching with latent space learning. LSDM unifies two major paradigms: the fast single-pass generation characteristic of GAN-like methods, and the ability of latent space models (e.g., latent diffusion models) to exploit unpaired data. We derived finite sample convergence rates that characterize how generation quality depends on key design factors, including latent dimension, smoothness parameters, and sample sizes, and showed that unpaired response data enhances approximation of the true data support, thereby improving perceptual quality. Experiments on class‑conditional generation and image super‑resolution validated the theoretical insights and showed that LSDM achieves strong performance with limited paired data.

Several directions warrant further investigation. First, our analysis assumes that the unpaired response data $Y$ share the same distribution as the responses in the paired dataset. In practice, distributional shifts may arise. Investigating the effects of such shifts and devising robust variants of LSDM would increase its practical utility. Second, although this study concentrated on unpaired response data $Y$, many semi‑supervised scenarios feature plentiful unpaired predictors $X$. Extending LSDM to leverage unlabeled $X$ could further enhance its applicability and impact.

{\bf \noindent Data Availability Statement}

The data supporting this study are publicly available from the following sources: MNIST is accessible at https://archive.ics.uci.edu/dataset/683/mnist+database+of+handwritten+digits, and CelebA data can be found at https://mmlab.ie.cuhk.edu.hk/projects/CelebA.html.

\bibliographystyle{apalike}
% \nocite{*}
\bibliography{reference}

@inproceedings{ChakrabortyBarlett2024,
  author       = {Saptarshi Chakraborty and
                  Peter L. Bartlett},
  title        = {A Statistical Analysis of Wasserstein Autoencoders for Intrinsically
                  Low-dimensional Data},
  booktitle    = {The Twelfth International Conference on Learning Representations,
                  {ICLR} 2024, Vienna, Austria, May 7-11, 2024},
  year         = {2024},
  url          = {https://openreview.net/forum?id=WjRPZsfeBO},
  bibsource    = {dblp computer science bibliography, https://dblp.org}
}

@misc{Pope2021,
      title={The Intrinsic Dimension of Images and Its Impact on Learning}, 
      author={Phillip Pope and Chen Zhu and Ahmed Abdelkader and Micah Goldblum and Tom Goldstein},
      year={2021},
      eprint={2104.08894},
      archivePrefix={arXiv},
      primaryClass={cs.CV},
      url={https://arxiv.org/abs/2104.08894}, 
}

@article{FID,
  title={Gans trained by a two time-scale update rule converge to a local nash equilibrium},
  author={Heusel, Martin and Ramsauer, Hubert and Unterthiner, Thomas and Nessler, Bernhard and Hochreiter, Sepp},
  journal={Advances in neural information processing systems},
  volume={30},
  year={2017}
}

@article{OptimalSafe2024,
	author = {Siyi Deng and Yang Ning and Jiwei Zhao and Heping Zhang},
	doi = {10.1080/01621459.2023.2277409},
	eprint = {https://doi.org/10.1080/01621459.2023.2277409},
	journal = {Journal of the American Statistical Association},
	number = {0},
	pages = {1--12},
	publisher = {ASA Website},
	title = {Optimal and Safe Estimation for High-Dimensional Semi-Supervised Learning},
	url = {https://doi.org/10.1080/01621459.2023.2277409},
	volume = {0},
	year = {2024}}

@article{SemiSupLinReg2022,
	author = {David Azriel and Lawrence D. Brown and Michael Sklar and Richard Berk and Andreas Buja and Linda Zhao},
	doi = {10.1080/01621459.2021.1915320},
	eprint = {https://doi.org/10.1080/01621459.2021.1915320},
	journal = {Journal of the American Statistical Association},
	number = {540},
	pages = {2238--2251},
	publisher = {ASA Website},
	title = {Semi-Supervised Linear Regression},
	url = {https://doi.org/10.1080/01621459.2021.1915320},
	volume = {117},
	year = {2022}}

@article{Buja2019,
	author = {Andreas Buja and Lawrence Brown and Richard Berk and Edward George and Emil Pitkin and Mikhail Traskin and Kai Zhang and Linda Zhao},
	doi = {10.1214/18-STS693},
	journal = {Statistical Science},
	number = {4},
	pages = {523 -- 544},
	publisher = {Institute of Mathematical Statistics},
	title = {{Models as Approximations I: Consequences Illustrated with Linear Regression}},
	url = {https://doi.org/10.1214/18-STS693},
	volume = {34},
	year = {2019}}

@article{ClusterAssump2006,
  title={Generalization Error Bounds in Semi-supervised Classification Under the Cluster Assumption.},
  author={Rigollet, Philippe},
  journal={Journal of Machine Learning Research},
  volume={8},
  number={7},
  year={2007}
}

@article{ManifoldAssump2004,
  title={Semi-Supervised Learning on Riemannian Manifolds},
  author={Mikhail Belkin and Partha Niyogi},
  journal={Machine Learning},
  year={2004},
  volume={56},
  pages={209-239},
  url={https://api.semanticscholar.org/CorpusID:17133491}
}

@inproceedings{lowdenAssump2005,
  title={Semi-Supervised Classification by Low Density Separation},
  author={Olivier Chapelle and Alexander Zien},
  booktitle={International Conference on Artificial Intelligence and Statistics},
  year={2005},
  url={https://api.semanticscholar.org/CorpusID:14283441}
}

@article{GeneralMEstSS2024,
	author = {Shanshan Song and Yuanyuan Lin and Yong Zhou},
	doi = {10.1080/01621459.2023.2169699},
	eprint = {https://doi.org/10.1080/01621459.2023.2169699},
	journal = {Journal of the American Statistical Association},
	number = {546},
	pages = {1065--1075},
	publisher = {ASA Website},
	title = {A General M-estimation Theory in Semi-Supervised Framework},
	url = {https://doi.org/10.1080/01621459.2023.2169699},
	volume = {119},
	year = {2024}}

@article{VAE,
  title={Auto-encoding variational bayes},
  author={Kingma, Diederik P},
  journal={arXiv preprint arXiv:1312.6114},
  year={2013}
}

@article{GAN,
  title={Generative adversarial nets},
  author={Goodfellow, Ian and Pouget-Abadie, Jean and Mirza, Mehdi and Xu, Bing and Warde-Farley, David and Ozair, Sherjil and Courville, Aaron and Bengio, Yoshua},
  journal={Advances in neural information processing systems},
  volume={27},
  year={2014}
}

@article{DDPM2020,
  title={Denoising diffusion probabilistic models},
  author={Ho, Jonathan and Jain, Ajay and Abbeel, Pieter},
  journal={Advances in neural information processing systems},
  volume={33},
  pages={6840--6851},
  year={2020}
}

@article{iWGAN,
  title={Inferential Wasserstein generative adversarial networks},
  author={Chen, Yao and Gao, Qingyi and Wang, Xiao},
  journal={Journal of the Royal Statistical Society Series B: Statistical Methodology},
  volume={84},
  number={1},
  pages={83--113},
  year={2022},
  publisher={Oxford University Press}
}

@article{AdaptiveGAN,
	author = {Yixuan Qiu and Qingyi Gao and Xiao Wang},
	doi = {10.1080/01621459.2024.2408778},
	eprint = {https://doi.org/10.1080/01621459.2024.2408778},
	journal = {Journal of the American Statistical Association},
	number = {0},
	pages = {1--13},
	publisher = {ASA Website},
	title = {Adaptive Learning of the Latent Space of Wasserstein Generative Adversarial Networks},
	url = {https://doi.org/10.1080/01621459.2024.2408778},
	volume = {0},
	year = {2024}}

@article{cGAN,
  title={Conditional Generative Adversarial Nets},
  author={Mehdi Mirza and Simon Osindero},
  journal={ArXiv},
  year={2014},
  volume={abs/1411.1784},
  url={https://api.semanticscholar.org/CorpusID:12803511}
}

@article{DCG,
	author = {Xingyu Zhou and Yuling Jiao and Jin Liu and Jian Huang},
	doi = {10.1080/01621459.2021.2016424},
	eprint = {https://doi.org/10.1080/01621459.2021.2016424},
	journal = {Journal of the American Statistical Association},
	number = {543},
	pages = {1837--1848},
	publisher = {ASA Website},
	title = {A Deep Generative Approach to Conditional Sampling},
	url = {https://doi.org/10.1080/01621459.2021.2016424},
	volume = {118},
	year = {2023}}

@ARTICLE{ImgSupRes2,
  author={Saharia, Chitwan and Ho, Jonathan and Chan, William and Salimans, Tim and Fleet, David J. and Norouzi, Mohammad},
  journal={IEEE Transactions on Pattern Analysis and Machine Intelligence}, 
  title={Image Super-Resolution via Iterative Refinement}, 
  year={2023},
  volume={45},
  number={4},
  pages={4713-4726},
  doi={10.1109/TPAMI.2022.3204461}}

@inproceedings{ImgSupRes1,
  title={Photo-realistic single image super-resolution using a generative adversarial network},
  author={Ledig, Christian and Theis, Lucas and Husz{\'a}r, Ferenc and Caballero, Jose and Cunningham, Andrew and Acosta, Alejandro and Aitken, Andrew and Tejani, Alykhan and Totz, Johannes and Wang, Zehan and others},
  booktitle={Proceedings of the IEEE conference on computer vision and pattern recognition},
  pages={4681--4690},
  year={2017}
}

@INPROCEEDINGS{DomainShift,
  author={Wang, Wei and Zhang, Haochen and Yuan, Zehuan and Wang, Changhu},
  booktitle={2021 IEEE/CVF International Conference on Computer Vision (ICCV)}, 
  title={Unsupervised Real-World Super-Resolution: A Domain Adaptation Perspective}, 
  year={2021},
  volume={},
  number={},
  pages={4298-4307},
  doi={10.1109/ICCV48922.2021.00428}}

@article{WassDCG,
  title={Wasserstein generative learning of conditional distribution},
  author={Liu, Shiao and Zhou, Xingyu and Jiao, Yuling and Huang, Jian},
  journal={arXiv preprint arXiv:2112.10039},
  year={2021}
}

@article{Cai1,
  title={Efficient and adaptive linear regression in semi-supervised settings},
  author={Abhishek Chakrabortty and Tianxi Cai},
  journal={The Annals of Statistics},
  year={2017},
  url={https://api.semanticscholar.org/CorpusID:88520809}
}

@article{Cai2,
  title={Semi-supervised inference: General theory and estimation of means},
  author={Anru R. Zhang and Lawrence D. Brown and T. Tony Cai},
  journal={The Annals of Statistics},
  year={2016},
  url={https://api.semanticscholar.org/CorpusID:23696481}
}

@article{Cai3,
  title={Semisupervised inference for explained variance in high dimensional linear regression and its applications},
  author={T. Tony Cai and Guo, Zijian},
  journal={Journal of the Royal Statistical Society Series B: Statistical Methodology},
  volume={82},
  number={2},
  pages={391--419},
  year={2020},
  publisher={Oxford University Press}
}

@article{TemperFlow,
	author = {Yixuan Qiu and Xiao Wang},
	doi = {10.1080/01621459.2023.2198059},
	eprint = {https://doi.org/10.1080/01621459.2023.2198059},
	journal = {Journal of the American Statistical Association},
	number = {546},
	pages = {1446--1460},
	publisher = {ASA Website},
	title = {Efficient Multimodal Sampling via Tempered Distribution Flow},
	url = {https://doi.org/10.1080/01621459.2023.2198059},
	volume = {119},
	year = {2024}}

@article{GANError,
  title={An error analysis of generative adversarial networks for learning distributions},
  author={Huang, Jian and Jiao, Yuling and Li, Zhen and Liu, Shiao and Wang, Yang and Yang, Yunfei},
  journal={Journal of machine learning research},
  volume={23},
  number={116},
  pages={1--43},
  year={2022}
}

@article{WAE,
  title={Wasserstein auto-encoders},
  author={Tolstikhin, Ilya and Bousquet, Olivier and Gelly, Sylvain and Schoelkopf, Bernhard},
  journal={arXiv preprint arXiv:1711.01558},
  year={2017}
}

@book{Kallenberg,
  title={Foundations of modern probability},
  author={Kallenberg, Olav},
  volume={2},
  year={1997},
  publisher={Springer}
}

@book{villani2008,
  title={Optimal transport: old and new},
  author={Villani, C{\'e}dric and others},
  volume={338},
  year={2008},
  publisher={Springer}
}

@article{WGANGP,
  title={Improved training of wasserstein gans},
  author={Gulrajani, Ishaan and Ahmed, Faruk and Arjovsky, Martin and Dumoulin, Vincent and Courville, Aaron C},
  journal={Advances in neural information processing systems},
  volume={30},
  year={2017}
}

@article{ADAM,
  title={Adam: A method for stochastic optimization},
  author={Kingma, Diederik P},
  journal={arXiv preprint arXiv:1412.6980},
  year={2014}
}

@article{kolmogorov1961,
	author = {Kolmogorov, Andrey N. and Tikhomirov, Vladimir M.},
	journal = {American Mathematical Society Translations: Series 2},
	pages = {277--364},
	title = {$\epsilon$-entropy and $\epsilon$-capacity of sets in functional spaces},
	volume = {17},
	year = {1961}}

@article{lu2020,
  title={A universal approximation theorem of deep neural networks for expressing probability distributions},
  author={Lu, Yulong and Lu, Jianfeng},
  journal={Advances in neural information processing systems},
  volume={33},
  pages={3094--3105},
  year={2020}
}

@article{GenerativePretraining2010,
  author  = {Dumitru Erhan and Yoshua Bengio and Aaron Courville and Pierre-Antoine Manzagol and Pascal Vincent and Samy Bengio},
  title   = {Why Does Unsupervised Pre-training Help Deep Learning?},
  journal = {Journal of Machine Learning Research},
  year    = {2010},
  volume  = {11},
  number  = {19},
  pages   = {625--660},
  url     = {http://jmlr.org/papers/v11/erhan10a.html}
}

@inproceedings{CelebA,
  title = {Deep Learning Face Attributes in the Wild},
  author = {Liu, Ziwei and Luo, Ping and Wang, Xiaogang and Tang, Xiaoou},
  booktitle = {Proceedings of International Conference on Computer Vision (ICCV)},
  month = {December},
  year = {2015} 
}

@inproceedings{Pythae,
 author = {Chadebec, Cl\'{e}ment and Vincent, Louis and Allassonniere, Stephanie},
 booktitle = {Advances in Neural Information Processing Systems},
 editor = {S. Koyejo and S. Mohamed and A. Agarwal and D. Belgrave and K. Cho and A. Oh},
 pages = {21575--21589},
 publisher = {Curran Associates, Inc.},
 title = {Pythae: Unifying Generative Autoencoders in Python - A Benchmarking Use Case},
 volume = {35},
 year = {2022}
}

@inproceedings{LDM,
  title={High-resolution image synthesis with latent diffusion models},
  author={Rombach, Robin and Blattmann, Andreas and Lorenz, Dominik and Esser, Patrick and Ommer, Bj{\"o}rn},
  booktitle={Proceedings of the IEEE/CVF conference on computer vision and pattern recognition},
  pages={10684--10695},
  year={2022}
}

@article{Umap,
  title={Umap: Uniform manifold approximation and projection for dimension reduction},
  author={McInnes, Leland and Healy, John and Melville, James},
  journal={arXiv preprint arXiv:1802.03426},
  year={2018}
}

@article{MetricBounds,
  title={On choosing and bounding probability metrics},
  author={Gibbs, Alison L and Su, Francis Edward},
  journal={International statistical review},
  volume={70},
  number={3},
  pages={419--435},
  year={2002},
  publisher={Wiley Online Library}
}

@article{zhang2019semi,
  title={Semi-supervised Inference: General Theory and Estimation of Means.},
  author={Zhang, Anru and Brown, Lawrence D and Cai, T Tony},
  journal={The Annals of Statistics},
  volume={47},
  number={5},
  pages={2538--2566},
  year={2019},
  publisher={JSTOR}
}

@article{W2_clustering,
  title={Wasserstein $ K $-means for clustering probability distributions},
  author={Zhuang, Yubo and Chen, Xiaohui and Yang, Yun},
  journal={Advances in Neural Information Processing Systems},
  volume={35},
  pages={11382--11395},
  year={2022}
}

@inproceedings{W2_domainadaptation,
  title={Wasserstein distance guided representation learning for domain adaptation},
  author={Shen, Jian and Qu, Yanru and Zhang, Weinan and Yu, Yong},
  booktitle={Proceedings of the AAAI conference on artificial intelligence},
  volume={32},
  year={2018}
}

@article{WGR,
  title={Wasserstein generative regression},
  author={Song, Shanshan and Wang, Tong and Shen, Guohao and Lin, Yuanyuan and Huang, Jian},
  journal={arXiv preprint arXiv:2306.15163},
  year={2023}
}

@article{FlowMatchingLatent,
  title={Flow matching in latent space},
  author={Dao, Quan and Phung, Hao and Nguyen, Binh and Tran, Anh},
  journal={arXiv preprint arXiv:2307.08698},
  year={2023}
}

@article{fGAN,
  title={f-gan: Training generative neural samplers using variational divergence minimization},
  author={Nowozin, Sebastian and Cseke, Botond and Tomioka, Ryota},
  journal={Advances in neural information processing systems},
  volume={29},
  year={2016}
}

@article{ConditionalDiffusionConvergence,
  title={Unveil conditional diffusion models with classifier-free guidance: A sharp statistical theory},
  author={Fu, Hengyu and Yang, Zhuoran and Wang, Mengdi and Chen, Minshuo},
  journal={arXiv preprint arXiv:2403.11968},
  year={2024}
}

@article{VQVAE,
  title={Neural discrete representation learning},
  author={Van Den Oord, Aaron and Vinyals, Oriol and others},
  journal={Advances in neural information processing systems},
  volume={30},
  year={2017}
}

@article{OverviewDM,
  title={An overview of diffusion models: Applications, guided generation, statistical rates and optimization},
  author={Chen, Minshuo and Mei, Song and Fan, Jianqing and Wang, Mengdi},
  journal={arXiv preprint arXiv:2404.07771},
  year={2024}
}

@article{cW1Inequality,
  title={Conditional Wasserstein distances with applications in Bayesian OT flow matching},
  author={Chemseddine, Jannis and Hagemann, Paul and Steidl, Gabriele and Wald, Christian},
  journal={Journal of Machine Learning Research},
  volume={26},
  number={141},
  pages={1--47},
  year={2025}
}

@book{Bakry,
  title={Analysis and Geometry of Markov Diffusion Operators},
  author={Bakry, D. and Gentil, I. and Ledoux, M.},
  isbn={9783319002279},
  lccn={2013952461},
  series={Grundlehren der mathematischen Wissenschaften},
  url={https://books.google.com.hk/books?id=gU3ABAAAQBAJ},
  year={2013},
  publisher={Springer International Publishing}
}

@inproceedings{cVAE,
 author = {Sohn, Kihyuk and Lee, Honglak and Yan, Xinchen},
 booktitle = {Advances in Neural Information Processing Systems},
 editor = {C. Cortes and N. Lawrence and D. Lee and M. Sugiyama and R. Garnett},
 pages = {},
 publisher = {Curran Associates, Inc.},
 title = {Learning Structured Output Representation using Deep Conditional Generative Models},
 url = {https://proceedings.neurips.cc/paper_files/paper/2015/file/8d55a249e6baa5c06772297520da2051-Paper.pdf},
 volume = {28},
 year = {2015}
}

@article{Euler-Maruyama,
author = {Gelbrich, Matthias and R\"{o}misch, Werner},
title = {Numerical Solution of Stochastic Differential Equations (Peter E. Kloeden and Eckhard Platen)},
journal = {SIAM Review},
volume = {37},
number = {2},
pages = {272-275},
year = {1995},
doi = {10.1137/1037073}
}

@inproceedings{LPIPS,
  title={The unreasonable effectiveness of deep features as a perceptual metric},
  author={Zhang, Richard and Isola, Phillip and Efros, Alexei A and Shechtman, Eli and Wang, Oliver},
  booktitle={Proceedings of the IEEE conference on computer vision and pattern recognition},
  pages={586--595},
  year={2018}
}

@article{VGG,
  title={Very deep convolutional networks for large-scale image recognition},
  author={Simonyan, Karen and Zisserman, Andrew},
  journal={arXiv preprint arXiv:1409.1556},
  year={2014}
}

@inproceedings{VQGAN,
  title={Taming transformers for high-resolution image synthesis},
  author={Esser, Patrick and Rombach, Robin and Ommer, Bjorn},
  booktitle={Proceedings of the IEEE/CVF conference on computer vision and pattern recognition},
  pages={12873--12883},
  year={2021}
}

@article{LogSoblev,
  title={Rapid convergence of the unadjusted langevin algorithm: Isoperimetry suffices},
  author={Vempala, Santosh and Wibisono, Andre},
  journal={Advances in neural information processing systems},
  volume={32},
  year={2019}
}

@article{MaxCoupling,
issn = {0002-9947},
journal = {Transactions of the American Mathematical Society},
language = {eng},
number = {12},
copyright = {Copyright 2019 by the authors},
pages = {8307-8345},
publisher = {American Mathematical Society},
title = {Mixing time and eigenvalues of the abelian sandpile Markov chain},
volume = {372},
year = {2019},
author = {Jerison, Daniel C. and Levine, Lionel and Pike, John},
address = {Providence, Rhode Island},
}

\end{document}

\def\spacingset#1{\renewcommand{\baselinestretch}%
{#1}\small\normalsize} \spacingset{1}

%%%%%%%%%%%%%%%%%%%%%%%%%%%%%%%%%%%%%%%%%%%%%%%%%%%%%%%%%%%%%%%%%%%%%%%%%%%%%%

\if1\blind
{
  \title{\bf Supplementary Material to ``Semi-Supervised Generative Learning via Latent Space Distribution Matching''}
  \author{}
    \date{}
  \maketitle
} \fi

\if0\blind
{
  \bigskip
  \bigskip
  \bigskip
  \begin{center}
    % {\LARGE\bf Semi-supervised Generative Learning via Autoencoding Wasserstein Generator}
        {\LARGE\bf Supplementary Material to ``Semi-Supervised Generative Learning via Latent Space Distribution Matching''}
\end{center}
  \medskip
} \fi

\bigskip

\spacingset{1.9} % DON'T change the spacing!
\appendix

In the supplementary material, we provide proofs for the Theorems along with several additional lemmas and propositions. Furthermore, we include more detailed discussions on the connection of LSDM to LDM, and information on the implementation of the numerical experiments.
\section{Definitions}

\begin{definition}[f-divergence] Let $P$ and $Q$ be two probability distributions on $\R^d$. Let $p$ and $q$ be the density function of $P$ and $Q$ with respect to a common dominant measure, respectively. Suppose $Q$ is absolutely continuous with respect to $P$, the f-divergence of $Q$ with respect to $P$ is defined by
    \begin{equation}
        \mathbb{D}_f\left(Q, P\right) = \int f\left(\frac{q(z)}{p(z)}\right) p(z)dz,
    \end{equation}
    where $f: [0,\infty) \to \R$ is a convex function with $f(1) = 0$ and is strictly convex at $x=1$. 

    (Kullback–Leibler divergence) The Kullback–Leibler (KL) divergence is a special case of f-divergence with $f(x) = x\log x$.

    (Jensen-Shannon divergence) The Jensen-Shannon (JS) divergence is a special case of f-divergence with $f(x) = \frac{1}{2} \left(x\log x - (x+1)\log\left(\frac{x+1}{2}\right)\right)$.

    ($\chi^2$ divergence) The $\chi^2$ divergence is a special case of f-divergence with $f(x) = \left(x-1\right)^2$.

    (Squared Hellinger divergence) The Squared Hellinger divergence is a special case of f-divergence with $f(x) =  \left(\sqrt{x}-1\right)^2$.
\end{definition}

\begin{definition} [Total variation distance] Let $(\Omega, \cA)$ be a measurable space and probability measures $P$ and $Q$ are defined on $(\Omega, \cA)$. The total variation distance between $P$ and $Q$ is defined as
    \begin{equation}
        \mathbb{D}_{TV}\left(P,Q\right) = \sup_{A \in \cA} \left| P(A)-Q(A)\right|.
    \end{equation}
    When the distributions have probability density function $p$ and $q$ respectively,
    \begin{equation}
         \mathbb{D}_{TV}\left(P,Q\right) = \frac{1}{2}\int \left| p(z) - q(z)\right| dz.
    \end{equation}
\end{definition}

\begin{definition}[Lévy–Prokhorov metric] Let $(\Omega,d)$ be a metric space with Borel sigma-algebra $\cB(\Omega)$. Let $\cP(\Omega)$ be the set of all probability measures on $(\Omega, \cB(\Omega))$. The Lévy–Prokhorov metric between $P$ and $Q$ in $\cP(\Omega)$ is
    \begin{equation}
        \mathbb{D}_{P}\left(P,Q\right) = \inf \{\epsilon > 0 | P(B)\leq Q(B^{\epsilon}) + \epsilon \text{ and }  Q(B)\leq P(B^{\epsilon}) + \epsilon \text{ for all } B \in \cB(\Omega)\},
    \end{equation}
    where $B^{\epsilon}:= \{p \in \Omega | \exists q \in B, d(p,q) < \epsilon\}.$
\end{definition}

\begin{definition}[Integral Probability Metrics, IPM] Let $\cF$ be a class of real-valued function, $\mu, \nu, \gamma$ be probability distributions over $\Omega$. The IPM between the two distributions $\mu, \nu$ are defined as 
    $$\dnorm{\mu - \nu}_\cF = \sup_{f \in \cF} |\E_{X \sim \mu} f(X) - \E_{Y \sim \nu}f(Y)| $$
    A function class $\cF$ is symmetric if $f \in \cF$ implies $-f \in \cF$. When $\cF$ is symmetric, the following holds
    \begin{enumerate}
        \item (Equivalent Definition) The absolute sign in IPM can be omitted
        \item (Symmetry) $\dnorm{\mu - \nu}_\cF = \dnorm{\nu - \mu}_\cF$
        \item (Triangle Inequality) $\dnorm{\mu - \nu }_\cF \leq \dnorm{\mu - \gamma }_\cF+\dnorm{\gamma - \nu }_\cF$
    \end{enumerate}
\end{definition}

\begin{definition}[$\epsilon$-cover] For a set $A \subseteq \R^k$, and $\epsilon>0$, we say that $C$ is an $\epsilon$-cover of $A$ w.r.t. $\dnorm{\cdot}_\infty$ norm if $C \subseteq A$ and for every $a \in A$ there is a $c \in C$ such that $\dnorm{c - a}_\infty < \epsilon$.
\end{definition}

\begin{definition}[Covering Number] For a set $A \subseteq \R^k$, and $\epsilon>0$, we define the covering number of $A$, denoted as $\cN(\epsilon, A, \dnorm{\cdot}_\infty)$, to be the minimal cardinality of an $\epsilon$-cover of $A$ w.r.t. $\dnorm{\cdot}_\infty$ norm. We say that $C$ is an optimal $\epsilon$-cover of $A$ if $C$ is an $\epsilon$-cover of $A$ and $|C| = \cN(\epsilon, A, \dnorm{\cdot}_\infty)$. 
\end{definition}

\begin{definition}[Minkowski dimension] For a bounded set $A \subset \R^d$, the upper Minkwoski dimension of $A$ is defined as 
    $$\overline{\dim}_M(A):= \limsup_{\epsilon\to 0} \frac{\log \cN(\epsilon,A, \dnorm{\cdot}_2)}{-\log \epsilon},$$
the lower Minkowski dimension of $A$ is defined as
$$\underline{\dim}_M(A):= \liminf_{\epsilon\to 0} \frac{\log \cN(\epsilon,A, \dnorm{\cdot}_2)}{-\log \epsilon},$$
where $\cN(\epsilon,A, \dnorm{\cdot}_2)$ is the covering number of $A$ w.r.t. $\dnorm{\cdot}_2$ norm. If $\overline{\dim}_M(A) = \underline{\dim}_M(A)$, then $\dim_M(A)$ is called the Minkowski dimension of the set $A$.
\end{definition}

\begin{definition}
    For a class of real-valued function $\cF: \cX \to \R$, and a sequence $X_{1:n}= (X_1,...,X_n) \in \cX^n$, we let
    $$\cF_{|X_{1:n}} = \{ \left(f(X_1), ..., f(X_n)\right): f\in \cF\} \subseteq \R^n$$
    For a class of vector-valued function $\cG: \cX \to \R^d$, we let 
    $$\cG_{|X_{1:n}} = \{ \left(g(X_1): ... : g(X_n)\right): g\in \cG\} \subseteq \R^{dn}$$
\end{definition}

\begin{definition}[SSIM]
The Structural Similarity Index Measure (SSIM) between two images $X$ and $Y$ with pixel values in $[0,1]$ is defined as
\begin{equation}
\operatorname{SSIM}(X,Y) = \frac{(2\mu_X\mu_Y + c_1)(2\sigma_{XY} + c_2)}{(\mu_X^2 + \mu_Y^2 + c_1)(\sigma_X^2 + \sigma_Y^2 + c_2)},
\end{equation}
where $\mu_X$ and $\mu_Y$ are the mean intensities, $\sigma_X^2$ and $\sigma_Y^2$ are the variances, and $\sigma_{XY}$ is the covariance, all computed over a local window. The constants $c_1 = (k_1L)^2$ and $c_2 = (k_2L)^2$ stabilize the division, with $L=1$ (the dynamic range of pixel values) and $k_1=0.01$, $k_2=0.03$. In implementation, SSIM is typically averaged over multiple windows rather than computed globally.
\end{definition}

\begin{definition}[FID]
Let $d_F(\mu,\nu)$ denote the Fréchet distance between two probability measures $\mu$ and $\nu$ on $\mathbb{R}^n$:
\begin{equation}
d_F(\mu,\nu) = \left( \inf_{\gamma \in \Gamma(\mu,\nu)} \int_{\mathbb{R}^n \times \mathbb{R}^n} \|x-y\|_2^2 \, d\gamma(x,y) \right)^{\frac{1}{2}},
\end{equation}
where $\Gamma(\mu,\nu)$ is the set of all couplings with marginals $\mu$ and $\nu$. Let $f$ be an Inception v3 model pretrained on ImageNet, with its final classification layer removed. The Fréchet Inception Distance (FID) between two image distributions $\Prob_X$ and $\Prob_Y$ is defined as
\begin{equation}
\operatorname{FID}(\Prob_X, \Prob_Y) = d_F\bigl(f_\#\Prob_X, f_\#\Prob_Y\bigr),
\end{equation}
where $f_\#\Prob$ denotes the pushforward distribution of $\Prob$ under the feature map $f$. 
\end{definition}

\begin{definition}[LPIPS]
Let $\mathcal{X} \subseteq \mathbb{R}^{H \times W \times 3}$ denote the space of RGB images. Let $\{\Phi^l\}_{l=1}^{L}$ be a collection of feature maps obtained from the intermediate layers of a pre‑trained convolutional neural network, where $\Phi^l: \mathcal{X} \to \mathbb{R}^{H_l \times W_l \times C_l}$. Each feature map is normalized such that for every spatial location $(h,w)$, the $C_l$-dimensional feature vector has unit $\ell_2$ norm.

For a pair of images $(x, x_0) \in \mathcal{X} \times \mathcal{X}$, define the layer‑wise squared perceptual discrepancy as
\begin{equation}
d^l(x, x_0) = \frac{1}{H_l W_l} \sum_{h=1}^{H_l} \sum_{w=1}^{W_l} \sum_{c=1}^{C_l} \gamma_c^l \bigl( \Phi^l_{h,w,c}(x) - \Phi^l_{h,w,c}(x_0) \bigr)^2,
\end{equation}
where $\boldsymbol{\gamma}^l = (\gamma_1^l, \dots, \gamma_{C_l}^l) \in \mathbb{R}_{\geq 0}^{C_l}$ are non‑negative scaling parameters associated with layer $l$. The full parameter vector $\boldsymbol{\gamma} = \{\boldsymbol{\gamma}^l\}_{l=1}^{L}$ is estimated from a dataset of human perceptual judgments in \citet{LPIPS}. The LPIPS distance between $x$ and $x_0$ is then defined as
\begin{equation}
d_{\boldsymbol{\gamma}}(x, x_0) = \sum_{l=1}^{L} d^l(x, x_0).
\end{equation}
\end{definition}

\section{Additional Discussions}

\subsection{Connections and Comparisons of LSDM with LDMs}

\paragraph{Step 1: Autoencoder training.}
In the original work of LDMs \citep{LDM}, Step 1 trains a deterministic autoencoder with MSE loss plus a vector‑quantization loss and a GAN loss:
\begin{align*}
\mathcal{L}_{\text{VQ}} &= \mathbb{E}_Y \big\| \mathrm{sg}[E(Y)] - q(E(Y)) \big\|_2^2 + \beta \, \mathbb{E}_Y \big\| E(Y) - \mathrm{sg}[q(E(Y))] \big\|_2^2, \\[4pt]
\mathcal{L}_{\text{GAN}} &= \max_\psi \; \mathbb{E}_Y \big[ \log f_\psi(Y) \big] + \mathbb{E}_Y \big[ \log\!\big(1 - f_\psi(D(E(Y)))\big) \big],
\end{align*}
where:
\begin{itemize}
    \item \(\mathcal{L}_{\text{VQ}}\) is the vector‑quantization loss, with \(\beta\) a commitment loss hyper‑parameter;
    \item \(\mathcal{L}_{\text{GAN}}\) is a GAN loss based on a learned discriminator \(f_\psi\) parameterized by \(\psi\);
    \item \(\mathrm{sg}[\cdot]\) denotes the stop‑gradient operator, which blocks gradient flow during backpropagation;
    \item \(q(z)\) returns the nearest embedding vector to \(z\) from a learned \(K\)-element codebook \(\{e_1,\dots,e_K\}\):
    \begin{equation*}
        q(z) = \min_{e \in \{e_1,\dots,e_K\}} \|z - e\|_2;
    \end{equation*}
    \item The quantization step is absorbed into the decoder, so that \(D(z) = \widetilde{D}(q(z))\) with a learned codebook decoder $\widetilde{D}$.
\end{itemize}
The full Step 1 objective in LDMs with VQ regularization is:
\begin{equation}
    \mathcal{L}_{\text{LDM}} = \mathbb{E}_Y \|Y - D \circ E(Y)\|_2^2 + R(E,D),
\end{equation}
where \(R(E,D) = \lambda \,  \mathcal{L}_{\text{GAN}} +  \mathcal{L}_{\text{VQ}}\), with \(\lambda\) a hyper‑parameter controlling the perceptual loss. Clearly, as \(\mathcal{L}_{\text{LDM}} \to 0\), we have \(\mathcal{L}_{\text{recon}} = \E_Y \|Y - D\circ E(Y)\|_2 \to 0\). Thus, autoencoder training in LDMs minimizes the Step 1 reconstruction objective in LSDM.

The connections of LSDM to LDM for Step 2 is discussed in the paper.

\section{Proofs of Main Results}
\subsection{Proof of Theorem \ref{Theorem:cLSDM}}
% ---------------------------------------------------------------

\textbf{Theorem \ref{Theorem:cLSDM}:} Let $E: \cY \to \cZ$ be an encoder. Suppose the generator $G$ has the form $G = D\circ H$, where $D: \cZ \to \R^q$ and $H: \cX \times \R^d \to \cZ$. Then, the 1-Wasserstein distance between the joint distributions of  $\left(X, G(X,\eta)\right)$ and $(X,Y)$ satisfies the bound:
\begin{equation}
\Wd{X, G(X, \eta)}{X,Y} \leq \E\dnorm{Y - D\circ E(Y)}_2 + \Wd{X, D\circ H(X, \eta)}{X, D\circ E(Y)},
\end{equation}
Consequently, if $(D,E,H)$ is a triplet that satisfies
\begin{equation}
    \E\dnorm{Y - D\circ E(Y)}_2=0,\quad 
    \Wd{X, D\circ H(X, \eta)}{X, D\circ E(Y)}=0,
\end{equation}
then the generator achieves conditional distribution matching:
\begin{align*}
    G(x,\eta) \sim \Prob_{Y|X=x}.
\end{align*}

\textbf{Proof:}
For any $(D,E)$,
\begin{align*}
    \Wd{X, Y}{X, D\circ E(Y)}&=  \inf_{\gamma' \in \Gamma(\Prob_{X,Y}, \Prob_{X, D\circ E(Y)})} \E_{(X_1,Y_1,X_2,Z) \sim \gamma'}\dnorm{(X_1,Y_1)-(X_2,Z)}_2\\
    &\leq\inf_{\gamma \in \Gamma\left(\Prob_{X,Y}, \Prob_{X,Y}\right)}\E_{(X_1,Y_1,X_2,Y_2)\sim \gamma}\dnorm{(X_1, Y_1) - (X_2, D\circ E(Y_2))}_2\\
    &\leq \E_{(X,Y)\sim\Prob_{X,Y}}\dnorm{(X, Y) - (X, D\circ E(Y))}_2\\
    &= \E_{Y \sim \Prob_Y}\dnorm{(Y- D\circ E(Y)}_2
\end{align*}
The first line is definition \ref{Definition:WassersteinDistance}.  The second line is due to the fact that, for any $(X_1,Y_1,X_2, Y_2) \sim \gamma \in \Gamma\left(\Prob_{X,Y}, \Prob_{X,Y}\right)$, $\left(X_1,Y_1,X_2, D\circ E(Y_2)\right)$ has a joint distribution with marginal distributions $\left(\Prob_{X,Y}, \Prob_{X,D\circ E(Y)}\right)$. The third line follows from the fact that $(X, Y, X, Y)$ has a joint distribution with marginals $\left(\Prob_{X,Y},\Prob_{X, Y}\right)$, and is thus an upper bound. By the triangle inequality of the 1-Wasserstein distance:
\begin{align*}
    \Wd{X, G(X, \eta)}{X,Y} &= \Wd{X, D\circ H(X, \eta)}{X,Y}\\
    &\leq \Wd{X, D\circ H(X, \eta)}{X, D\circ E(Y)} + \Wd{X, D\circ E(Y)}{X, Y}\\
    &\leq \Wd{X, D\circ H(X, \eta)}{X, D\circ E(Y)} + \E\dnorm{Y- D\circ E(Y)}_2
\end{align*}

Since the 1-Wasserstein distance is non-negative, if $\Wd{X, D\circ H(X, \eta)}{X, D\circ E(Y)}=0$ and $ \E\dnorm{Y- D\circ E(Y)}=0$, then $\Wd{X, G(X, \eta)}{X,Y} =0$. By lemma \ref{Lemma:WassersteinConvergence}, $(X,D\circ H(X,\eta)) \eqd (X,Y)$. Consequently, by lemma \ref{Lemma:JointDistributionMatching}, $G(x,\eta)\sim \Prob_{Y|X=x}, \hspace{0.5em}  G(x,\eta) \sim \Prob_{Y|X=x},$ for $\Prob_X$-almost every $x\in \cX$.

% ---------------------------------------------------------------
\subsection{Proof of Proposition \ref{Proposition:ExistenceOfH}}
\textbf{Proposition \ref{Proposition:ExistenceOfH}:}
For arbitrary $(\hD,\hE)$, there exists a measurable function $H: \cX \times \R^d \to \cZ$ such that $(X,  H(X,\eta)) = (X,  \hE(Y)) \hspace{0.5em} a.s.$ and $(X, \hD\circ H(X,\eta)) = (X, \hD\circ \hE(Y)) \hspace{0.5em} a.s.$ If they have finite first moment, then both $ \Wd{X, \hD\circ H(X, \eta)}{X,\hD\circ \hE(Y)} $ and $ \Wd{X,  H(X, \eta)}{X,\hE(Y)} $ are zero.

\hfill

\textbf{Proof:}
Given a pair $(\hD,\hE)$, consider the random vector $(X,\hE(Y))$. By Lemma \ref{Lemma:NnoiseOutsourcingLemma}, there exists a measurable function $H$ such that $\eta \sim \cN(0, I_d)$ is independent of $X$ and $(X, H(X,\eta)) = (X,\hE(Y)) \hspace{0.5em} a.s.$. Since $\hD$ is measurable, $(X,\hD\circ H(X,\eta)) = (X,\hD\circ \hE(Y)) \hspace{0.5em} a.s.$. Since they have finite first moment and the same distribution, by lemma \ref{Lemma:WassersteinConvergence}, $\Wd{X, \hD\circ H(X,\eta)}{X,\hD\circ \hE(Y)} = 0$ and $ \Wd{X,  H(X, \eta)}{X,\hE(Y)}  = 0$.

\subsection{Proof of Theorem \ref{Theorem:dLSDM}}
\textbf{Theorem \ref{Theorem:dLSDM}:}
 If the decoder $D$ is $K$-Lipschitz continuous, then
    \begin{equation}
        \Wd{X, D\circ H(X,\eta)}{X, D\circ E(Y)} \leq \left(1 \vee K\right) \hspace{0.25em}\Wd{X, H(X,\eta)}{X,E(Y)}.
    \end{equation}
Consequently, if $(D,E,H)$ is a triplet that satisfies
\begin{equation}
    \E\dnorm{Y - D\circ E(Y)}_2=0,\quad 
    \Wd{X, H(X, \eta)}{X, E(Y)}=0,
\end{equation}
then the generator achieves conditional distribution matching:
\begin{align*}
 G(x,\eta) \sim \Prob_{Y|X=x}.
\end{align*}

\textbf{Proof:} Let $\mu = \Prob_{X, H(X,\eta)}$, $\nu = \Prob_{X, E(Y)}$, and $f(x,y) = (x, D(y))$. Then $f$ is a $(1\vee K)$-Lipschitz function, since
\begin{align*}
    \dnorm{f(x_1 , y_1) - f(x_2,y_2)}_2 &= \sqrt{ \dnorm{x_1 - x_2}_2^2 + \dnorm{D(y_1) - D(y_2)}_2^2}\\
    &\leq  \sqrt{ \dnorm{x_1 - x_2}_2^2 + K^2\dnorm{ y_1 -  y_2}_2^2}\\
    &\leq \sqrt{ (1\vee K)^2\dnorm{x_1 - x_2}_2^2 + (1\vee K)^2\dnorm{ y_1-y_2}_2^2}\\
    &\leq  (1\vee K) \sqrt{\dnorm{x_1 - x_2}_2^2 + \dnorm{ y_1-y_2}_2^2}\\ 
    &= (1\vee K)  \dnorm{\left(x_1 - x_2,  y_1-y_2\right)}_2
\end{align*}

Let $f_{\#\mu },f_{\#\nu }$ denotes the pushforward measure of $\mu$ and $\nu$ by $f$, respectively. 
\begin{align*}
    \Wd{X, D\circ H(X, \eta)}{X,D\circ E(Y)}  &= W_1\left(f_{\#\mu},f_{\#\nu }\right)\\
    &= \inf_{\gamma \in \Gamma(f_{\#\mu}, f_{\#\nu })}\E_{(x,y)\sim \gamma} \dnorm{x-y}_2\\
    &\leq \inf_{\gamma' \in \Gamma(\mu ,\nu)} \E_{(x',y')\sim \gamma'}\dnorm{f(x') - f(y')}_2\\
    &\leq (1\vee K)\inf_{\gamma' \in \Gamma(\mu ,\nu)} \E_{(x',y')\sim \gamma'}\dnorm{x' - y'}_2\\
    &=  (1\vee K) W_1\left(\mu , \nu\right) \\
    &= (1\vee K)\Wd{X, H(X,\eta)}{X, E(Y)}
\end{align*}
where $\Gamma(f_{\#\mu}, f_{\#\nu })$ is the set of all couplings of $f_{\#\mu}$ and $f_{\#\nu }$. The second line is by definition \ref{Definition:WassersteinDistance}. The first inequality is due to the fact that, for any $(x', y') \sim \gamma' \in \Gamma(\mu, \nu)$, $(f(x'), f(y'))$ has a joint distribution with marginals $(f_{\#\mu}, f_{\#\nu })$. The rest of the theorem follows from Theorem \ref{Theorem:cLSDM}.

\subsection{Proof of Proposition \ref{Proposition:ConnectionsWithfGAN}}
 Let $\mathbb{D}_f\left(\mu, \nu\right)$ denote one of the following divergences between two probability measures $\mu$ and $\nu$ with bounded support $\Omega$: (i) Kullback–Leibler, (ii) $\chi^2$, (iii) Jensen-Shannon, or (iv) total variation (formal definitions are provided in the supplementary material). Then
    \begin{equation}
    W_1\left(\mu, \nu\right)\leq 2\,\text{diam}(\Omega) \max\biggl\{\mathbb{D}_f\left(\mu, \nu\right), \sqrt{\frac{1}{2}\mathbb{D}_f\left(\mu, \nu\right)}\biggr\}.
\end{equation}
where $\text{diam}(\Omega)=\sup_{x,y \in \Omega}\dnorm{x-y}_2$ denotes the diameter of $\Omega$.

\textbf{Proof:\\}
When the support is bounded, \cite{MetricBounds} summarized a wide range of bounds relating the 1-Wasserstein distance to other probability metrics and divergences. 

\begin{enumerate}
    \item (Total variation distance): By Theorem 4 in \cite{MetricBounds}:
$$W_1(\mu, \nu ) \leq \text{diam}(\Omega) \dnorm{\mu - \nu}_{TV}.$$
\item (KL divergence): By Pinsker’s inequality,
$\dnorm{\mu - \nu}_{TV}\leq \sqrt{\frac{1}{2} \mathbb{D}_{KL}(\mu,\nu )}$. Therefore, 
$$W_1(\mu, \nu ) \leq \text{diam}(\Omega) \sqrt{\frac{1}{2} \mathbb{D}_{KL}(\mu,\nu )}.$$

\item (Jensen-Shannon distance): By definition,
$$\mathbb{D}_{JS}(\mu,\nu)
:= \frac{1}{2} \mathbb{D}_{KL}(\mu , m) + \frac{1}{2} \mathbb{D}_{KL}(\nu , m),
\qquad m := \frac{1}{2}(\mu+\nu),$$
Since $\mu - m \;=\; \mu - \frac12(\mu+\nu) \;=\; \frac12(\mu - \nu)$ and $\|\mu - \nu\|_{TV} := \sup_{A \in \mathcal{B}(\Omega)} |\mu(A) - \nu(A)|$, we have
\[
\|\mu - m\|_{TV} = \frac12 \|\mu - \nu\|_{TV}.
\]
Similarly,
\[
\|\nu - m\|_{TV} = \frac12 \|\mu - \nu\|_{TV}.
\]
By Pinsker's inequality,
\[
\|\mu - m\|_{TV}^2 \;\le\; \frac{1}{2} \mathbb{D}_{KL}(\mu , m),
\qquad
\|\nu - m\|_{TV}^2 \;\le\; \frac{1}{2}\mathbb{D}_{KL}(\nu , m).
\]
Using $\|\mu-m\|_{TV} = \tfrac12 \|\mu-\nu\|_{TV}$, we obtain
\[
\mathbb{D}_{KL}(\mu , m)
\;\ge\; 2\|\mu-m\|_{TV}^2
\;=\; 2\left(\frac12 \|\mu-\nu\|_{TV}\right)^2
\;=\; \frac12 \|\mu-\nu\|_{TV}^2,
\]
and the same bound holds for $\mathbb{D}_{KL}(\nu , m)$:
\[
\mathbb{D}_{KL}(\nu, m) \;\ge\; \frac12 \|\mu-\nu\|_{TV}^2.
\]
Therefore
\[
\begin{aligned}
\mathbb{D}_{JS}(\mu,\nu)
&= \frac{1}{2} \mathbb{D}_{KL}(\mu , m) + \frac{1}{2} \mathbb{D}_{KL}(\nu , m)\\
&\ge \frac12 \cdot \frac12 \|\mu-\nu\|_{TV}^2
   + \frac12 \cdot \frac12 \|\mu-\nu\|_{TV}^2 \\
&= \frac12 \,\|\mu-\nu\|_{TV}^2.
\end{aligned}
\]
Equivalently,
\begin{equation}\label{eq:TV-JS}
\|\mu-\nu\|_{TV} \;\le\; \sqrt{2\,\mathbb{D}_{JS}(\mu,\nu)}.
\end{equation}
Combine with $W_1(\mu, \nu ) \leq \text{diam}(\Omega) \dnorm{\mu - \nu}_{TV}$,
\[
W_1(\mu,\nu)
\;\le\; \mathrm{diam}(\Omega)\,\|\mu-\nu\|_{TV}
\;\le\; 2\, \mathrm{diam}(\Omega)\sqrt{\frac{1}{2}\mathbb{D}_{JS}(\mu,\nu)}.
\]
\item ($\chi^2$ divergence) By Theorem 5 in \cite{MetricBounds}, $\mathbb{D}_{KL}(\mu,\nu ) \leq \mathbb{D}_{\chi^2}(\mu,\nu )$. Therefore,
$$W_1(\mu,\nu)\leq \text{diam}(\Omega)\sqrt{\frac{1}{2} \mathbb{D}_{\chi^2} (\mu, \nu)}.$$
\end{enumerate}

\subsection{Lemma \ref{Lemma:W1_cW1_Bound}}
\label{Lemma:W1_cW1_Bound} Consider two probability distributions with finite first moment, $\Prob_{X,Y}$ and $\Prob_{X,Z}$ on $\R^{p + q}$, where they share the same marginal $\Prob_X$. The following inequality holds
$$\Wd{X,Y}{X,Z} \leq \E_X W_1(\Prob_{Y|X}, \Prob_{Z|X})$$
\textbf{Proof:\\}
This is proved in \citet{cW1Inequality}. We provide a proof here for completeness. By definition of the 1-Wasserstein distance:
\begin{align*}
    \Wd{X,Y}{X,Z} &= \inf_{\gamma' \in \Gamma(\Prob_{X,Y}, \Prob_{X,Z})} \E_{(X_1,Y_1,X_2,Z) \sim \gamma'}\dnorm{(X_1,Y_1)-(X_2,Z)}_2\\
    &=^{\text{Lemma }\ref{Lemma:KantorovichRubinsteinDuality}}\sup_{f \in \cF^1} \E_{X,Y} f(X,Y) - \E_{X,Z} f(X,Z)\\
    &= \sup_{f \in \cF^1} \E_X \biggl\{ \E_{Y|X} f(X,Y) - \E_{Z|X} f(X,Z)\biggr\}\\
    &\leq \E_X \sup_{f \in \cF^1} \biggl\{ \E_{Y|X} f(X,Y) - \E_{Z|X} f(X,Z)\biggr\}\\
    &\leq \E_X \sup_{f' \in \cF^1 }\biggl\{ \E_{Y|X} f'(Y) - \E_{Z|X} f'(Z)\biggr\}\\
    &=^{\text{Lemma }\ref{Lemma:KantorovichRubinsteinDuality}}\E_X W_1(\Prob_{Y|X}, \Prob_{Z|X}).
\end{align*}
Here, Lemma \ref{Lemma:KantorovichRubinsteinDuality} is the Kantorovich–Rubinstein duality. The second inequality is due to the fact that, for any $f\in \cF^1(\R^{p\times q}, \R)$, $f^x(y):= f(x,y)$ is a 1-Lipschitz function, since $|f^x(y_1) - f^x(y_2)|= |f(x,y_1)-f(x,y_2)| \leq \dnorm{y_1-y_2}$. Therefore, $f^x(y) \in \cF^1(\R^{q}, \R)$.

\subsection{Proof of Proposition \ref{Proposition:ConnectionsWithLDM}}

\textbf{Proof:\\}
The proof is a modification and extension of Theorem 4.2 in \citet{ConditionalDiffusionConvergence}. We extend it to bound the joint 1-Wasserstein distance using Pinsker's inequality and Lemma \ref{Lemma:W1_cW1_Bound}. 

Denote the generated distribution from backward process using true initial distribution $\Prob_{Z_T|X=x}$ and score $s(z,x,t)$ as $\Prob_{H_s|X=x}'$, the generated distribution from backward process using approximate initial distribution $\cN(0, I_m)$ and score $s(z,x,t)$ as $\Prob_{H_s|X=x}$. Note that the generated distribution from backward process using true initial distribution $\Prob_{Z_T|X=x}$ and true score $\nabla \log p_t(z|x)$ is $\Prob_{Z|X=x}$. Then we have
\begin{align*}
    \dnorm{\Prob'_{H_s|X=x}-\Prob_{H_s|X=x} }_{TV} &\leq \frac{1}{\sqrt{2}}\sqrt{\mathbb{D}_{KL}(\Prob'_{H_s|X=x},\Prob_{H_s|X=x})}\\
    &\leq \frac{1}{\sqrt{2}}\sqrt{\mathbb{D}_{KL}(\Prob_{Z_{T}|X=x} , \cN(0,I_m))}\\
    &\lesssim \frac{1}{\sqrt{2}} e^{-T} \sqrt{\mathbb{D}_{KL}(\Prob_{Z|X=x} , \cN(0,I_m))}\\
    &\leq \frac{1}{\sqrt{2}} e^{-T} \sqrt{C_1}.
\end{align*}
The first inequality is the Pinsker's inequality (2 in Lemma~\ref{Proposition:ConnectionsWithfGAN}). The second inequality is the data processing inequality. The third inequality is the logarithmic Sobolev inequality (Theorem 5.2.1, Exponential decay in entropy, in \citealp{Bakry}; Theorem 4 in \citealp{LogSoblev}). The fourth inequality is by assumption (1) in the proposition. 

Now, decompose the risk according to
\begin{align*}
   &\dnorm{\Prob_{H_s|X=x} - \Prob_{Z|X=x}}_{TV}\\
   &\leq  \dnorm{\Prob_{H_s|X=x}  - \Prob_{H_s|X=x} '}_{TV} + \dnorm{\Prob_{Z|X=x} - \Prob_{H_s|X=x} '}_{TV}
\end{align*}
By Lemma \ref{UnveilLDM:lemmaD4} and the Pinsker's inequality,
\begin{align*}
     \dnorm{\Prob_{Z|X=x} - \Prob_{H_s|X=x}'}_{TV} &\lesssim \sqrt{\mathbb{D}_{KL}(\Prob_{Z|X=x} ,\Prob_{H_s|X=x}')}\\
    & \lesssim\sqrt{\int_{0}^T \frac{1}{2}\E_{Z_t|X=x} \dnorm{s(z_t,x,t) - \nabla \log p_t(z_t|x)}_2^2 \, dt}.
\end{align*}
By the maximal coupling lemma (Proposition 4.7, \citealp{MaxCoupling}), there exists a coupling $\gamma^*$ of $A \sim \mu= \Prob_{H_s|X=x} $ and $ B \sim  \nu=\Prob_{Z|X=x} $ such that $\dnorm{\mu - \nu}_{TV} = \Prob(A \neq B)$. Using this coupling $\gamma^*$, by definition of 1-Wasserstein distance,
\begin{align*}
     W_1(\Prob_{H_s|X=x}, \Prob_{Z|X=x})  &= \inf_{(A,B) \sim \Gamma(\Prob_{H_s|X=x}, \Prob_{Z|X=x})} \E \dnorm{A-B}_2\\
     &\leq \E_{(A,B)\sim \gamma^*} \dnorm{A-B}_2\\
     &= \E_{(A,B)\sim \gamma^*} \dnorm{A-B}_2 \mathbbm{1}(A\neq B)\\
     &\leq \sqrt{\E_{\gamma^*} \dnorm{A-B}_2^2} \sqrt{\E_{\gamma^*} \mathbbm{1}(A\neq B)}\\
     &\leq \sqrt{2 (\E \dnorm{A}_2^2 + \E \dnorm{B}_2^2)} \sqrt{\dnorm{\mu - \nu }_{TV}}\\
     &\lesssim\sqrt{\dnorm{\mu - \nu}_{TV}},
\end{align*}
where the second line is due to $\gamma^* \in \Gamma(\Prob_{H_s|X=x}, \Prob_{Z|X=x})$ being a coupling, fourth line is the Cauchy–Schwarz inequality, fifth line is due to $\dnorm{\mu - \nu}_{TV} = \Prob(A \neq B)$, and last line is by assumptions.

Combining the above, we have
\begin{align*}
    W_1(\Prob_{H_s|X=x}, \Prob_{Z|X=x}) &\lesssim  \sqrt{\dnorm{\Prob_{H_s|X=x} - \Prob_{Z|X=x}}_{TV} }\\
    &\lesssim e^{-T/2} +  \left(\int_{0}^T \frac{1}{2}\E_{Z_t|X=x} \dnorm{s(z_t,x,t) - \nabla \log p_t(z_t|x)}_2^2 \, dt\right)^{1/4} .
\end{align*}

Take expectation w.r.t. the predictor $X$, and apply Lemma \ref{Lemma:W1_cW1_Bound}, we have
\begin{align*}
   W_1(\Prob_{X,H_s}, \Prob_{X,Z}) &\leq  \E_X   W_1(\Prob_{H_s|X=x}, \Prob_{Z|X=x})\\
   &\lesssim  e^{-T/2} + \E_X \left(\int_{0}^T \frac{1}{2}\E_{Z_t|X=x} \dnorm{s(z_t,x,t) - \nabla \log p_t(z_t|x)}_2^2 \, dt \right)^{1/4}\\
   &\lesssim  e^{-T/2} + \left(\int_{0}^T \E_X  \E_{Z_t|X=x} \dnorm{s(z_t,x,t) - \nabla \log p_t(z_t|x)}_2^2 \, dt\right)^{1/4}\\
   &=  e^{-T/2} + \left(T \cL_{SM}(s)\right)^{1/4} .
\end{align*}

\subsection{Proof of Theorem \ref{Theorem:ReconstructionError}:}

\hfill

\textbf{Proof:}
We define the empirical and population losses as:
\begin{align*}
   \hL(D,E) =  \widehat{\cL}_{recon}(D,E) &= \frac{1}{n+N} \sum_{i=1}^{n+N} \dnorm{Y_i - D\circ E(Y_i)}_2,\\
    L(D,E)= \cL_{recon}(D,E) &= \E \dnorm{Y - D\circ E(Y)}_2.
\end{align*}
Let $(\hD,\hE)$ be a pair which satisfies
$$(\hD,\hE) \in \argmin_{D \in \cD, E\in \cE} \hL(D,E)$$
For any $D\in \cD, E\in \cE$:
\begin{align*}
    L(\hD, \hE) &= L(D,E) + (L(\hD, \hE) - \hL(\hD, \hE)) + (\hL(\hD, \hE) - L(D,E))\\
    &\leq L(D,E) + (L(\hD, \hE) - \hL(\hD, \hE)) + (\hL(D,E) - L(D,E))\\
    &\leq L(D,E) + 2 \sup_{D \in \cD, E\in\cE} |\hL(D,E) - L(D,E)|
\end{align*}
Let $\cF = \bigl\{f(y) = \dnorm{y - D\circ E(y)}_2: D\in \cD, E\in \cE\bigr\}$ and take infimum over $D\in \cD, E\in \cE$,
\begin{align*}
    L(\hD, \hE) &\leq \inf_{D\in \cD, E\in \cE}L(D,E) + 2 \sup_{D \in \cD, E\in\cE} |\hL(D,E) - L(D,E)|\\
    &=\inf_{D\in \cD, E\in \cE}L(D,E) + 2 \dnorm{\widehat{\Prob}_{Y} - \Prob_Y }_\cF\\
    &:= \Delta_{miss} + 2*\Delta_{gen}
\end{align*}
where $\hProb_{Y}$ is the empirical measure of combined response $\cP_Y \cup \cU$, and 
\begin{equation}
    \Delta_{miss} = \inf_{D\in \cD, E\in \cE}L(D,E)
\end{equation}
\begin{equation}
    \Delta_{gen} = \sup_{D \in \cD, E\in\cE} |\hL(D,E) - L(D,E)| = \dnorm{\widehat{\Prob}_{Y} - \Prob_Y }_\cF
\end{equation}
$\Delta_{miss}$ is misspecification error and $\Delta_{gen}$ is generalization error. Take expectation over $\{Y_i\}_{i=1}^{N+n}$, 
$$\E  L(\hD, \hE) \leq \Delta_{miss} + 2*\E\Delta_{gen}$$

\paragraph{Bounding $\Delta_{miss}$: }By Assumption \ref{Assumption:Existence}, for any $m \geq d_\cY$, one can construct $D^* \in \cH^{\beta_d}([0,1]^{m}, [0,1]^q, C)$ and $E^* \in \cH^{\beta_e}([0,1]^q, [0,1]^{m}, C)$ satisfying $\cL_{\text{recon}}(D^*,E^*) = 0$, by ignoring the redundant dimensions $m-d_\cY$. Therefore, for any $D \in \cD, E \in \cE$,
\begin{align*}
    L(D,E) &= \E \dnorm{Y - D\circ E (Y)}_2\\
    &= \E \dnorm{Y - D^*\circ E^* (Y) + D^*\circ E^* (Y) - D\circ E(Y)}_2\\
    &=   \E \dnorm{D^*\circ E^* (Y) - D\circ E(Y)}_2\\
    &\leq \dnorm{D^*\circ E^*- D\circ E}_{L_\infty(\cY)}\\
    &= \dnorm{D^*\circ E^* - D^* \circ E + D^*\circ E - D\circ E}_{L_\infty(\cY)}\\
    &\leq \dnorm{D^*\circ E^* - D^* \circ E }_{L_\infty(\cY)} + \dnorm{ D^*\circ E - D\circ E }_{L_\infty(\cY)}\\
    &\leq  C \dnorm{E^* - E }_{L_\infty(\cY)}^{\beta_d\wedge 1} + \dnorm{ D^* - D}_{L_\infty([0,1]^m)}
\end{align*}

Apply Lemma \ref{chakraborty:theorem18} with $\epsilon_d$ for $\dnorm{ D^* - D}_{L_\infty([0,1]^m)}$ term and $\epsilon_e$ for $\dnorm{E^* - E }_{L_\infty(\cY)}$ term. For any $s > d_\cY$, there exists a constant $\alpha>0$ (which may depend on $\beta_d, \beta_e,m,p,C$) and $\epsilon_0$ (which may depend on $\cY$), such that if $\epsilon_d, \epsilon_e \in (0, \epsilon_0]$, ReLU networks  $ D'$ and $E'$ can be constructed with $\cL(D') \leq L_d=\alpha \log(1/\epsilon_d)$, $\cL(E') \leq L_e=\alpha \log(1/\epsilon_e)$ , $\cS(D')\leq S_d=\alpha \log(1/\epsilon_d)\epsilon_d^{-m/\beta_d}$, $\cS(E^*)\leq S_e=\alpha \log(1/\epsilon_e)\epsilon_e^{-s/\beta_e}$, such that $\dnorm{ D^* - D'}_{L_\infty([0,1]^m)} \leq \epsilon_d, \dnorm{E^* - E' }_{L_\infty(\cY)} \leq \epsilon_e$, by approximating $D^*_i$ and $E^*_i$ for each coordinate output and stacking the ReLU networks obtained parallel. Hence, by setting $\cD = \cR\cN(L_d,S_d,1), \cE=\cR\cN(L_e,S_e,1)$,
\begin{align*}
    \Delta_{miss} &= \inf_{D\in \cD, E\in\cE} L(D,E) \leq L(D', E')\\
    &\leq C \epsilon_e^{\beta_d \wedge 1} + \epsilon_d
\end{align*}

\paragraph{Bounding $\E\Delta_{gen}$: } 
Since the range of $D\in \cD=\cR\cN(L_d,S_d,B_d=1)$ and the support $\cY$ is bounded, $\sup_{f \in \cF} \dnorm{f}_\infty \leq B$ for some constant $B$.  By Lemma \ref{Lemma12Huang},
\begin{equation}\label{Equation:chaining}
    \E\Delta_{gen} \leq 8 \E \inf_{0<\delta<B/2}\left( \delta + \frac{3}{\sqrt{n+N}} \int_{\delta}^{B/2} \sqrt{\log \cN \left(\epsilon, \cF_{|Y_{1:n+N}}, \dnorm{\cdot}_\infty \right)} d\epsilon\right) 
\end{equation}
$\log \cN \left(\epsilon, \cF_{|Y_{1:n+N}}, \dnorm{\cdot}_\infty \right)$ can be bounded with $ \log \cN \left(\epsilon, (\cD \circ \cE)_{|Y_{1:n+N}}, \dnorm{\cdot}_\infty \right) $, where $\cD \circ \cE = \{f(y) = D\circ E(y): D\in \cD, E\in \cE\}$:

Let $A \subseteq \R^{(n+N)q}$ be an optimal $\epsilon$-cover of $(\cD \circ \cE)_{|Y_{1:n+N}}$ in $\dnorm{\cdot}_\infty$ . That is, for any $D \in \cD, E \in \cE $, there is a $v\in A$ such that $\dnorm{v - (D\circ E)_{|Y_{1:n+N}}}_\infty \leq \epsilon$, which means $\dnorm{v_i - D\circ E(Y_i)}_\infty \leq \epsilon, \forall i=1,2,..., n+N$, where $v_i \in \R^q$. Define $B = \bigl\{ \left(\dnorm{Y_1 - v_1}_2, ..., \dnorm{Y_{n+N} - v_{n+N}}_2\right) :  v \in A\bigr\} \subseteq \R^{n+N}$. Clearly, $|B| = |A|$.  Then, for any $D \in \cD, E \in \cE$,  $\vert  \dnorm{Y_i - v_i}_2 - \dnorm{Y_i - D\circ E(Y_i)}_2 \vert \leq  
 \dnorm{v_i - D\circ E(Y_i)}_2 \leq \epsilon, \forall i=1,...,n+N$. Thus, $\forall f \in \cF$,  $ \dnorm{f_{|Y_{1:n+N}} - b}_\infty \leq \epsilon$ for some $b=\left(\dnorm{Y_1 - v_1}_2, ..., \dnorm{Y_{n+N} - v_{n+N}}_2 \right) \in B$. This implies that $B$ is an $\epsilon$-cover of $\cF_{|Y_{1:n+N}}$ w.r.t. $\dnorm{\cdot}_\infty$.  Therefore,
$$\cN \left(\epsilon, \cF_{|Y_{1:n+N}}, \dnorm{\cdot}_\infty \right) \leq \cN \left(\epsilon, (\cD \circ \cE)_{|Y_{1:n+N}}, \dnorm{\cdot}_\infty \right)$$
By Lemma \ref{metricentropybound}, there is a positive constant $\theta$ (which may depend on $q$) such that, if $L_d, L_e \geq 3$, $S_e\ge 6q+2qL_e, S_d \geq 6q+2qL_d$, and $n+N\geq \theta(S_d+S_e)(L_e+L_d)\left(\log \left(S_e+S_d\right) + L_e+L_d\right)$, then 
$$\log \cN \left(\epsilon, (\cD \circ \cE)_{|Y_{1:n+N}}, \dnorm{\cdot}_\infty \right)  \lesssim (S_e+S_d)(L_e+L_d)\left(\log \left(S_e+S_d\right) + L_e+L_d\right)\log \left(\frac{(n+N)q}{\epsilon}\right)$$
If $n+N< \theta(S_d+S_e)(L_e+L_d)\left(\log \left(S_e+S_d\right) + L_e+L_d\right)$, then 
$$ \cN \left(\epsilon, (\cD \circ \cE)_{|Y_{1:n+N}}, \dnorm{\cdot}_\infty \right) \leq \cN \left(\epsilon, [0,1]^{q(n+N)}, \dnorm{\cdot}_\infty \right) \leq \lceil {2/\epsilon}\rceil ^{q(n+N)},$$
which gives the same bound. Thus, it holds for all $n+N$. 

Substitute it into (\ref{Equation:chaining}),
\begin{align*}
    \E\Delta_{gen} &\lesssim \E \inf_{0<\delta<B/2}\left( \delta + \frac{3}{\sqrt{n+N}} \int_{\delta}^{B/2} \sqrt{\log \cN \left(\epsilon, \cF_{|Y_{1:n+N}}, \dnorm{\cdot}_\infty \right)} d\epsilon\right) \\
    &\lesssim  \frac{1}{\sqrt{n+N}} \int_{0}^{B/2} \sqrt{(S_e+S_d)(L_e+L_d)\left(\log \left(S_e+S_d\right) + L_e+L_d\right)\log \left(\frac{(n+N)q}{\epsilon}\right)}d\epsilon\\
    &\leq  \sqrt{\frac{(S_e+S_d)(L_e+L_d)\left(\log \left(S_e+S_d\right) + L_e+L_d\right)}{n+N}} \int_{0}^{B} \sqrt{\log \left(\frac{(n+N)q}{\epsilon}\right)}d\epsilon\\
    &\lesssim \sqrt{\frac{(S_e+S_d)(L_e+L_d)\left(\log \left(S_e+S_d\right) + L_e+L_d\right) \log\left((n+N)q\right)}{n+N}}
\end{align*}

\paragraph{Bounding $\E L(\hD, \hE)$:}
Combining the above,
\begin{align*}
    \E L(\hD, \hE) &= \Delta_{miss} + 2*\Delta_{gen}\\
    % & \lesssim \epsilon_e^{\beta_d \wedge 1} + \epsilon_d + \sqrt{\frac{(S_e+S_d)(L_e+L_d)(\log(S_e+S_d)+L_e+L_g)\log(n+N)}{n+N}}\\
    &\lesssim   \epsilon_e^{\beta_d \wedge 1} + \epsilon_d + \sqrt{\frac{(S_e+S_d)(L_e+L_d)(\log(S_e+S_d)+L_e+L_d)\log\left((n+N)q\right)}{n+N}}\\
    &\lesssim  \epsilon_e^{\beta_d \wedge 1} + \epsilon_d + \sqrt{\frac{(S_e+S_d)(L_e+L_d)(\log(S_e+S_d)+L_e+L_d)\log(n+N)}{n+N}}\\
    &\lesssim  \epsilon_e^{\beta_d \wedge 1} + \epsilon_d + \sqrt{\frac{(S_e+S_d)(L_e+L_d)(\log(S_e+S_d))\log(n+N)}{n+N}}\\
    &\lesssim \epsilon_e^{\beta_d \wedge 1} + \epsilon_d + \left(\log (\frac{1}{\epsilon_e \wedge\epsilon_d})\right)^{3/2}\left(\sqrt{\frac{(\epsilon_e^{-s/\beta_e }+ \epsilon_d^{-m/\beta_d})\log (n+N)}{n+N}}\right)\\
    &\lesssim \epsilon_e^{\beta_d \wedge 1} + \epsilon_d + \left(\log (\frac{1}{\epsilon_e \wedge\epsilon_d})\right)^{3/2}\left(\sqrt{\frac{\epsilon_e^{-s/\beta_e }\log (n+N)}{n+N}} + \sqrt{\frac{\epsilon_d^{-m/\beta_d}\log (n+N)}{n+N}}\right)
\end{align*}
Choose $\epsilon_d \asymp (n+N)^{-\frac{1}{2+\frac{m}{\beta_d}}}$ and $\epsilon_e \asymp (n+N)^{-\frac{1}{2(\beta_d \wedge1)+\frac{s}{\beta_e}}}$, then
$$\E L(\hD, \hE) \lesssim \log^2(n+N) \times (n+N)^{-\frac{1}{\max \{2+\frac{m}{\beta_d}, 2+\frac{s}{\beta_e(\beta_d \wedge 1)}\}}}$$

\subsection{Lemma \ref{Lemma:GeneralizationErrorLipschitzClass}}
Let $\FLip$ be the class of 1-Lipschitz functions from $[0,1]^{p+q}$ to $\R$, and  $\FLip_1$ be the class of bounded 1-Lipschitz functions from $[0,1]^{p+q}$ to $\R$ with $\sup_{f \in \FLip_1}\dnorm{f}_{L_\infty([0,1]^{p+q})} \leq 1$. 

\begin{lemma} \label{Lemma:GeneralizationErrorLipschitzClass}
    Suppose there are $n$ iid samples $Z_i=(X_i, Y_i) \in \R^p \times \R^d$, and $\cD=\cR\cN(L_d,S_d, B_d)$ is a class of neural networks with output dimension $q$. For function class $ \cK = \bigl\{k(x,y) = f(x, D(x,y)): f \in \FbLip, D \in \cD \bigr\}$ or $ \cK = \bigl\{k(x,y) = f(x, D(y)): f \in \FbLip, D \in \cD \bigr\}$, if $L_d\geq 3$, $S_d\geq 6q+2qL_d$, then
    $$\E_{Z_{1:n}} \dnorm{\Prob_{Z} - \hProb_{Z}}_\cK \lesssim  n^{-\frac{1}{p+q}} + \sqrt{\frac{S_d L_d \left(\log(S_d) +L_d\right)\log(nq)}{n}}$$

\end{lemma}

\hfill

\textbf{Proof:} Consider $ \cK = \bigl\{k(x,y) = f(x, D(x,y)): f \in \FbLip, D \in \cD \bigr\}$. By Lemma \ref{metricentropybound}, there is a positive constant $\theta'$ (which may depends on $m,q$), such that if  $n\geq \theta'(S_d+6q+2qL_d)(L_d+3)(\log(S_d+6q+2qL_d) + L_d + 3)$,
$$\log \cN(\epsilon, \cD_{|Z_{1:n}}, \dnorm{\cdot}_\infty) 
 \lesssim  \left(S_d+6q+2qL_d\right) \left(L_d+3\right)\left(\log(S_d+6q+2qL_d)+L_d +3\right) \log \left(\frac{nq}{\epsilon}\right)$$
If $L_d\geq 3$, $S_d\geq 6q+2qL_d$, then $\theta'(S_d+6q+2qL_d)(L_d+3)(\log(S_d+6q+2qL_d) + L_d + 3) \leq  4\theta'S_dL_d(\log2 + \log S_d+2L_d)\leq 8\theta'S_dL_d(\log S_d+L_d)$. Choose $\theta = 8\theta'$, then if $n \geq \theta S_d L_d(\log(S_d) + L_d)$,
$$\log \cN(\epsilon, \cD_{|Z_{1:n}}, \dnorm{\cdot}_\infty) 
 \lesssim  S_d L_d\left(\log(S_d)+L_d\right) \log \left(\frac{nq}{\epsilon}\right)$$
 If $n \leq \theta S_d L_d(\log(S_d) + L_d) $, then
$$\cN(\epsilon, \cD_{|Z_{1:n}}, \dnorm{\cdot}_\infty)  \leq \cN(\epsilon, [0,1]^{qn}, \dnorm{\cdot}_\infty) \leq \lceil2/\epsilon\rceil^{qn},$$
 which gives the same bound.
 
 Let $A$ be an optimal $\epsilon$-cover of $\cD_{|Z_{1:n}} \subseteq \R^{nq}$. That is, $\forall D \in \cD$, there $\exists v_D \in A\subseteq \R^{n q}$ such that $\dnorm{v_D - D_{|Z_{1:n}}}_\infty \leq \epsilon$. This implies that $\dnorm{v_{D,i} - D(Z_i)}_\infty \leq \epsilon, \forall i=1,2,..., n$, where $v_{D,i}\in \R^q$. Let $P = (\tf_1, ..., \tf_r )$ be an optimal $\epsilon$-cover of $\FbLip$ w.r.t $\dnorm{\cdot}_\infty$. That is, $\forall f \in \FbLip$, there $\exists\tf \in P$ s.t $\dnorm{f - \tf}_\infty\leq \epsilon$. By Lemma \ref{metricentropyHolder}, $\log r\leq C_1\epsilon^{-(p+q)}$ for some constant $C_1$. Consider the set $M = \{ (\tf(X_1, v_1), ..., \tf(X_n, v_n)), \tf \in P, v\in A\}\subseteq \R^n$. Then, $|M|= r |A|$.  And, $\forall k \in \cK_{|Z:_{1:n}}\subseteq \R^{n}, k=(f(X_1,D(Z_1)), ..., f(X_n, D(Z_n)))$ for some $D\in \cD, f\in \FbLip$. Let $\tf \in P$ be the corresponding function s.t. $\dnorm{f-\tf}_\infty \leq \epsilon$, then  $|f(X_i,D(Z_i) ) - \tf(X_i, v_{D,i})| \leq |f(X_i, D(Z_i)) - \tf(X_i, D(Z_i))| + |\tf(X_i, D(Z_i)) - \tf(X_i, v_{D,i})| \leq \dnorm{f - \tf}_\infty + \dnorm{(X_i, D(Z_i)) - (X_i, v_{D,i})}_2\leq \epsilon + \sqrt{q} \epsilon = (\sqrt{q}+1)\epsilon, \forall i=1,2,...,n$. Hence, $M$ is an $(\sqrt{q}+1) \epsilon$-cover of $\cK_{|Z_{1:n}}$ w.r.t. $\dnorm{\cdot}_\infty$. Thus, 
$$\cN\left(\epsilon, \cK_{|Z_{1:n}}, \dnorm{\cdot}_\infty\right) \leq \cN\left(\frac{\epsilon}{2}, \FbLip, \dnorm{\cdot}_\infty\right) *\cN\left(\frac{\epsilon}{2\sqrt{q}}, \cD_{|Z_{1:n}}, \dnorm{\cdot}_\infty\right)$$
Consequently,
$$\log \cN(\epsilon, \cK_{|Z_{1:n}}, \dnorm{\cdot}_\infty) 
 \lesssim \epsilon^{-(p+q)}+ S_d L_d \left(\log(S_d)+L_d\right) \log \left(\frac{nq}{\epsilon}\right)$$
The same holds for $ \cK = \bigl\{k(x,y) = f(x, D(y)): f \in \FbLip, D \in \cD \bigr\}$ by replacing $\cD_{|Z_{1:n}}$ with $\cD_{|Y_{1:n}}$. Because $\sup_{k \in \cK} \dnorm{k}_\infty = 1 <2$. By Lemma \ref{Lemma12Huang},
\begin{align*}
    \E_{Z_{1:n}}\dnorm{\Prob_{Z} - \hProb_{Z}}_\cK &\leq 8 \E_{Z_{1:n}} \inf_{0<\delta<1} \left( \delta +\frac{3}{\sqrt{n}} \int_{\delta}^1 \sqrt{\log \cN(\epsilon, \cK_{|Z_{1:n}}, \dnorm{\cdot}_\infty)} d\epsilon\right)\\
    &\lesssim  \E \inf_{0<\delta<1}\left( \delta + \frac{3}{\sqrt{n}}\int_{\delta}^1\sqrt{   \epsilon^{-(p+q)}+ S_d L_d(\log(S_d) +L_d)\log \left(\frac{nq}{\epsilon}\right)     } d\epsilon \right)\\
    &\lesssim  \E \inf_{0<\delta<1}\left( \delta + \frac{3}{\sqrt{n}}\int_{\delta}^1\sqrt{   \epsilon^{-(p+q)}   } + \sqrt{S_d L_d(\log(S_d) +L_d) \log \left(\frac{nq}{\epsilon}\right) }  \hspace{0.25em}d\epsilon \right)\\
    &\lesssim  n^{-\frac{1}{p+q}} + \sqrt{\frac{S_d L_d(\log(S_d) +L_d)\log(nq)}{n}}
\end{align*}

\subsection{Proof of Theorem \ref{Theorem:DistributionMatchingConsistency}}

Since $\cX \times \cY \subseteq [0,1]^{p+q}$ is bounded by Assumption \ref{Assumption:Boundedness}, for any probability measures $\mu, \nu,\gamma$ on $\cX \times \cY$
\begin{equation}\label{Equation:LipschitzClassBound}
    \dnorm{\mu - \nu }_{\FbLip} \leq W_1(\mu, \nu) = \dnorm{\mu-\nu}_{\FLip} \leq  \sqrt{p+q} \dnorm{\mu - \nu}_{\FbLip}
\end{equation}
The first inequality is due to $\FLip_1 \subseteq \FLip$. The equality is the Kantorovich–Rubinstein duality. For the second inequality, see Remark 6 of \cite{GANError}. The same argument applies to $\cX \times \cZ \subseteq [0,1]^{p+m}$.

Because $\FbLip$ is a symmetric class, the absolute sign in the definition of IPM can be omitted, for every $f\in \FbLip$ its negation $-f$ is also in $\FbLip$.  Moreover, it satisfies $\dnorm{\mu - \nu}_{\FbLip} = \dnorm{\nu - \mu}_{\FbLip}$, and the triangle inequality $\dnorm{\mu - \nu}_{\FbLip} \leq \dnorm{\mu - \gamma}_{\FbLip} + \dnorm{\gamma - \nu}_{\FbLip}$. The properties also hold for IPM defined by $\FLip$. We will use the properties throughout the proof.

We consider the case of cLSDM and dLSDM separately. For notational simplicity, we write $\Prob_{X, D\circ E}$ to denote the distribution of $(X, D\circ E(Y))$, and similarly for the other terms, throughout the proof.

\subsubsection{Distribution matching error of cLSDM}
The empirical risk minimizer $\hH$ satisfies
\begin{equation} \label{Equation:ERMOfcLSDM}
    \hH \in \argmin_{ H \in \cH} \dnorm{\hProb_{X, \hD\circ H } - \hProb_{X, \hD \circ \hE}}_{\FLip}.
\end{equation}
The true risk in Step 2 is $\dnorm{\Prob_{X, \hD\circ \hH } - \Prob_{X, \hD \circ \hE}}_{\FLip}$, we decompose the risk as follow:
\begin{align*}
    \dnorm{\Prob_{X, \hD\circ \hH } - \Prob_{X, \hD \circ \hE}}_{\FLip} &\leq \dnorm{\Prob_{X, \hD\circ \hH } - \hProb_{X, \hD\circ \hH }}_{\FLip} + \dnorm{ \hProb_{X, \hD\circ \hH } - \hProb_{X, \hD \circ \hE}}_{\FLip} + \dnorm{\hProb_{X, \hD \circ \hE} - \Prob_{X, \hD \circ \hE}}_{\FLip} \\
    &:= \delta_1 + \Delta_{app} +\delta_2. 
\end{align*}

\textbf{Decomposing $\delta_1$: }We apply Assumption \ref{Assumption:Smoothness}, \ref{Assumption:Boundedness} and Theorem \ref{Theorem:dLSDM} to bound the term.
\begin{align*}
    \delta_1 &=  \dnorm{\Prob_{X, \hD\circ \hH } - \hProb_{X, \hD\circ \hH }}_{\FLip} \\
    &=  \dnorm{\Prob_{X, \hD\circ \hH } - \hProb_{X, \hD\circ \hH }}_{\FLip}  \mathbbm{1}(\cL_{recon}(\hD,\hE) \geq \delta) +  \dnorm{\Prob_{X, \hD\circ \hH } - \hProb_{X, \hD\circ \hH }}_{\FLip}  \mathbbm{1}(\cL_{recon}(\hD,\hE) < \delta)\\
    &\leq \sqrt{p+q}\mathbbm{1}(\cL_{recon}(\hD,\hE) \geq \delta) + \dnorm{\Prob_{X, \hD\circ \hH } - \hProb_{X, \hD\circ \hH }}_{\FLip}  \mathbbm{1}(\cL_{recon}(\hD,\hE) < \delta)\\
    &\leq \sqrt{p+q}\mathbbm{1}(\cL_{recon}(\hD,\hE) \geq \delta) + (K\vee 1) \dnorm{\Prob_{X, \hH } - \hProb_{X, \hH }}_{\FLip}  \mathbbm{1}(\cL_{recon}(\hD,\hE) < \delta)\\
    &\leq \sqrt{p+q}\mathbbm{1}(\cL_{recon}(\hD,\hE) \geq \delta) + (K\vee 1) \sqrt{p+m}\dnorm{\Prob_{X, \hH } - \hProb_{X, \hH }}_{\FbLip}  \mathbbm{1}(\cL_{recon}(\hD,\hE) < \delta)\\
    &\leq \sqrt{p+q}\mathbbm{1}(\cL_{recon}(\hD,\hE) \geq \delta) + (K\vee 1) \sqrt{p+m}\dnorm{\Prob_{X, \hH } - \hProb_{X, \hH }}_{\FbLip}
\end{align*}
The first inequality is due to Assumption \ref{Assumption:Boundedness} which implies  $\dnorm{\Prob_{X, \hD\circ \hH } - \hProb_{X, \hD\circ \hH }}_{\FLip} \leq \sqrt{p+q}$. The second inequality is due to Assumption \ref{Assumption:Smoothness} and the Lipschitz transfer property of the 1-Wasserstein distance when $\hD$ is Lipschitz (See proof in Theorem \ref{Theorem:dLSDM}). The third inequality is due to (\ref{Equation:LipschitzClassBound}). The last inequality is due to $\mathbbm{1}(\cL_{recon}(\hD,\hE) < \delta)\leq 1$. We further bound the second term:
\begin{align*}
        & \dnormF{\Prob_{X,\hH} - \hProb_{X, \hH}}\\
    &= \sup_{f \in \FbLip} \biggl\{\E_{X,\eta} f(X, \hH(X,\eta)) - 
    \frac{1}{n}\sum_{i=1}^n f(X_i,  \hH (X_i, \eta_i))\biggr\}\\
    &\leq \sup_{f \in \FbLip, H\in \cH} \biggl\{ \E_{X,\eta} f(X,  H(X,\eta)) - 
    \frac{1}{n}\sum_{i=1}^n f(X_i, H (X_i, \eta_i))  \biggr\}\\
    &=  \sup_{k \in \cK_1} \biggl\{ \E_{X,\eta}k(X,\eta) - \frac{1}{n}\sum_{i=1}^n k(X_i, \eta_i)\biggr\}\\
    &= \dnorm{\Prob_{X,\eta} - \hProb_{X,\eta}}_{\cK_1}\\
    &:= \Delta_{gen}^1
\end{align*}
where $\cK_1 = \bigl\{k(x,\eta) = f(x, H(x,\eta)): f \in \FbLip, H \in \cH \bigr\}$ is a symmetric function class, since $\FbLip$ is symmetric. Applying the Markov Inequality, 
$$\E\delta_1 \leq \sqrt{p+q} \frac{\E \cL_{recon}(\hD,\hE)}{\delta} +  (K\vee 1)\sqrt{p+m }  \E \Delta_{gen}^1. $$

\textbf{Decomposing $\delta_2$: } With a similar argument as in $\delta_1$, we have
$$\E\delta_2 \leq \sqrt{p+q} \frac{\E \cL_{recon}(\hD,\hE)}{\delta} +  (K\vee 1)\sqrt{p+m }  \E \Delta_{gen}^2. $$
where $\E \Delta_{gen}^2 =\dnorm{\Prob_{X,Y} - \hProb_{X,Y}}_{\cK_2}$ and $\cK_2 = \bigl\{k(x,y) = f(x, E(y)): f \in \FbLip, E \in \cE \bigr\}$ is a symmetric function class. 

\textbf{Decomposing $E\Delta_{app}$: } We use the empirical risk minimizing property of $\hH$ to bound the approximation error. For any $H \in \cH$:
\begin{align*}
    \Delta_{app} &= \dnorm{ \hProb_{X, \hD\circ \hH } - \hProb_{X, \hD \circ \hE}}_{\FLip} \\
    &= \dnorm{ \hProb_{X, \hD\circ \hH } - \hProb_{X, \hD \circ \hE}}_{\FLip}\mathbbm{1}(\cL_{recon}(\hD,\hE) \geq \delta) +\dnorm{ \hProb_{X, \hD\circ \hH } - \hProb_{X, \hD \circ \hE}}_{\FLip} \mathbbm{1}(\cL_{recon}(\hD,\hE) < \delta) \\
    &\leq \sqrt{p+q} \mathbbm{1}(\cL_{recon}(\hD,\hE) \geq \delta) + \dnorm{ \hProb_{X, \hD\circ \hH } - \hProb_{X, \hD \circ \hE}}_{\FLip} \mathbbm{1}(\cL_{recon}(\hD,\hE) < \delta)\\
    &\leq^{(\ref{Equation:ERMOfcLSDM})}  \sqrt{p+q} \mathbbm{1}(\cL_{recon}(\hD,\hE) \geq \delta) + \dnorm{ \hProb_{X, \hD\circ H } - \hProb_{X, \hD \circ \hE}}_{\FLip} \mathbbm{1}(\cL_{recon}(\hD,\hE) < \delta)\\
    &\leq\sqrt{p+q} \mathbbm{1}(\cL_{recon}(\hD,\hE) \geq \delta) + (K\vee 1) \dnorm{ \hProb_{X,  H } - \hProb_{X, \hE}}_{\FLip} \mathbbm{1}(\cL_{recon}(\hD,\hE) < \delta)\\
    &\leq\sqrt{p+q} \mathbbm{1}(\cL_{recon}(\hD,\hE) \geq \delta) + (K\vee 1) \dnorm{ \hProb_{X,  H } - \hProb_{X, \hE}}_{\FLip} 
\end{align*}
The first inequality is due to Assumption \ref{Assumption:Boundedness} which implies  $\dnorm{\Prob_{X, \hD\circ \hH } - \hProb_{X, \hD\circ \hH }}_{\FLip} \leq \sqrt{p+q}$. The second inequality is due to the empirical minimizing property (\ref{Equation:ERMOfcLSDM}). The third inequality is due to Assumption \ref{Assumption:Smoothness} and the Lipschitz transfer property of the 1-Wasserstein distance when $\hD$ is Lipschitz (See proof in Theorem \ref{Theorem:dLSDM}). Take infemum over $H \in \cH$ and further bound the second term using the optimal latent generator $H^*$ from Assumption \ref{Assumption:Smoothness}:
\begin{align*}
    \inf_{H \in \cH}\dnorm{ \hProb_{X,  H } - \hProb_{X,  \hE}}_{\FLip} &\leq  \inf_{H \in \cH} \frac{1}{n}\sum_{i=1}^n \dnorm{H(X_i,\eta_i) - \hE(Y_i)}_2\\
    &\leq \inf_{H \in \cH} \frac{1}{n} \sum_{i=1}^n  \left(\dnorm{H(X_i,\eta_i) - H^*(X_i,\eta_i)}_2 +   \dnorm{H^*(X_i,\eta_i) - \hE(Y_i))}_2\right)\\
    &= \inf_{H \in \cH} \frac{1}{n} \sum_{i=1}^n  \dnorm{H(X_i,\eta_i) - H^*(X_i,\eta_i)}_2 \\
    &\leq \inf_{H \in \cH} \dnorm{H - H^*}_{L_{\infty}([0,1]^{p+d})}
\end{align*}
The first inequality is due to the definition of 1-Wasserstein distance, and the fact that the empirical distribution of $\left(X_i,H(X_i,\eta_i),X_i, \hE(Y_i)\right)$ has a joint distribution with marginal distributions $\left( \hProb_{X,  H } , \hProb_{X,  \hE}\right)$. By Lemma \ref{chakraborty:theorem18} , there exists a constant $\epsilon_0$ and $\alpha_h$, such that if $\epsilon \in (0, \epsilon_0]$, a ReLU network $H'$ can be constructed with $\cL(H') \leq  \alpha_h \log(1/\epsilon_h), \cS(H') \leq   \alpha_h \log(1/\epsilon_h)\epsilon_h^{-\frac{p+d}{\beta_h}}$
such that $\dnorm{H' - H^*}_{L_\infty([0,1]^{p+d})} \leq \hspace{0.1em}\epsilon_h$ (by approximating each coordinate output $H_{i}^*$ and stacking the ReLU networks obtained parallel). Set the maximum depth and size of the neural network class $\cH$ as $L_h = \alpha_h \log(1/\epsilon_h) $ and $S_h = \alpha_h \log(1/\epsilon_h)\epsilon_h^{-\frac{p+d}{\beta_h}}$, then we have $\inf_{H \in \cH}\dnorm{ \hProb_{X,  H } - \hProb_{X,  \hE}}_{\FLip}  \leq \epsilon_h$. We choose $\epsilon_h \asymp n^{-\frac{1}{2 +\frac{p+d}{\beta_h}}}$. By the Markov inequality, 
\begin{align*}
    \E \Delta_{app} &\leq  \sqrt{p+q} \frac{\E \cL_{recon}(\hD,\hE)}{\delta} + (K\vee 1)n^{-\frac{1}{2 +\frac{p+d}{\beta_h}}}
\end{align*}

\textbf{Bounding $\E\Delta_{gen}^1$: }By Lemma \ref{Lemma:GeneralizationErrorLipschitzClass}, if $L_h\geq 3$, $S_h\geq 6q+2qL_h$, then
    $$\E_M\Delta_{gen}^1 \lesssim  n^{-\frac{1}{p+m}} + \sqrt{\frac{S_h L_h \left(\log(S_h) +L_h\right)\log(nm)}{n}}$$
Substitute $L_h=\alpha_h \log(1/\epsilon_h), S_h= \alpha_h \log(1/\epsilon_h)\epsilon_h^{-\frac{p+d}{\alpha_h}}$ with $\epsilon_h \asymp n^{-\frac{1}{2+\frac{p+d}{\beta_h}}}$, then
$$\E_M\Delta_{gen}^1
    \lesssim n^{-\frac{1}{p+m}} + (\log n)^2n^{-\frac{1}{2+\frac{p+d}{\beta_h}}}$$

\textbf{Bounding $\E\Delta_{gen}^2$: } By Lemma \ref{Lemma:GeneralizationErrorLipschitzClass}, if $L_e\geq 3$, $S_e\geq 6q+2qL_e$, then 
    $$\E_M\Delta_{gen}^2\lesssim  n^{-\frac{1}{p+m}} + \sqrt{\frac{S_e L_e \left(\log(S_e) +L_e\right)\log(nm)}{n}}$$
Substitute $L_e\lesssim \log n$, $S_e \lesssim   n^{\frac{s}{2\beta_e(\beta_d\wedge1)+s}}\log n$ (since we have $n \asymp N$), then,
$$\E_M\dnorm{\Prob_{X,Y} - \hProb_{X,Y}}_{\cK_2} \lesssim  n^{-\frac{1}{p+m}} + \left(\log n\right)^2 n^{-\frac{1}{2+\frac{s}{\beta_e(\beta_d\wedge 1)}}} $$

\textbf{Bounding $\E\cL_{recon}(\hD,\hE)$: } By Theorem \ref{Theorem:ReconstructionError} and $n \asymp N$, with the choices of $L_e, S_e,L_d, S_d$,
    $$\E_M \cL_{recon}(\hD,\hE) \lesssim (\log n)^2 n^{-\frac{1}{\max\{ 2 + \frac{m}{\beta_d},  2+\frac{s}{\beta_e(\beta_d\wedge 1)} \}}}$$

\textbf{Combining all: }
\begin{align*}
      \dnorm{\Prob_{X, \hD\circ \hH } - \Prob_{X, \hD \circ \hE}}_{\FLip} \lesssim \left(\log n\right)^2n^{-\frac{1}{\max\{ 2 + \frac{m}{\beta_d},  2+\frac{s}{\beta_e(\beta_d\wedge 1)} , 2+\frac{p+d}{\beta_h}, p+m\}}}.
\end{align*}

\subsubsection{Distribution matching error of dLSDM} 
The empirical risk minimizer $\hH$ satisfies
\begin{equation} \label{Equation:ERMOfdLSDM}
    \hH \in \argmin_{ H \in \cH} \dnorm{\hProb_{X,  H } - \hProb_{X,  \hE}}_{\FLip}.
\end{equation}
We consider the risk $\dnorm{\Prob_{X, \hD\circ \hH } - \Prob_{X, \hD \circ \hE}}_{\FLip}$ (i.e., the same risk as cLSDM). We perform the same decomposition as cLSDM:
\begin{align*}
    \dnorm{\Prob_{X, \hD\circ \hH } - \Prob_{X, \hD \circ \hE}}_{\FLip} &\leq \dnorm{\Prob_{X, \hD\circ \hH } - \hProb_{X, \hD\circ \hH }}_{\FLip} + \dnorm{ \hProb_{X, \hD\circ \hH } - \hProb_{X, \hD \circ \hE}}_{\FLip} + \dnorm{\hProb_{X, \hD \circ \hE} - \Prob_{X, \hD \circ \hE}}_{\FLip} \\
    &:= \delta_1 + \Delta_{app} +\delta_2. 
\end{align*}
For $\Delta_{app}$, we reverse the order of Lipschitz transfer argument and empirical risk minimizing argument. For any $H \in \cH$:
\begin{align*}
    \Delta_{app} &= \dnorm{ \hProb_{X, \hD\circ \hH } - \hProb_{X, \hD \circ \hE}}_{\FLip} \\
    &= \dnorm{ \hProb_{X, \hD\circ \hH } - \hProb_{X, \hD \circ \hE}}_{\FLip}\mathbbm{1}(\cL_{recon}(\hD,\hE) \geq \delta) +\dnorm{ \hProb_{X, \hD\circ \hH } - \hProb_{X, \hD \circ \hE}}_{\FLip} \mathbbm{1}(\cL_{recon}(\hD,\hE) < \delta) \\
    &\leq \sqrt{p+q} \mathbbm{1}(\cL_{recon}(\hD,\hE) \geq \delta) + \dnorm{ \hProb_{X, \hD\circ \hH } - \hProb_{X, \hD \circ \hE}}_{\FLip} \mathbbm{1}(\cL_{recon}(\hD,\hE) < \delta)\\
    &\leq  \sqrt{p+q} \mathbbm{1}(\cL_{recon}(\hD,\hE) \geq \delta) + (K\vee 1) \dnorm{ \hProb_{X, \hH } - \hProb_{X, \hE}}_{\FLip} \mathbbm{1}(\cL_{recon}(\hD,\hE) < \delta)\\
    &\leq^{(\ref{Equation:ERMOfdLSDM})}  \sqrt{p+q} \mathbbm{1}(\cL_{recon}(\hD,\hE) \geq \delta) + (K\vee 1) \dnorm{ \hProb_{X,  H } - \hProb_{X, \hE}}_{\FLip} \mathbbm{1}(\cL_{recon}(\hD,\hE) < \delta)\\
    &\leq\sqrt{p+q} \mathbbm{1}(\cL_{recon}(\hD,\hE) \geq \delta) + (K\vee 1) \dnorm{ \hProb_{X,  H } - \hProb_{X, \hE}}_{\FLip} 
\end{align*}
The rest of the proof is exactly the same as cLSDM. Therefore, dLSDM has the same rate as cLSDM.

\subsubsection{Distribution matching error when Step 1 is pretrained using only unpaired data}
Consider $(\hD, \hE)$ trained only on the unpaired dataset $\cU$. That is, $\widehat{\cL}_{\text{recon}}(D,E) := \frac{1}{N} \sum_{i=n+1}^{n+N} \dnorm{Y_i - D\circ E(Y_i)}_2$, and that $\cD,\cE$ are set with $L_e, L_d= \alpha \log N$, $S_d = \alpha N^{\frac{m}{2\beta_d+m}}\log N$, and  $S_e = \alpha N^\frac{s}{2\beta_e(\beta_d\wedge1)+s}\log N$. 

For cLSDM, following the same proof as previous, we have 
\begin{equation}
        \dnorm{\Prob_{X, \hD\circ \hH } - \Prob_{X, \hD \circ \hE}}_{\FLip} \leq  \delta_1 + \Delta_{app} +\delta_2,
\end{equation}
where
\begin{align*}
    \delta_1 &=  \dnorm{\Prob_{X, \hD\circ \hH } - \hProb_{X, \hD\circ \hH }}_{\FLip} \\
     \delta_2 &= \dnorm{\hProb_{X, \hD \circ \hE} - \Prob_{X, \hD \circ \hE}}_{\FLip} \\
     \Delta_{app} &= \dnorm{ \hProb_{X, \hD\circ \hH } - \hProb_{X, \hD \circ \hE}}_{\FLip} .
\end{align*}
For $\delta_1$ and $\Delta_{app}$, we use the same decomposition, while accounting for the fact that the Step 1 rate now depends on the unpaired sample size $N$ instead of $n+N$. Thus, without the $n \asymp N$ requirement, 
$$\E_M \cL_{recon}(\hD,\hE) \lesssim (\log N)^2 N^{-\frac{1}{\max\{ 2 + \frac{m}{\beta_d},  2+\frac{s}{\beta_e(\beta_d\wedge 1)} \}}},$$
For $\delta_2$, use the same truncation argument through $\mathbbm{1}(\cL_{recon}(\hD,\hE) \geq \delta) $, but with a different treatment for the generalization error term $\dnormF{\Prob_{X,\hE} - \hProb_{X, \hE}}$. Conditioning on $\cU$, $\hE$ is fixed. So $Z = \hE(Y)$ is a random variable independent of $\cP$. Then,
\begin{align*}
        & \dnormF{\Prob_{X,\hE} - \hProb_{X, \hE}}\\
    &= \sup_{f \in \FbLip} \biggl\{\E_{X,Y} f(X, \hE(Y)) - 
    \frac{1}{n}\sum_{i=1}^n f(X_i,  \hE (Y_i))\biggr\}\\
    &= \dnorm{\Prob_{X,Z} - \hProb_{X,Z}}_{\FbLip}\\
    &\leq \dnorm{\Prob_{X,Z} - \hProb_{X,Z}}_{\cF^1}
\end{align*}
By Lemma \ref{lu2020} and Assumption~\ref{Assumption:Boundedness}, assuming $p+m \geq 3$, 
\begin{align*}
    \E_\cP \dnorm{\Prob_{X,Z} - \hProb_{X,Z}}_{\cF^1}&= \E_{\cP} W_1(\Prob_{X,Z}, \hProb_{X,Z})\\
    &\lesssim  n^{-\frac{1}{p+m}}.\\
    \E_M   \dnorm{\Prob_{X,Z} - \hProb_{X,Z}}_{\cF^1}&= \E_\cU \E_{\cP} \dnorm{\Prob_{X,Z} - \hProb_{X,Z}}_{\cF^1}\\
    &\lesssim n^{-\frac{1}{p+m}}.
\end{align*}
Thus, without the $n\asymp N$ requirement to control the generalization error $\dnormF{\Prob_{X,\hE} - \hProb_{X, \hE}}$, we combine all terms as previous to obtain
$$ \dnorm{\Prob_{X, \hD\circ \hH } - \Prob_{X, \hD \circ \hE}}_{\FLip} \lesssim (\log N)^2 N^{-\frac{1}{\max\{ 2 + \frac{m}{\beta_d},  2+\frac{s}{\beta_e(\beta_d\wedge 1)} \}}} +  \left(\log n\right)^2n^{-\frac{1}{\max\{ 2+\frac{p+d}{\beta_h}, p+m\}}}.$$
For dLSDM, as previous, we reverse the order of Lipschitz transfer argument and empirical risk minimizing argument, to arrive at the same rate.

\subsection{Proof of Theorem \ref{Theorem:RangeProximity}}
Suppose the conditions of Corollary \ref{Corollary:FullReconstruction} hold. Let $g(z) = \text{dist}(\hD(z), \cY)$ and assume there exists a constant $\rho \in (0,1)$ such that $\int_{B}p(\hat{z})d\hat{z}\geq \rho, B = \bigl\{z\in \cZ:g(z)\geq \sup_{z \in \cZ}g(z)/2\bigr\}$ almost surely. Then,
$$ \E_M \sup_{z \in \cZ} \,\text{dist}(\hD(z), \cY) \lesssim \rho^{-1} (n+N)^{-1/\xi} \log^2(n+N),$$
where $\xi = 2+m\vee s$, $M = \cP \cup \cU$ and $s>d_\cY$ is arbitrary. 

\textbf{Proof:\\}
Let $g(z) = \text{dist}(\hD(z), \cY), z \in \cZ$. For any $y \in \cY$, let $z_y = \hE(y)$, and 
\begin{align*}
    \dnorm{\hD(\hE(y)) - y}_2&= \dnorm{\hD(z_y) - y}_2\\
    &\geq \inf_{y' \in \cY} \dnorm{\hD(z_y) - y'}_2\\
    &= g(z_y).
\end{align*}
Let $c^* = \sup_{z \in \cZ} g(z)$ be the worse case distance of a decoded point to the support $\cY$, and $B =\{z: g(z) \geq c^*/2\}$ be the set of bad points in the latent space. Then
\begin{align*}
    \E_Y \dnorm{\hD \circ \hE(Y) - Y}_2&\geq \E_{\widehat{Z}} g(\widehat{Z})\\
   & \geq \int_B g(\hat{z}) p(\hat{z}) dz\\
    &\geq \frac{c^*}{2} \Prob(B)\\
    &\geq \frac{\rho}{2}c^*\\
    c^*&\leq \frac{2}{\rho}\E_Y \dnorm{\hD\circ\hE(Y)-Y}_2.
\end{align*}
The first inequality is due to the law of unconscious statistician. Take expectation w.r.t. the data $M$ and apply Corollary \ref{Corollary:FullReconstruction} gives the result.

\section{Supporting Results}
\begin{lemma}\label{Lemma:WassersteinConvergence}
   \citep[Theorem 6.9,][]{villani2008} Let $\{\Prob_n\}_{n \in \mathbb{N}}$ and $ \Prob$ be probability distributions on $\R^d$ with finite first moment, then   $W_1\left(\Prob_n, \Prob\right) \to 0$   is equivalent to  $\Prob_n \xrightarrow{D} \Prob$ and $\E_{X \sim \Prob_n}\dnorm{X}_2 \to \E_{X \sim \Prob}\dnorm{X}_2$.
\end{lemma}

\begin{lemma}\label{Lemma:NnoiseOutsourcingLemma}
\citep[Theorem 5.10,][]{Kallenberg} Let $(X,Y)$ be a random pair taking values in $\cX \times \cY$ with joint distribution $\Prob_{X,Y}$. Suppose $\cY$ is a standard Borel space. Then there exist a random variable $\eta \sim \cN(0, I_d)$, for any $d\geq 1$, and a Borel-measurable function $G: \cX \times \R^d \to \cY$ such that $\eta$ is independent of $X$ and 
$$(X,Y) = (X, G(X,\eta)) \quad a.s.$$
\end{lemma}
While the original lemma assumes $\eta \sim U([0,1])$, it can be extended to $\eta =(\eta_1,...,\eta_d) \sim \cN(0, I_d)$, for any $d\geq 1$, via $\Phi^{-1}(\eta_1) \sim U([0,1])$ and ignore the remaining $\eta_i$, where $\Phi$ is the cumulative distribution function of standard Gaussian.

\begin{lemma}
    \label{Lemma:JointDistributionMatching}\citep[Lemma 2.2,][]{DCG} Suppose that $\eta$ is independent of $X$. Then $G(x,\eta) \sim \Prob_{Y|X=x}, \text{ for $\Prob_X$-almost every $x$}$ if and only if 
    $$\left(X,G(X,\eta)\right) \eqd (X,Y)$$
\end{lemma}

\begin{lemma}
    \label{Lemma12Huang}\citep[Lemma 12,][]{GANError} For any function class $\cF$, assume $\sup_{f\in \cF}\dnorm{f}_\infty \leq B$, then we have the following bounds
\begin{align*}
    \E_{X_{1:n}} \dnorm{\mu - \hat{\mu}_n}_\cF & \leq   2\E _{X_{1:n}} \cR(\cF_{|X_{1:n}})\\
    &\leq 8 \E _{X_{1:n}}\inf_{0<\delta<B/2}\left(\delta + \frac{3}{\sqrt{n}}\int_\delta^{B/2} \sqrt{\log \cN(\epsilon, \cF_{|X_{1:n}},\dnorm{\cdot}_\infty )} d \epsilon\right)
\end{align*}
where $\cR(A):= \E_{\xi}\left[ \sup_{(a_1,\dots,a_n)\in A}\frac{1}{n} \sum_{i=1}^n \xi_ia_i\right]$ for any $A\subseteq \R^n$ and $\xi_i$ are  iid Rademacher variables, $\xi = (\xi_1,\dots,\xi_n)$.
    % $$\E_{X_{1:n}} \dnorm{\mu - \hat{\mu}_n}_\cF \leq 8 \E _{X_{1:n}}\inf_{0<\delta<B/2}\left(\delta + \frac{3}{\sqrt{n}}\int_\delta^{B/2} \sqrt{\log \cN(\epsilon, \cF_{|X_{1:n}},\dnorm{\cdot}_\infty )} d \epsilon\right)$$
\end{lemma}

\begin{lemma}
    \label{lu2020}\citep[Proposition 3.1,][]{lu2020}
Assume that probability distribution $\pi$ on $\R^d$ satisfy $M_3 = \E_{X\sim \pi } |X|^3 < \infty$, and let $\hat{\pi}_n$ be its empirical distribution of sample size $n$. Then there exists a constant $C$ depending on $M_3$ such that
$$
\E W_1(\pi, \hat{\pi}_n)\leq C
\begin{cases}
n^{-\frac{1}{2}}, \hspace{4.2em} d=1\\
n^{-\frac{1}{2}}\log n, \hspace{2em} d=2\\
n^{-\frac{1}{d}}, \hspace{4.2em} d\geq 3
\end{cases}
$$
\end{lemma}

\begin{lemma}
\label{metricentropyHolder} \citep{kolmogorov1961} 
    The $\epsilon$-covering number of $\cH^{\beta}([0,1]^d, \R,1)$ can be bounded as,
    $$\log \cN(\epsilon, \cH^{\beta}([0,1]^d,\R,1), \dnorm{\cdot}_\infty)\lesssim \epsilon^{-d/\beta}$$
    
\end{lemma}

 \begin{lemma}
    \label{metricentropybound} \citep[Lemma 20,][]{ChakrabortyBarlett2024} Suppose that $n\geq 6$ and $\cF:\R^d \to \R^{d'}$ be a class of bounded neural networks with depth at most $L$ and the number of weights at most $S$. Furthermore, the activation functions are piece-wise polynomial activation with the number of pieces and degree at most $k \in \mathbb{N}$. Then, there exists positive constant $\theta$ (that might depend on $d$ and $d'$), such that if  $n\geq \theta(S+6d'+2d'L)(L+3)(\log(S+6d'+2d'L) + L + 3)$,
     $$\log \cN(\epsilon, \cF_{|X_{1:n}}, \dnorm{\cdot}_\infty)\lesssim (S+6d'+2d'L)(L+3)(\log(S+6d'+2d'L) + L + 3)\log\left(\frac{nd'}{\epsilon}\right)$$
 \end{lemma}

\begin{lemma} \label{chakraborty:theorem18}
\citep[Theorem 18,][]{ChakrabortyBarlett2024} Let $f \in \cH^\beta(\R^d, \R, C)$, where $C>0$, and $\Prob_Y$ be a measure on $\R^d$ with support $\cY  \subseteq [0,1]^d$. Then for any $s>d_\cY$, there exists constants $\epsilon_0$ (which may depend on $\cY$) and $\alpha$ (which may depend on $\beta, d, \text{ and } C$), such that if $\epsilon\in (0, \epsilon_0]$, a ReLU network $\hat{f}$ can be constructed with $\cL(\hat{f})\leq \alpha \log (1/\epsilon)$ and $\cS(\hat{f}) \leq \alpha \log(1/\epsilon)\epsilon^{-s/\beta}$, satisfying the condition $\dnorm{f - \hat{f}}_{L_{\infty}(\cY)} \leq \epsilon$. Here, $d_\cY$ is the upper Minkowski dimension of $\cY$. 
\end{lemma}

\begin{lemma} \label{UnveilLDM:lemmaD4}
\citep[Lemma D.4,][]{ConditionalDiffusionConvergence} Let $p_0$ be a probability distribution, and let $Y = \{Y_t\}_{t \in [0,T]}$ and $Y' = \{Y'_t\}_{t \in [0,T]}$ be two stochastic processes that satisfy the following SDEs:
\begin{align*}
    dY_t &= s(Y_t,t)dt + dW_t, \quad Y_0 \sim p_0\\
    dY'_t &= s'(Y'_t,t)dt + dW_t, \quad Y'_0 \sim p_0
\end{align*}
We further define the distributions of $Y_t$ and $Y_t'$ by $p_t$ and $p_t'$. Suppose that
$$\int_x p_t(x) \dnorm{(s - s')(x,t)}^2 dx \leq C$$
for any $t\in [0,T]$. Then we have
$$\mathbb{D}_{KL}(p_T, p_T')\leq \int_0^T\frac{1}{2}\int_x p_t(x)\dnorm{(s-s')(x,t)}^2dxdt.$$
\end{lemma}

\begin{lemma} \label{UnveilLDM:lemmaD5}
\citep[Lemma D.5,][]{ConditionalDiffusionConvergence} Suppose that (1) $p(z|x) \in \cH^{\beta}(\cZ \times [0,1]^p, \R, B)$ for some constants $\beta, B > 0$ (2)  $p(z|x)\leq C_1 \exp(-C_2 \dnorm{z}_2^2/2) $ for all $x $ and some constants $C_1,C_2>0$. We have for any $x \in [0,1]^p$, 
$$\dnorm{\Prob_{Z|X=x} - \Prob_{Z_{t_0}|X=x}}_{TV}\lesssim\sqrt{t_0} \log^{\frac{m+1}{2}}\left(\frac{1}{t_0}\right).$$
\end{lemma}

\begin{lemma}
 \label{Lemma:KantorovichRubinsteinDuality} The 1-Wasserstein distance between two probability measures $\mu,\nu$ admits the duality form:
 \begin{equation} 
    W_1(\mu,\nu) =  \sup_{f\in \mathcal{F}^1}\E_{x\sim\mu} f(x) - \E_{y\sim \nu}f(y)
\end{equation}
where $\mathcal{F}^1  = \Lip(\Omega, 1)$  is the 1-Lipschitz class. 
\end{lemma}

\section{Additional Experiments}
\subsection{Simulation}
\begin{figure}[ht]
\begin{minipage}{0.5\linewidth}
\centering
\includegraphics[scale=0.55]{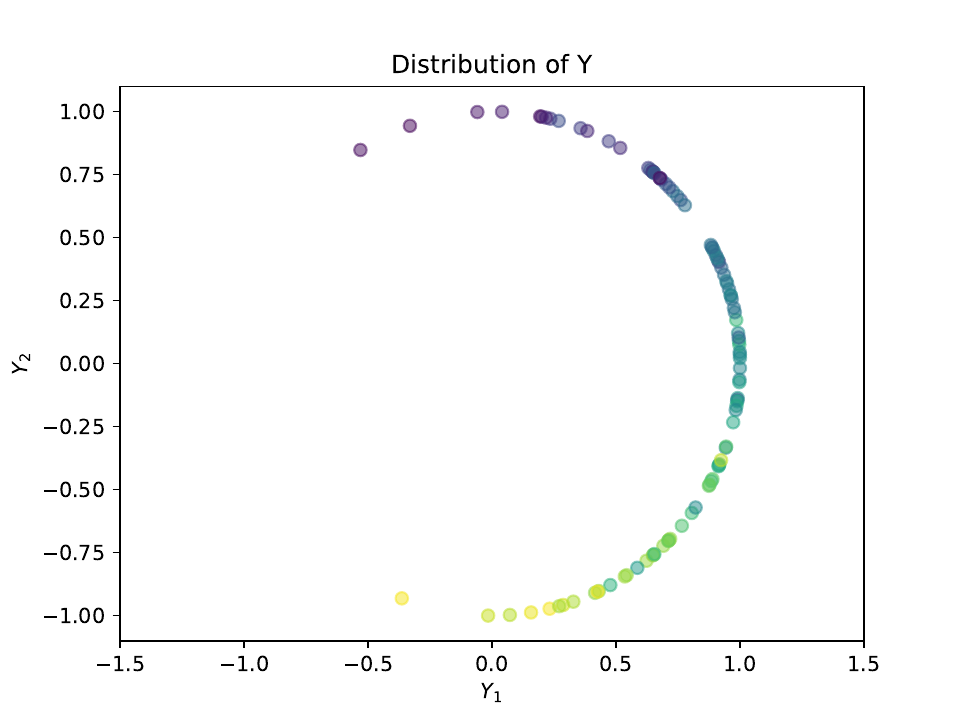}
\end{minipage}
\begin{minipage}{0.44\linewidth}
\centering
\includegraphics[scale=0.55]{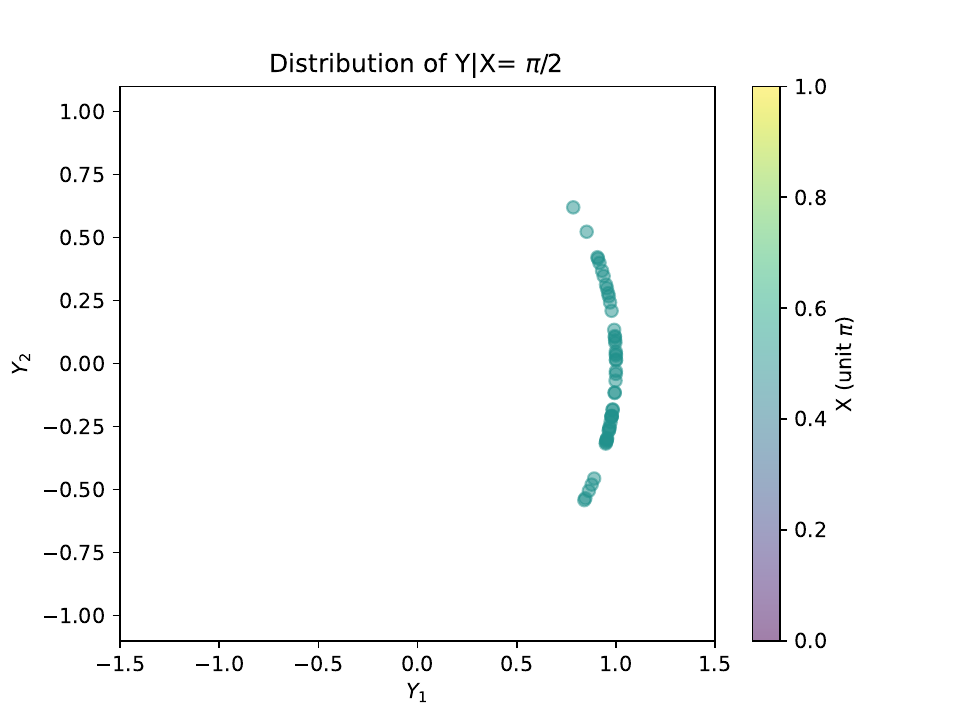}
\end{minipage}
\vspace{5pt}
\small
\caption{Illustration of $Y$ model.}\label{Response:Simulation:Model}
\end{figure}

We conducted a new simulation experiment with a known conditional density as follows.
\begin{align*}
X_i &\sim \text{Unif}\left([0, \pi]\right) \in \mathbb{R},\\
Y_i &= \bigl(\sin (X_i + \epsilon_i), \cos(X_i + \epsilon_i)\bigr) \in \mathbb{R}^2,\\
\epsilon_i &\sim \mathcal{N}\left(0, \pi^2 / 100\right).
\end{align*}
The support $\mathcal{Y}$ of $Y_i$ is the unit circle, which constitutes a one-dimensional manifold embedded in $\mathbb{R}^2$. Accordingly, we set the latent dimension to $m = 1$. Conditional on $X_i = x$, $Y_i$ follows a continuous distribution along the line obtained by unrolling the unit circle. While the exact analytical form of this conditional distribution is difficult to express in closed form, it can nonetheless be sampled straightforwardly for evaluation purposes. A visualization of the model is provided in Figure~\ref{Response:Simulation:Model}.

%This aligns well with our theoretical setting, which assumes that $Y$ has low‑dimensional intrinsic structure.
\FloatBarrier
\FloatBarrier

\paragraph{Ablation studies.}
Using this model, we examine how the sample sizes $(n,N)$ affect generation quality. Performance is measured by the 1‑Wasserstein distance between joint distributions $W_1\left(\hProb_{X, Y}, \hProb_{X, \hD \circ \hH(X,\eta) }\right)$ evaluated on a test set of size $500$ (computing the exact $W_1$ distance on a large sample size is extremely slow). We consider two scenarios:
\begin{enumerate}
\item[(1)] Fixed total sample size $n+N=1000$, varying paired proportion $n$.
\item[(2)] Fixed paired sample size $n=50$, varying unpaired size $N$.
\end{enumerate}

These experiments allow us to isolate the contributions of paired and unpaired data to the final generation quality. Results are shown below.

\begin{figure}[ht]
\centering
\begin{tikzpicture}[scale=0.825] 
\begin{axis}[
    name=ax1,
    title={Fix n+N=1000, vary n},
    width=10cm,
    height=7cm,
    xlabel={$n$},
    ylabel={$W_1\left(\hProb_{X, Y}, \hProb_{X, \widehat{G}}\right)$},
    xmin=0, xmax=1100,
    ymin=0.10, ymax=0.55,
    xtick={25, 100,  300, 500, 700, 900, 1000},
    ytick={0.15, 0.25, 0.35, 0.45, 0.55},
    legend pos=north east,
    tick style={thin},
    ticklabel style={font=\footnotesize},
    label style={font=\small},
    title style={font=\small\bfseries},
    legend style={font=\small, draw=none, fill=white, fill opacity=0.9, text opacity=1},
    every axis plot/.append style={thick},
    scaled y ticks=false,
    y tick label style={
        /pgf/number format/fixed,
        /pgf/number format/.cd,
        fixed zerofill,
        precision=2
    },
]
\addplot[
    color=red!90!black,
    mark=square*,
    ]
    coordinates {(25,0.454957012)(50,0.320769331)
(100, 0.240205915)(300,0.143647621)(500,0.151171115)(700,0.137425305)(900,0.150852741)(1000,0.12539817)
   };
% \addplot[
%     color=blue!90!black,
%     mark=triangle*,
%     ]
%     coordinates {(2.5, 25.50) (5, 24.15) (7.5,23.59) (10, 23.08) (12.5, 20.98) (15, 21.02) (17.5, 20.31) (20, 20.56) (22.5, 20.53) (25,20.14) (27.5, 19.93) (30, 19.76)
% };
% \label{ablation:dLSDM}
\end{axis}

\hfill

\begin{axis}[
        at={(ax1.south east)},
        xshift=1.5cm,
            width=10cm,
            height=7cm,
    title={Fix n=250, vary n+N},
    xlabel={$n+N$ (hundreds)},
    ylabel={},
    xmin=0, xmax=1200,
    ymin=0.10, ymax=0.65,
    xtick={100,  300, 500, 700, 900, 1000},
    ytick={0.15, 0.25, 0.35, 0.45, 0.55, 0.65},
    legend pos=north east,
    tick style={thin},
    ticklabel style={font=\footnotesize},
    label style={font=\small},
    title style={font=\small\bfseries},
    legend style={font=\small, draw=none, fill=white, fill opacity=0.9, text opacity=1},
    every axis plot/.append style={thick},
    scaled y ticks=false,
    y tick label style={
        /pgf/number format/fixed,
        /pgf/number format/.cd,
        fixed zerofill,
        precision=2
    },
]
\addplot[
    color=red!90!black,
    mark=square*,
    ]
    coordinates {
(100, 0.1725505623574010)(300,0.2019259560914525)(500,0.15498484250081784)(700,0.134892951127615)(900,0.1671766055277750)(1100, 0.12980042951955006)
    };

% \addplot[
%     color=blue!90!black,
%     mark=triangle*,
%     ]
%     coordinates {
%     (2.5,107.43380)(5,101.90734)(7.5,72.22736)(10, 56.60303)(12.5,47.28194)(15,38.780715)(17.5,31.69911)(20, 29.66109) (22.5, 29.83943)(25,26.78973)(27.5,25.04384)(30,24.0118)};\label{plot:dLSDM}

\end{axis}
\end{tikzpicture}
\vspace*{-10pt} 
\small
\caption{Ablation study on simulation experiment ($m=1$, cLSDM). 
}\label{Simulation:Ablation}
\end{figure}
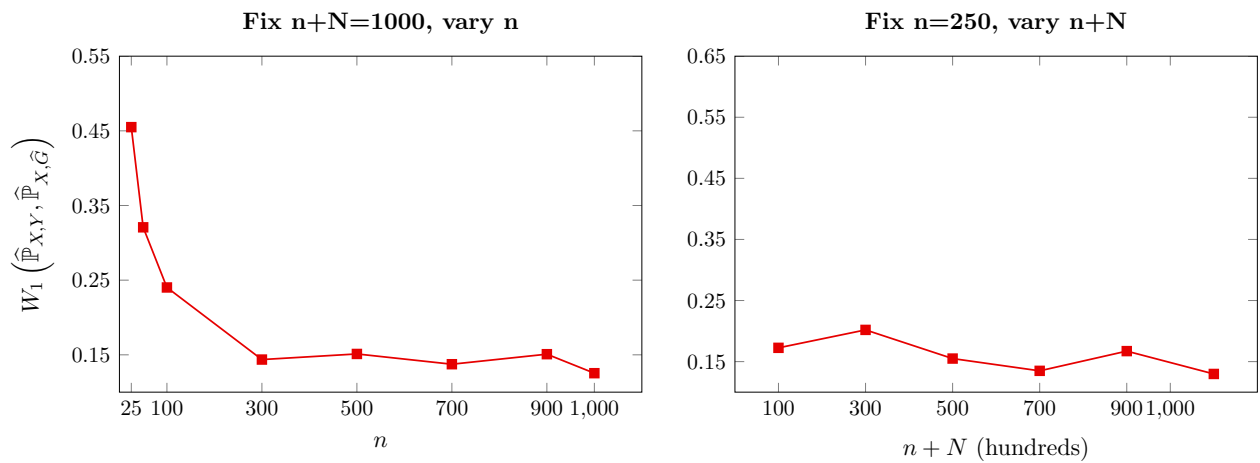

\begin{figure}[ht]
\centering
\includegraphics[scale=0.44]{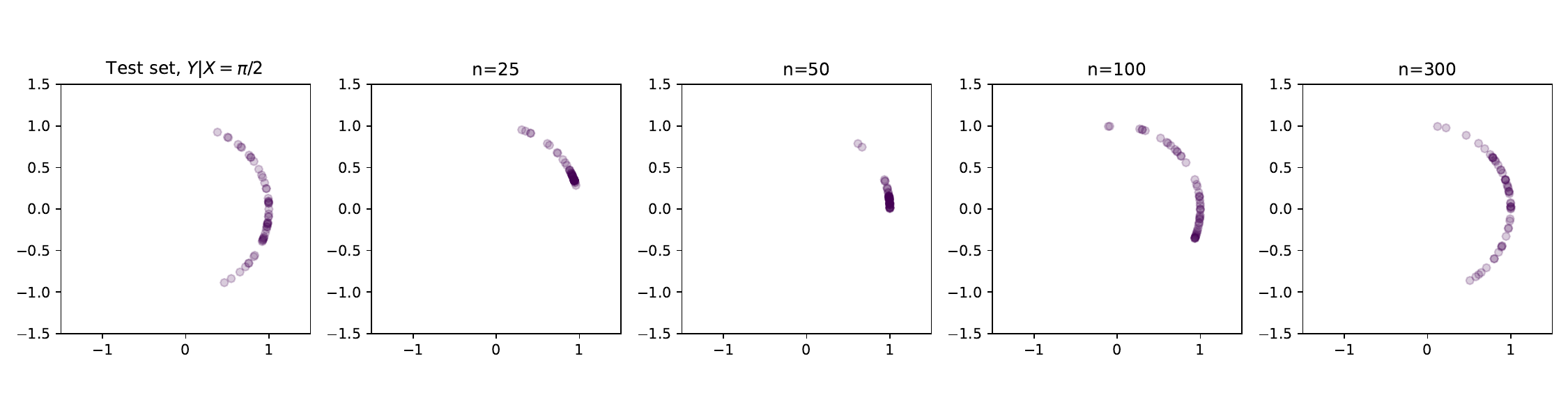}
\small
\vspace{-20pt}
\caption{Visualization of the distribution of $\hH(\pi /2, \eta)$ in simulation experiment. (Vary $n$, fix $n+N=1{,}000$, cLSDM, $m =1$). }\label{Simulation:Varyn}
\end{figure}

\begin{figure}[ht]
\centering
\includegraphics[scale=0.44]{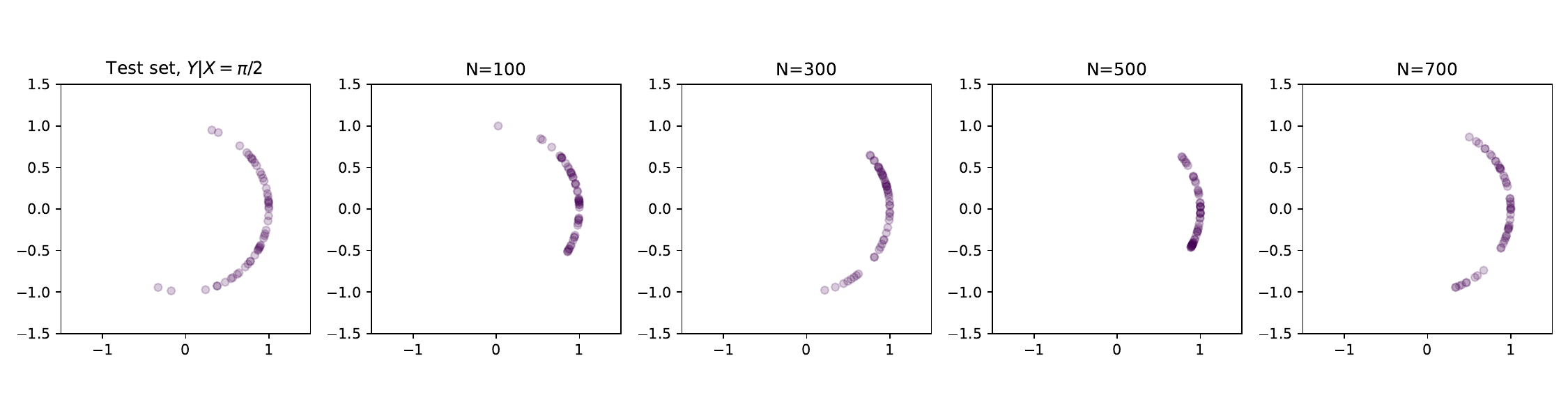}
\small
\vspace{-20pt}
\caption{Visualization of the distribution of $\hH(\pi /2, \eta)$ in simulation experiment. (Fix $n=250$, vary $n+N$, cLSDM, $m =1$). }\label{Simulation:VarynpN}
\end{figure}
\FloatBarrier

Moreover, we considered shifted unpaired responses generated as follows:
\begin{align*}
Y^{\cU} &= \bigl(\sin (X_i + \epsilon_i) + c_1, \cos(X_i + \epsilon_i) + c_1\bigr) \in \mathbb{R}^2,\\
\epsilon_i &\sim \mathcal{N}\left(c_2, \pi^2 / 100\right),
\end{align*}
where the constant $c_1$ shifts the support $\cY$ and $c_2$ shifts the conditional distribution. The paired response $Y^\cP$ sets $c_1=c_2=0$. The autoencoder is trained on $Y^\cP \bigcup Y^\cU$, while the distribution matching step and the test set uses $(X, Y^\cP)$. Ablation studies varying $c_1$ and $c_2$ are presented below.

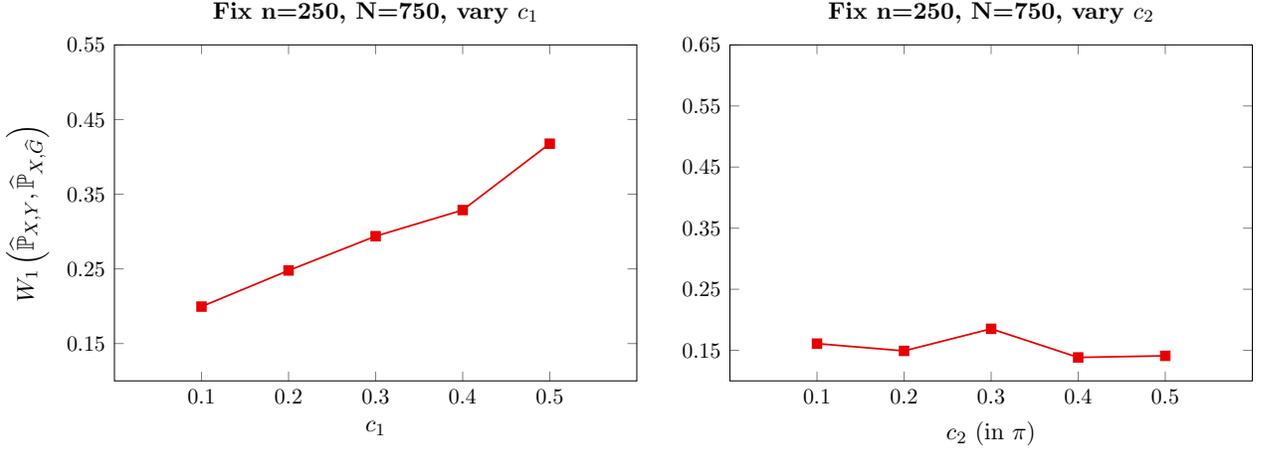
\begin{figure}[H]
\centering
\begin{tikzpicture}[scale=0.825] 
\begin{axis}[
    name=ax1,
    title={Fix n=250, N=750, vary $c_1$},
    width=10cm,
    height=7cm,
    xlabel={$c_1$},
    ylabel={$W_1\left(\hProb_{X, Y}, \hProb_{X, \widehat{G}}\right)$},
    xmin=0, xmax=0.6,
    ymin=0.10, ymax=0.55,
    xtick={0.1, 0.2, 0.3, 0.4, 0.5},
    ytick={0.15, 0.25, 0.35, 0.45, 0.55},
    legend pos=north east,
    tick style={thin},
    ticklabel style={font=\footnotesize},
    label style={font=\small},
    title style={font=\small\bfseries},
    legend style={font=\small, draw=none, fill=white, fill opacity=0.9, text opacity=1},
    every axis plot/.append style={thick},
    scaled y ticks=false,
    y tick label style={
        /pgf/number format/fixed,
        /pgf/number format/.cd,
        fixed zerofill,
        precision=2
    },
]
\addplot[
    color=red!90!black,
    mark=square*,
    ]
    coordinates {(0.1,0.1995)(0.2,0.24786)(0.3,0.2938)(0.4,0.3288)(0.5,0.41768797833828913)
   };
\end{axis}

\hfill

\begin{axis}[
        at={(ax1.south east)},
        xshift=1.5cm,
            width=10cm,
            height=7cm,
    title={Fix n=250, N=750, vary $c_2$},
    xlabel={$c_2$ (in $\pi$)},
    ylabel={},
    xmin=0, xmax=0.6,
    ymin=0.10, ymax=0.65,
    xtick={0.1, 0.2, 0.3, 0.4, 0.5},
    ytick={0.15, 0.25, 0.35, 0.45, 0.55, 0.65},
    legend pos=north east,
    tick style={thin},
    ticklabel style={font=\footnotesize},
    label style={font=\small},
    title style={font=\small\bfseries},
    legend style={font=\small, draw=none, fill=white, fill opacity=0.9, text opacity=1},
    every axis plot/.append style={thick},
    scaled y ticks=false,
    y tick label style={
        /pgf/number format/fixed,
        /pgf/number format/.cd,
        fixed zerofill,
        precision=2
    },
]
\addplot[
    color=red!90!black,
    mark=square*,
    ]
    coordinates {
(0.1, 0.1610)(0.2,0.1491)(0.3, 0.18522)(0.4, 0.13839771)(0.5,0.14102048571998183)
    };

% \addplot[
%     color=blue!90!black,
%     mark=triangle*,
%     ]
%     coordinates {
%     (2.5,107.43380)(5,101.90734)(7.5,72.22736)(10, 56.60303)(12.5,47.28194)(15,38.780715)(17.5,31.69911)(20, 29.66109) (22.5, 29.83943)(25,26.78973)(27.5,25.04384)(30,24.0118)};\label{plot:dLSDM}

\end{axis}
\end{tikzpicture}
\vspace*{-10pt} 
\small
\caption{Ablation study on simulation experiment ($m=1$, cLSDM). 
}\label{ShiftSimulation:Ablation}
\end{figure}

\begin{figure}[H]
\centering
\includegraphics[scale=0.44]{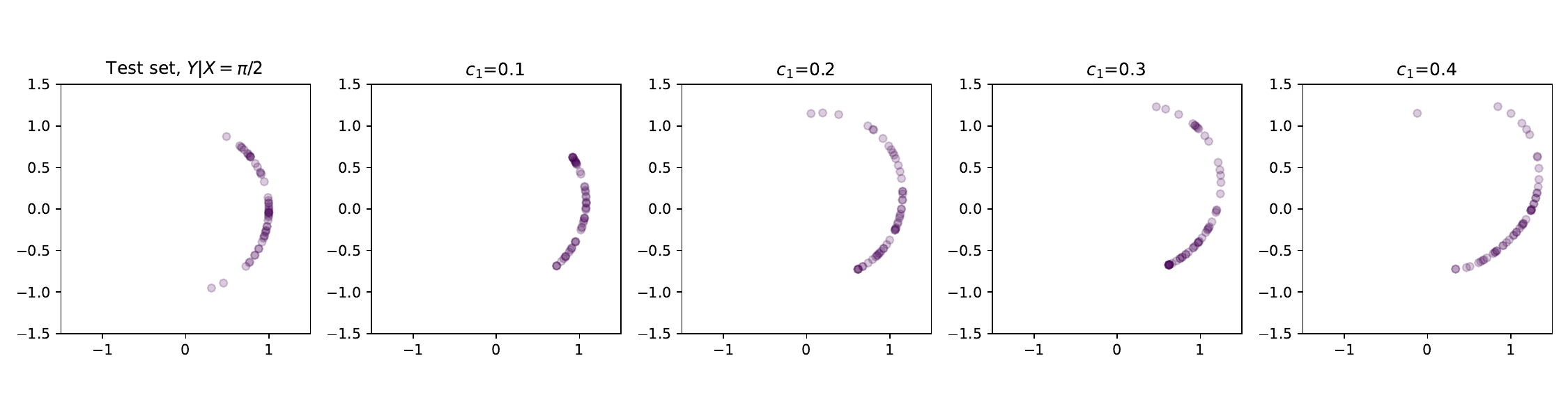}
\small
\vspace{-20pt}
\caption{Visualization of the distribution of $\hH(\pi /2, \eta)$ in simulation experiment. Distribution of $Y^\cU$  is shifted via $c_1$ ( $n=250$, $N=750$, cLSDM, $m =1$). }\label{ShiftSimulation:Varyn}
\end{figure}

\begin{figure}[H]
\centering
\includegraphics[scale=0.44]{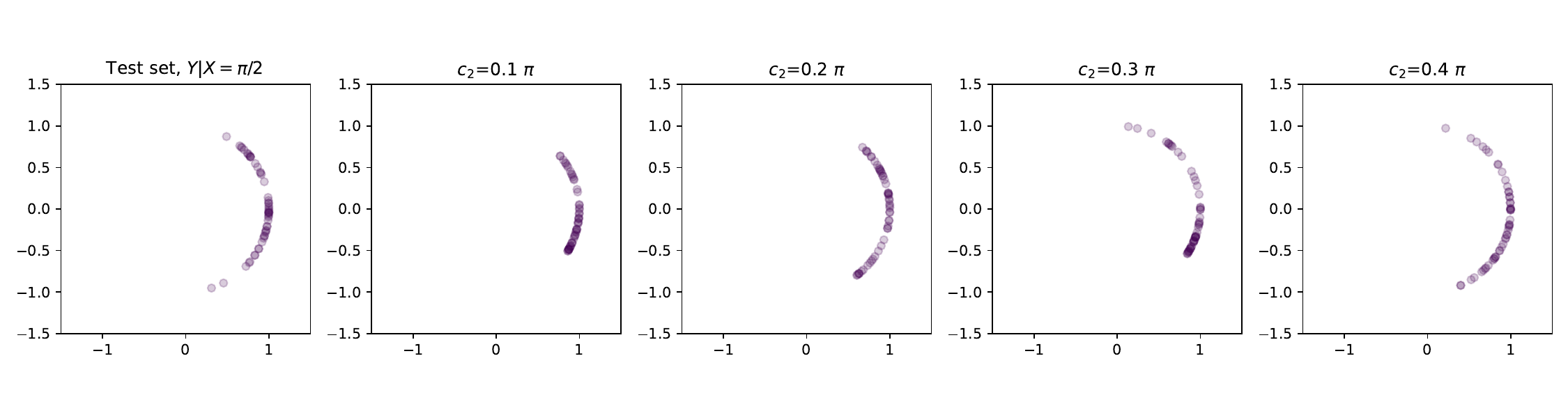}
\small
\vspace{-20pt}
\caption{Visualization of the distribution of $\hH(\pi /2, \eta)$ in simulation experiment. Distribution of $Y^\cU$  is shifted via $c_2$ ($n=250$, $N=750$, cLSDM, $m =1$). }\label{ShiftSimulation:VarynpN}
\end{figure}
\FloatBarrier

\section{Duality-Based Implementation}\label{dualityimplementation}
In this section, we describe the duality-based implementation of LSDM, where the functions $D_\theta$, $E_\phi$, $H_\gamma$ and $f_\omega$ are represented by deep neural networks parameterized by $\theta$, $\phi$, $\gamma$, and $\omega$ respectively.

For the representation learning step, it is formulated as an optimization over the network parameters. The corresponding empirical objective is
\begin{equation}\label{empobj:1}
    (\hat{\theta}, \hat{\phi}) = \argmin_{\theta, \phi} \hspace{1em} \frac{1} {n+N}\sum_{i=1}^{n+N} \dnorm{Y_i - D_{\theta}(E_\phi(Y_i))}_2 \hspace{0.5em}
\end{equation}

For the conditional generation step, we use the gradient penalty algorithm \citep{WGANGP} to enforce the Lipschitz constraint. 
The objective is formulated as:
\begin{equation}\label{empobj:2a}
    \hat{\gamma} = \argmin_{\gamma} \sup_{\omega} \frac{1}{n}\sum_{i=1}^n \biggl\{ f_\omega(X_i, T_\gamma(X_i, \eta_i)) - f_\omega(X_i,Z_i) + \lambda \left(\dnorm{\nabla_{(x,z)}f_\omega(X_i,Z_i)}_2-1\right)^2\biggr\}
\end{equation}

For cLSDM, $T_\gamma = D_{\hat{\theta}}\circ H_{\gamma}$ and $Z_i = D_{\hat{\theta}} \circ E_{\hat{\phi}} (Y_i)$, whereas for dLSDM, $T_\gamma =  H_{\gamma}$ and $Z_i = E_{\hat{\phi}} (Y_i)$. For both models, $\nabla_{(x,z)} f_\omega(X_i,Z_i)$ is the gradient of $f_\omega$ at $(X_i,Z_i)$, and $\lambda$ is a hyper-parameter controlling the strength of the gradient penalty. 

The parameters $(\theta, \phi, \gamma, \omega)$ are optimized using the ADAM algorithm \citep{ADAM}, a widely adopted optimizer for gradient-based training. The ADAM optimizer, $\text{ADAM}_{\alpha,\beta_1,\beta_2}$ is defined by three hyperparameters: $\alpha$ (learning rate), $\beta_1$ (momentum parameter), and $\beta_2$ (second-moment paramter). 

The implementation is summarized as follows. The final estimated generator is given by $\hG = \hD\circ \hH$.

{\tiny
\begin{algorithm}[H]
\caption{Step 1: Representation Learning}\label{algo:1}
\begin{algorithmic}[1]
\Require iterations $T$, batch size $B$, $\text{ADAM}_{\alpha, \beta_1,\beta_2}$, decoder $D_\theta$, encoder $E_\phi$
% , initial parameters $(\theta^{(0)}, \phi^{(0)})$
\For{t =0, 1, 2, ..., $T-1$}
        \State $\{Y_i\}_{i=1}^B \gets \text{Sample}(\cU\cup\cP_Y, B)$
        \State $L\gets \frac{1}{B}\sum_{i=1}^B\dnorm{Y_i - D_{\theta^{(t)}}\circ E_{\phi^{(t)}} (Y_i)}_2$
        \State $(\theta^{(t+1)}, \phi^{(t+1)}) \gets (\theta^{(t)}, \phi^{(t)}) + \text{ADAM}_{\alpha, \beta_1, \beta_2}(-\nabla_{(\theta^{(t)},\phi^{(t)})}L)$
        \EndFor
        \State \Return $\hD = D_{\theta^{(T)}}$, $\hE = E_{\phi^{(T)}}$
\end{algorithmic}
\end{algorithm}

\begin{algorithm}[H]
\caption{Step 2: Conditional Generation}\label{algo:2}
\begin{algorithmic}[1]
\Require iterations $T$, batch size $B$, $\text{ADAM}_{\alpha^1, \beta_1^1,\beta_2^1}$, $\text{ADAM}_{\alpha^2, \beta_1^2,\beta_2^2}$, gradient penalty parameter $\lambda$, trained decoder $\hD$, trained encoder $\hE$, latent code conditional generator $H_\gamma$, critic $f_\omega$, critic iterations $J$
% , initial parameters $(\gamma^{(0)}, \omega^{(0)})$
\For{t =0, 1, 2, ..., $T-1$}
        \State $\{(X_i,Y_i)\}_{i=1}^B \gets \text{Sample}(\cL, B)$; \hspace{0.5em} $\{\eta_i\}_{i=1}^B \simiid \cN(0, I_d)$
        \State \textbf{(cLSDM)} $Z_i \gets \hat{D}\circ \hE(Y_i)$
            , $T_{\gamma^{(t)}} \gets \hat{D}\circ H_{\gamma^{(t)}}$; \textbf{(dLSDM)} $Z_i \gets \hE(Y_i)$, $T_{\gamma^{(t)}} \gets H_{\gamma^{(t)}}$
        % \If{cLSDM}
        %     \State $Z_i \gets \hat{D}\circ \hE(Y_i)$
        %     ;\hspace{0.5em} $T_{\gamma^{(t)}} \gets \hat{D}\circ H_{\gamma^{(t)}}$
        % \ElsIf{dLSDM}
        %     \State $Z_i \gets \hE(Y_i)$
        %     ;\hspace{0.5em} $T_{\gamma^{(t)}} \gets H_{\gamma^{(t)}}$
        % \EndIf
        \For{j=0, 1, 2, ..., $J-1$}
            \State $\{\epsilon_i\}_{i=1}^B \simiid U([0,1])$
            \State $V \gets \frac{1}{B} \sum_{i=1}^B f_{\omega^{(t)}}(X_i,Z_i) - f_{\omega^{(t)}}(X_i,T_{\gamma^{(t)}}(X_i, \eta_i))  + \lambda  \left(\dnorm{\nabla_{(x,z)}f_{\omega^{(t)}}(X_i, Z_i)}_2-1\right)^2 $
        \State $\omega^{(t)} \gets \omega^{(t)} + \text{ADAM}_{\alpha^1, \beta_1^1,\beta_2^1}(-\nabla_{\omega^{(t)}} V)$
        \EndFor
        \State $\omega^{(t+1)} \gets \omega^{(t)}$, $L \gets \frac{1}{B} \sum_{i=1}^Bf_{\omega^{(t+1)}}(X_i,T_{\gamma^{(t)}}(X_i, \eta_i))$
        % \State $\omega^{(t+1)} \gets \omega^{(t)}$
        % \State $L \gets \frac{1}{B} \sum_{i=1}^Bf_{\omega^{(t+1)}}(X_i,T_{\gamma^{(t)}}(X_i, \eta_i))$
        \State $\gamma^{(t+1)} \gets \gamma^{(t)} + \text{ADAM}_{\alpha^2, \beta_1^2,\beta_2^2}(-\nabla_{\gamma^{(t)}} L)$
        \EndFor
        \State \Return $\hG = \hD \circ H_{\gamma^{(T)}}$
\end{algorithmic}
\end{algorithm}
}

\section{Implementation Details of Experiments}
We denote a fully connected layer with output size $x$ by "$\text{FC}_x$"; a convolution layer with parameters (output channels, kernel size, stride, padding) = ($o,k,s,p$) by "$\text{Conv}(o,k,s,p)$", without a bias; a transposed convolution layer with parameters ($o,k,s,p$) by "$\text{ConvTrans}(o,k,s,p)$", without a bias; concatenation operation with vector $v$ by "$\text{Cat}(v)$"; batch normalization, layer normalization, spectral normalization, instance normalization module as "$\text{BN}$", "$\text{LN}$", "$\text{SN}$", "$\text{IN}$", respectively, where input channel size is given from context; Leaky ReLU activation with slope parameter $a$ as "$\text{LeakyReLU}(a)$";  An embedding layer with input $X$, number of discrete input $C$ and output shape $l$ as $\text{Embed}(X,C,l)$; Average pooling 2d by kernel size $k$ as $\text{AvgPool}(X,k)$;

\subsection{CelebA Image Super-resolution}
The architecture of neural networks are shown as follow:\\
\textbf{Residual Block:}
\begin{align*}
    \text{Res}_{i,o}(x) &\to \text{Conv}_{o,3,1,1} \to \text{BN} \to \text{ReLU} \\
    &\to \text{Conv}_{o,3,1,1} \to \text{BN} \to z'  \\
    &\to \text{ReLU}(\text{skip}(x)+z')\\
    \text{DownRes}_{i,o}(x) &\to \text{Conv}_{o,3,2,1} \to \text{BN} \to \text{ReLU} \\
    &\to \text{Conv}_{o,3,1,1} \to \text{BN} \to z'  \\
    &\to \text{ReLU}(\text{skip}(\text{AvgPool}(x,2))+z')\\
     \text{UpRes}_{i,o}(x) &\to \text{Up}(x,2,\text{bilinear})\to \text{Conv}_{o,3,1,1} \to \text{BN} \to \text{ReLU} \\
    &\to \text{Conv}_{o,3,1,1} \to \text{BN} \to z'  \\
    &\to \text{ReLU}(\text{skip}(\text{Up}(x,2,\text{bilinear}))+z')\\
    \text{skip}(x) &\to x \text{ if } i == o  \text{ else } \text{Conv}_{o,1,1,0}(x)
\end{align*}
If $\text{BN}$ is replaced with other normalization module, or $\text{ReLU}$ is replaced with other activation function, or that a different upsampling mode is used, it is denoted in the superscript. 

\textbf{Encoder ($E$):}
\begin{align*}
    Y &\to \text{Conv}_{64,3,1,1} \to \text{BN} \to \text{ReLU}\\
    &\to \text{Res}_{64,64} \to \text{DownRes}_{64,128}\\
    &\to \text{Res}_{128,128} \to \text{DownRes}_{128,256}\\
    &\to  \text{Res}_{256,256}  \to \text{Res}_{256,256} \\
    &\to \text{Conv}_{m_c, 3,1,1}
\end{align*}

\textbf{Decoder ($D$):}
\begin{align*}
    E(Y) &\to \text{Conv}_{256,3,1,1} \to \text{BN} \to \text{ReLU}\\
    &\to \text{Res}_{256,256} \to \text{Res}_{256,256}\\
    &\to \text{UpRes}_{256,128} \to \text{Res}_{128,128}\\
    &\to  \text{UpRes}_{128,64} \to \text{Res}_{64,64}\\
    &\to \text{Conv}_{m_c, 3,1,1} \to \text{Tanh}
\end{align*}

\textbf{Latent Code Generator ($H$):}
\begin{align*}
    X &\to \text{Cat}(X,\eta)\\
    &\to \text{Conv}_{128,3,1,1}\\
    &\to \text{Res}^{\text{IN}}_{128,128}\to \text{Res}^{\text{IN}}_{128,128}\to  \text{Res}^{\text{IN}}_{128,128}\\
    &  \to \text{Conv}_{m_c,3,1,1}\\
\end{align*}
where $\eta $ is $3\times 16\times 16$ i.i.d. standard Gaussian variable.

\textbf{cLSDM critic ($f$):}
\begin{align*}
   \text{Cat}( \text{Up}(X,4,\text{bilinear}), Y) &\to \text{Conv}_{64,3,1,1}\\
    &\to \text{DownRes}_{64,64} \to \text{DownRes}_{64,128 } \to \text{DownRes}_{256} \to \text{DownRes}_{256,512}\\
    &\to \text{AdaptiveAvgPool2d}(1)\\
    &\to \text{FC}_{1}
\end{align*}
In $\text{DownRes}$, BN are removed as suggested in the original work of WGAN-GP, and $\text{LeakyReLU} (x,0.2)$ is used instead of $\text{ReLU}$.

\textbf{cGAN critic ($f$):}
\begin{align*}
   \text{Cat}( \text{Up}(X,4,\text{bilinear}), Y) &\to \text{SN(Conv)}_{64,3,1,1}\\
    &\to \text{DownRes}_{64,64} \to \text{DownRes}_{64,128 } \to \text{DownRes}_{256} \to \text{DownRes}_{256,512}\\
    &\to \text{AdaptiveAvgPool2d}(1)\\
    &\to \text{SN(FC)}_{1}
\end{align*}
In $\text{DownRes}$, SN is applied to all Conv layer as in the original work of spectral norm GAN and $\text{LeakyReLU} (x,0.2)$ is used instead of $\text{ReLU}$. 

\textbf{dLSDM critic ($f$):}
\begin{align*}
   \text{Cat}(X, Y) &\to \text{Conv}_{64,3,1,1}\\
    &\to \text{DownRes}_{64,64} \to \text{DownRes}_{64,128 } \to \text{DownRes}_{256} \to \text{DownRes}_{256,512}\\
    &\to \text{AdaptiveAvgPool2d}(1)\\
    &\to \text{FC}_{1}
\end{align*}

\textbf{cGAN, cWGAN generator ($G$):}
\begin{align*}
    X  &\to \text{Cat}(X,\eta)\\
    &\to \text{Res}^{\text{IN}}_{128,128}\to \text{UpRes}^{\text{IN}}_{128,128} \to \text{UpRes}^{\text{IN}}_{128,128}\\
    &\to \text{Conv}_{3,3,1,1}\to \text{Tanh} \\
\end{align*}
where $\eta $ are $3\times 28\times 28$ i.i.d. standard Gaussian variable.

\textbf{cVAE encoder:}
\begin{align*}
   \text{Cat}( \text{Up}(X,4,\text{bilinear}), Y) &\to \text{Conv}_{128,3,1,1}\\
    &\to \text{Res}^{\text{IN}}_{128,128} \to \text{DownRes}^{\text{IN}}_{128,128 } \to \text{DownRes}^{\text{IN}}_{128,128} \\
    &\to (\text{Conv}_{2,3,1,1}, \text{Conv}_{2,3,1,1})
\end{align*}

\textbf{cVAE decoder:}
\begin{align*}
   \text{Cat}( X,Z) &\to \text{Conv}_{128,3,1,1} \\
    &\to \text{UpRes}^{\text{IN}}_{128,128} \to \text{UpRes}^{\text{IN}}_{128,128 } \to \text{Res}^{\text{IN}}_{128,128} \\
    &\to  \text{Conv}_{3,3,1,1}\to \text{Tanh}
\end{align*}

\textbf{LDM network:}
\begin{align*}
\text{TimeEmbed}(\sigma,256) &\to  (\sin(\log (\sigma ) * \alpha ), \cos(\log(\sigma) * \alpha))\\
&\to \text{FC}_{256}\to \text{SiLU} \to  \text{FC}_{256} \to Z_0\\
Z _0&\to \text{FC}_{3\times 16\times 16} \to Z_1\\
Z _0&\to \text{FC}_{3\times 16\times 16} \to Z_2\\
Z _0&\to \text{FC}_{3\times 16\times 16} \to Z_3\\
   \text{Cat}(Z,X,Z_1) &\to \text{Conv}_{128,3,1,1} \to \text{Res}^{\text{IN}}_{128,128}  \to  h_1\\
   \text{Cat}(h_1,X,Z_2) &\to \text{Res}^{\text{IN}}_{128,128}  \to  h_2\\
   \text{Cat}(h_2,X,Z_3) &\to \text{Res}^{\text{IN}}_{128,128}  \to  \text{Conv}_{m_c,3,1,1} \\
\end{align*}

For training the VQVAE in comparison with baselines, $1000$ sample size is taken as the validation set from the train set, trained for $150$ epochs with a batch size of $128$, and the optimizer $\text{ADAM}_{5e-4, 0.9,0.999}$. VQVAE uses commitment loss factor of $0.25$, quantization loss factor of $1$, EMA averaging of decay $0.99$, and number of embeddings $512$.  The model with the lowest validation reconstruction error is taken as the final model. We adopt the implementation of \cite{Pythae} to ensure correctness. The optimizers have a scheduler that reduce the learning rate by a factor of $2$, at epochs $30,50,75, 120$. 

For cLSDM critic $f$, we use critic iteration $J=5$, and trained with an optimizer $\text{ADAM}_{7.5e-4, 0,0.9}$ with Lipschitz penalty $\lambda=10$. $H$ is trained for $400$ epochs with a batch size of $25$, optimizer $\text{ADAM}_{7.5e-4, 0,0.9}$. For dLSDM, we use learning rates of $4e-4$ for both critic and generator, and a batch size of $50$, all else equal.

For cWGAN and cVAE, are trained for $400$ epochs with a batch size of $50$, optimizer $\text{ADAM}_{4e-4, 0,0.9}$, with critic iteration $J=5$. For LDM, it is trained for $50000$ epochs with optimizer $\text{AdamW}_{4e-4, \gamma =1e-2}$, $\gamma $ is weight decay, batch size of $50$. For cGAN, it is trained for $4000$ epochs, all else the same as cWGAN.

All models use ema averaging with decay rate $0.99$.  Further implementation details, including the specific commands to reproduce the results, can be found in the code provided.

The architecture of neural networks are shown as follow:\\
\textbf{Encoder ($E$):}
\begin{align*}
    Y\in \R^{32\times 32} &\to \text{Conv}_{128,3,1,1} \to \text{BN} \to LeakyReLU(0.1)\\
    &\to  \text{Conv}_{256,4,2,1} \to \text{BN} \to \text{LeakyReLU}(0.1)\\
    &\to  \text{Conv}_{256,4,2,1} \to \text{BN} \to \text{LeakyReLU}(0.1)\\
    &\to  \text{Conv}_{512,4,2,1} \to \text{BN} \to \text{LeakyReLU}(0.1)\\
    &\to  \text{Conv}_{1024,4,2,1} \to \text{BN} \to \text{LeakyReLU}(0.1)\\
    &\to \text{FC}_{256} \to \text{LeakyReLU}(0.1) \to \text{FC}_{m}\\
\end{align*}

\textbf{Decoder ($D$):}
\begin{align*}
    E(Y)\in \R^{m} &\to \text{FC}_{128\times 3\times 3} \to \text{BN} \to \text{ReLU}\\
    &\to \text{ConvTrans}_{64,5,2,1} \to \text{BN} \to \text{ReLU}\\
    &\to \text{ConvTrans}_{32,4,2,1} \to \text{BN} \to \text{ReLU}\\
    &\to \text{ConvTrans}_{1,4,2,1} \to \text{Sigmoid}
\end{align*}

\textbf{Latent Code Generator ($H$):}
\begin{align*}
    X\in \R^{p} &\to \text{Conv}_{128,5,1,1} \to  \text{ReLU}\\
    &\to  \text{Conv}_{256,4,1,2}  \to \text{ReLU}\\
    &\to  \text{Conv}_{512,4,1,2}  \to \text{ReLU}\\
    &\to  \text{Conv}_{1024,4,1,1}  \to \text{ReLU}\\
    &\to \text{Cat}(\eta) \to \text{FC}_{512} \to \text{ReLU}\\
    &\to \text{FC}_{256} \to \text{ReLU}\to \text{FC}_{128} \to \text{ReLU}\to \text{FC}_{m}\\
\end{align*}

\textbf{cLSDM critic ($f$):}
\begin{align*}
    X\in \R^{p} &\to \text{Conv}_{128,5,1,1} \to  \text{ReLU}\\
    &\to  \text{Conv}_{256,4,1,2}  \to \text{ReLU}\\
    &\to  \text{Conv}_{512,4,1,2}  \to \text{ReLU}\\
    &\to  \text{Conv}_{1024,4,1,1}  \to \text{ReLU} \to A\\
    Y\in \R^{p} &\to \text{Conv}_{128,4,1,2} \to  \text{ReLU}\\
    &\to  \text{Conv}_{256,4,1,2}  \to \text{ReLU}\\
    &\to  \text{Conv}_{512,4,1,2}  \to \text{ReLU}\\
    &\to  \text{Conv}_{1024,4,1,2}  \to \text{ReLU} \to B\\
    \text{Cat}(A,B) &\to \text{FC}_{128} \to \text{ReLU}
    \to \text{FC}_{1}
\end{align*}

\textbf{dLSDM critic ($f$):}
\begin{align*}
    X\in \R^{p} &\to \text{Conv}_{128,5,1,1} \to  \text{ReLU}\\
    &\to  \text{Conv}_{256,4,1,2}  \to \text{ReLU}\\
    &\to  \text{Conv}_{512,4,1,2}  \to \text{ReLU}\\
    &\to  \text{Conv}_{1024,4,1,1}  \to \text{ReLU} \to A\\
    Z \in \R^m &\to \text{Cat}(A) \to \text{FC}_{512} \to \text{ReLU}\\
    &\to \text{FC}_{256} \to \text{ReLU}\to \text{FC}_{128} \to \text{ReLU}\to \text{FC}_{1} \\
\end{align*}

For training the encoder-decoder pair, 1000 sample size is taken as the validation set from the train set. The encoder-decoder pair is trained for 200 epochs with a batch size of 100, and the optimizer $\text{ADAM}_{5e-3, 0.9,0.999}$. The encoder-decoder pair with the lowest validation reconstruction error is taken as the final model. We adopt the implementation of \cite{Pythae} to ensure correctness.

For the cLSDM, $H$ is trained for 300 epochs with a batch size of 20, the noise vector $\eta \sim \cN(0, I_{10})$, and an optimizer $\text{ADAM}_{1e-4, 0.5,0.999}$. For critic $f$, critic iteration $J$ is set to 3, and trained with an optimizer $\text{ADAM}_{1e-5, 0.5,0.999}$ with Lipschitz penalty $\lambda=100$. The optimizers have a scheduler that reduce the learning rate by a factor of $2$, at epochs $60,120,160$. For dLSDM, we set $\lambda=500$ and train for $1000$ epoch, the learning rate is reduced by a factor of $2$, at epochs $60, 120, 160, 500, 750$. Other parameters being the same.

For the baseline model, its model architecture is $D\circ H$. It is additional trained for 200 more epochs, totaling 500 epochs with an optimizer $\text{ADAM}_{1e-4, 0,0.9}$, as its decoder component is not trained in Step 1. The learning rate is reduced by a factor of $2$, at epochs $260, 320, 360$. Moreover, we set $\lambda=10$. Other parameters being the same as $LSDM$. 

\subsection{MNIST Class-conditional Generation}
The architecture of neural networks are shown as follow:\\
\textbf{Encoder ($E$):}
\begin{align*}
    Y\in \R^{1\times 28\times 28} &\to \text{Conv}_{32,4,2,1} \to \text{BN} \to \text{ReLU}\\
    &\to \text{Conv}_{64,4,2,1} \to \text{BN}\to \text{ReLU}\\
    &\to \text{FC}_{128,4,2,1} \to \text{BN}\to \text{ReLU}\\
    &\to  \text{FC}_{m} 
\end{align*}

\textbf{Decoder ($D$):}
\begin{align*}
    E(Y)\in \R^{m} &\to \text{FC}_{128\times 3\times 3} \to \text{BN} \to \text{ReLU}\\
    &\to \text{ConvTrans}_{64,5,2,1} \to \text{BN} \to \text{ReLU}\\
    &\to \text{ConvTrans}_{32,4,2,1} \to \text{BN} \to \text{ReLU}\\
    &\to \text{ConvTrans}_{1,4,2,1} \to \text{Sigmoid}
\end{align*}

\textbf{Latent Code Generator ($H$):}
\begin{align*}
    X\in \R^{p} &\to \text{Cat}(\text{Embed}(X,10,3\times 28\times 28),\eta)\\
    &\to \text{Conv}_{128,3,2,1}\to \text{BN} \to  \text{ReLU}\\
    &\to \text{Conv}_{64,3,2,1}\to \text{BN} \to  \text{ReLU}\\
    & \to \text{Conv}_{32,3,2,1}\to \text{BN} \to  \text{ReLU}\\
    & \to \text{FC}_{m}\\
\end{align*}
where $\eta $ are $3\times 28\times 28$ i.i.d. standard Gaussian variable.
\textbf{cLSDM critic ($f$):}
\begin{align*}
   \text{Cat}( \text{Embed}(X,10,3\times 28\times 28), Y) &\to \text{Conv}_{32,3,2,1} \to \text{LN}\to \text{LeakyReLU}(0.2)\\
    &\to \text{Conv}_{64,3,2,1} \to \text{LN}\to \text{LeakyReLU}(0.2)\\
    &\to \text{Conv}_{128,3,2,1} \to \text{LN}\to \text{LeakyReLU}(0.2)\\
    &\to \text{Conv}_{1,4,1,0}
\end{align*}

\textbf{cGAN critic ($f$):}
\begin{align*}
   \text{Cat}( \text{Embed}(X,10,3\times 28\times 28), Y) &\to \text{SN(Conv)}_{32,3,2,1} \to \text{LN}\to \text{LeakyReLU}(0.2)\\
    &\to \text{SN(Conv)}_{64,3,2,1} \to \text{LN}\to \text{LeakyReLU}(0.2)\\
    &\to \text{SN(Conv)}_{128,3,2,1} \to \text{LN}\to \text{LeakyReLU}(0.2)\\
    &\to \text{SN(Conv)}_{1,4,1,0}
\end{align*}

\textbf{dLSDM critic ($f$):}
\begin{align*}
    \text{Cat}(\text{Embed}(X,10,3\times 28\times 28),\text{FC}_{1\times 28\times 28}(Z))&\to \text{Conv}_{32,3,2,1} \to \text{LN}\to \text{LeakyReLU}(0.2)\\
    &\to \text{Conv}_{64,3,2,1} \to \text{LN}\to \text{LeakyReLU}(0.2)\\
    &\to \text{Conv}_{128,3,2,1} \to \text{LN}\to \text{LeakyReLU}(0.2)\\
    &\to \text{Conv}_{1,4,1,0}
\end{align*}
\textbf{cGAN, cWGAN generator ($G$):}
\begin{align*}
    X\in \R^{p} &\to \text{Cat}(\text{Embed}(X,10,3\times 28\times 28),\eta)\\
    &\to \text{ConvTrans}_{128,5,2,1}\to \text{BN} \to  \text{ReLU}\\
    &\to \text{ConvTrans}_{64,4,2,1}\to \text{BN} \to  \text{ReLU}\\
    & \to \text{ConvTrans}_{32,4,2,1}\to \text{BN} \to  \text{ReLU}\\
    & \to \text{ConvTrans}_{1,3,1,1} \to \text{Sigmoid}\\
\end{align*}
where $\eta $ are $3\times 28\times 28$ i.i.d. standard Gaussian variable.

\textbf{cVAE encoder:}
\begin{align*}
   \text{Cat}( \text{Embed}(X,10,3\times 28\times 28), Y) &\to \text{Conv}_{32,4,2,1} \to \text{BN}\to \text{ReLU}\\
    &\to \text{Conv}_{64,4,2,1} \to \text{BN}\to \text{ReLU}\\
    &\to \text{Conv}_{128,4,2,1} \to \text{BN}\to \text{ReLU}\\
    &\to \text{Conv}_{13,3,1,0} \\
    &\to (\text{FC}_{13}, \text{FC}_{13})
\end{align*}

\textbf{cVAE decoder:}
\begin{align*}
   \text{Cat}( \text{Embed}(X,10,3\times 3\times 3), \text{FC}_{3\times 3\times 3}(Z)) &\to \text{ConvTrans}_{128,5,2,1} \to \text{BN}\to \text{ReLU}\\
    &\to \text{ConvTrans}_{64,4,2,1} \to \text{BN}\to \text{ReLU}\\
    &\to \text{ConvTrans}_{32,4,2,1} \to \text{BN}\to \text{ReLU}\\
    &\to \text{ConvTrans}_{1,3,1,1} \\
    &\to \text{Sigmoid}
\end{align*}

\textbf{LDM network:}
\begin{align*}
\text{TimeEmbed}(\sigma,64) &\to  (\sin(\log (\sigma ) * \alpha ), \cos(\log(\sigma) * \alpha))\\
&\to \text{FC}_{64}\to \text{SiLU} \to  \text{FC}_{64}\\
   \text{Cat}( \text{TimeEmbed}(\sigma, 64), Z) &\to \text{FC}_{3\times 28\times 28} \to Z'\\
   \text{Cat}( \text{Embed}(10, X, 3\times 28\times 28), Z')&\to \text{Conv}_{128,3,2,1} \to \text{BN}\to \text{ReLU}\\
    &\to \text{Conv}_{64,3,2,1} \to \text{BN}\to \text{ReLU}\\
    &\to \text{Conv}_{32,3,2,1} \to \text{BN}\to \text{ReLU}\\
    &\to \text{FC}_{13}
\end{align*}

For training the autoencoder in comparison with baselines, $1000$ sample size is taken as the validation set from the train set. All autoencoders (AE,WAE,VQVAE) are trained for $100$ epochs with a batch size of $100$, and the optimizer $\text{ADAM}_{1e-3, 0.9,0.999}$. The model with the lowest validation reconstruction error is taken as the final model. We adopt the implementation of \cite{Pythae} to ensure correctness. The optimizers have a scheduler that reduce the learning rate by a factor of $2$, at epochs $30,50,75$. For ablation studies with fixed $n$ and varying $N$, a learning rate of $5e-4$ and batch size of $20$ is used instead due to lower sample size and to prevent overfitting. Moreover, for $n+N=(250,500,750)$, a smaller sample size of $(20,20,50)$ is taken as the validation set, due to insufficient total sample size. For the ablation study with fixed $n+N=3000$ and varying $n$, a learning rate of $5e-4$ , batch size of $20$ and validation set size of $300$ is used. Other parameters remain the same.

For cLSDM and dLSDM critic $f$, critic iteration $J=5$, and trained with an optimizer $\text{ADAM}_{3e-4, 0,0.9}$ with Lipschitz penalty $\lambda=10$. $H$ is trained for $200$ epochs with a batch size of $25$, optimizer $\text{ADAM}_{3e-4, 0,0.9}$.

For cWGAN, cGAN, cVAE, models are trained for $200$ epochs with a batch size of $25$, optimizer $\text{ADAM}_{3e-4, 0,0.9}$, with critic iteration $J=5$. For LDM, it is trained for $20000$ epochs with optimizer $\text{AdamW}_{5e-4, \gamma =1e-2}$, $\gamma $ is weight decay, batch size of $125$. 

All models use ema averaging with decay rate $0.999$. Further implementation details, including the specific commands to reproduce the results, can be found in the code provided.

\bibliographystyle{apalike}
\bibliography{reference}